# Experimental evidence of progressive ChatGPT models self-convergence

**Konstantinos F. Xylogiannopoulos [a,b], Petros Xanthopoulos [c], Panagiotis Karampelas [d], Georgios A. Bakamitsos[e]**

[a] University of Calgary, Department of Computer Science, Calgary, AB, Canada, kostasfx@yahoo.gr
[b] Stetson University, School of Business Administration, DeLand, FL, USA, kxylogiannopoulos@stetson.edu
[c] Stetson University, Business Systems & Analytics Department, School of Business Administration, DeLand, FL, USA, pxanthopoulos@stetson.edu
[d] Hellenic Air Force Academy, Dekeleia, Greece, panagiotis.karampelas@hafa.haf.gr
[e] Stetson University, Marketing Department, School of Business Administration, DeLand, FL, USA, bakamitsos@stetson.edu

## Abstract

Large Language Models (LLMs) that undergo recursive training on synthetically generated data are susceptible to *model collapse*, a phenomenon marked by the generation of meaningless output. Existing research has examined this issue from either theoretical or empirical perspectives, often focusing on a single model trained recursively on its own outputs. While prior studies have cautioned against the potential degradation of LLM output quality under such conditions, no longitudinal investigation has yet been conducted to assess this effect over time. In this study, we employ a text similarity metric to evaluate different ChatGPT models' capacity to generate diverse textual outputs. Our findings indicate a measurable decline of recent ChatGPT releases' ability to produce varied text, even when explicitly prompted to do so, by setting the temperature parameter to one. The observed reduction in output diversity may be attributed to the influence of the amounts of synthetic data incorporated within their training datasets as the result of internet infiltration by LLM generated data. The phenomenon is defined as *model self-convergence* because of the gradual increase of similarities of produced texts among different ChatGPT versions.

## 1 Introduction

Generative pretrained models (GPT) are usually trained in large quantities of data, usually scrapped from the internet. Once a pretrained model is established one can come up with refinements that can happen with further training of these models with new, sometimes domain specific data. This process is known as fine tuning. In (Shumailov et al., 2024) authors showed that GPT models that are fine-tuned with machine generated data eventually collapse which means that after a few training recursions their output is predominantly “gibberish”. At the time of that publication, the majority of GPT models were trained primarily on human-generated data; however, the authors warned about potential adverse implications arising from the increasing incorporation of artificially generated data in the training of subsequent GPT models.

Subsequently, the phenomenon of model collapse has been examined from both theoretical and empirical perspectives, with researchers investigating its underlying causes and identifying the optimal balance between human-generated and synthetic data required to prevent its occurrence. In (Dohmatob, Feng, & Kempe, 2024) authors derive analytical formulas for understanding the model collapse problem in simpler data analysis methods such as regression. In (Barzilai & Shamir, 2025) authors identify special cases of algorithms where model collapse is avoidable. Other literature tries to establish training data strategies that mitigate the model collapse. In (Dey & Donoho, 2024; Gerstgrasser et al., 2024) authors provided theoretical evidence that if the real data used to train the original model are accumulated, instead of replaced then the test error of the subsequent models has a finite upper bound. In (Dohmatob, Feng, Yang, et al., 2024) authors study how synthetic data affect the neural network scaling laws.

Other literature proposes methods as to how to mitigate this negative effect by either introduce weights to data (He et al., 2025) however in (Dohmatob, Feng, Subramonian, et al., 2024) authors showed that the model collapse can’t be mitigated by data weighting unless one asymptotically removes all the synthetic data. Larger models can mitigate the effect but cannot avoid it. This negative result is known as strong model collapse.

It is interesting that the concept of model collapse is not limited to the LLMs but also extends to the visional language models (VLMs) and text to image diffusion models (Hu et al., 2025; Yoon et al., 2025).

The value of AI generated text detection has been identified as one approach that can help mitigate the negative effect of model collapse (Drayson et al., 2025). Even though most academic research makes the silent assumption that the training process is a well monitored and closed process the practical reality is vastly different. Many companies such as Google, Open AI or Anthropic use retrieval-augmented generation (RAG) methodologies to allow their LLM models access to the internet and therefore to enhance their answering abilities especially when they are prompted with questions for which the answer is not part of their training dataset. As one can easily point out the infusion of the LLMs with “lazy” internet information that might have been created by LLMs. There is preliminary evidence that RAG LLMs are prone to model collapse as well (Wang et al., 2025) and at the same time they are known to suffer from other issues such as safety (An et al., 2025).

The initial version of ChatGPT, released on November 30, 2022, was based on the ChatGPT-3.5 model and trained primarily on human-generated internet data. The most recent iteration, ChatGPT-5.2, operates in an online environment where a growing proportion of the available data is synthetic, generated by artificial systems. Despite this development, no longitudinal study has systematically examined potential model collapse patterns across successive GPT versions. In this paper, we employ textual pattern-matching techniques to evaluate the capacity of different GPT models to produce diverse textual outputs. Given the probabilistic nature of neural networks, we hypothesize that increasing exposure to synthetic data constrains model diversity, leading to convergence toward similar output structures, even when the stochastic parameter (temperature) is set to its maximum value of one.

### 1.a Large Language Models

A Large Language Model (LLM) as a specific type of Deep Neural Networks (DNN) is based on two fundamental pillars. First and foremost, as in every Artificial Intelligence technology, is the training dataset, i.e., internet, while the second is the algorithmic method used to create the model, i.e., transformers.

For LLMs the training dataset is very specific and simple in its source. Everything that exists in digital form is used to train an LLM. According to OpenAI every available digital information has been used as training dataset for ChatGPT, with different time milestones known as knowledge cutoff date (OpenAI, 2026a). Most models have different knowledge cutoff dates, however, there are some models with the same knowledge cutoff date, meaning that what practically differentiates one model from the other is the algorithmic approach, mostly from the perspective of parameters number, or fine tuning for specific purposes (OpenAI, 2026a).

From an algorithmic perspective, an LLM is in principle an autoregressive predictor of the next possible token (a sub-word sequence of characters), which will form a text using the Transformer architecture (Vaswani et al., 2017). First, the tokenization process is executed where the input text is broken down to small word subunits (tokens), including digits and special characters, and a numerical value is assigned to each one. Then the tokens pass through a Transformer architecture that uses the Self-Attention mechanism to calculate weights for each token. Each token weight represents the contextual relevance of the token in the sequence to every other token, regardless of its distance. For every token in the sequence, the model generates a Query (Q), Key (K) and Value (V) vector, which is used to calculate the Attention Score. This is calculated through the dot product of specific token Q with the K of all other tokens to measure relevance as attention scores. These scores, over the square root of the dimension $d_k$ of the K vectors (used for scaling purposes), are transformed into probabilities using a SoftMax function and then the final attention output is created by applying the probabilities as weights to the V, as presented by the formula:

$$Attention(Q,K,V) = SoftMax(\frac{QK^T}{\sqrt{d_k}})V$$

After all context related relationships have been calculated, the model produces a raw score $z_i$ for every possible next token that exists in its vocabulary. These scores are called logits and for better interpretation are converted into probabilities, after applying the Temperature (T) parameter to scale them, as in the formula:

$$P(x_i) = \frac{e^{(\frac{z_i}{T})}}{\sum_j e^{(\frac{z_j}{T})}}$$

This step is important because the Temperature is applied to scale logits and differentiate the model from deterministic when $T \rightarrow 0$ (practically T=0 is not possible since T is in the denominator, but it is usually used for simplicity), to stochastic when $T = 1$, and completely random when $T \rightarrow \infty$ since the distribution approaches uniform. The temperature definition describes the likelihood of a token to be selected and, therefore, determinism (low variance) and stochasticity (high variance) express the magnitude of this likelihood. If T=0 then high probable tokens get their scores further increased while low probable tokens have their scores further decreased, leading to a huge spread between them. On the contrary, if T=1 then this range is significantly reduced, allowing least probable tokens to be selected. For values T>1 the model will fluctuate more, and it could make it not applicable for meaningful text creation, yet, can be very useful for other abstract and fictional tasks like image or music creation (Appendix I).

After temperature application, the model states a probability distribution from the scaled logits. Then a Top-P or Top-K filtering is applied to restrict candidates to the most probable.

Instead of choosing from the entire LLM vocabulary, there is a filtering of the most probable tokens either with the use of the top probable tokens based on cumulative distribution (Top-P), e.g., the top tokens where their probability sums up to more than 80%, or with the use of the top tokens based on the order of probability (Top-K), e.g., the top 10 tokens as they are ordered based on their probability. With this process very unlikely to occur tokens, which could produce nonsense, are truncated from the selection list. Finally, a SoftMax function is applied to the filtered Top-P or Top-K logits to recreate a probability distribution across the filtered sample, and then the final output token is the random selection from the Top-P or Top-K sample. The model appends the selected token to the sequence, and the process repeats until termination, when the sequence reaches a predefined length or the model generates an End of Sequence (EoS) token.

According to this brief description it can be observed that what significantly differentiates models' performance is, from the one hand, the training dataset as each knowledge cutoff date includes more data, and, from the other hand, their size with regard the number of parameters used since the models began to get significantly larger with every new version.

### 1.b Pattern Detection

**Definition 1:** We define as "Similarity Percentage Ratio (SPR) between two texts A and B for pattern length $l$", the percentage ratio of the total number of words from the non-overlapping parts of the common patterns of the specific length $l$, over the length of each text. If the texts have different lengths, the similarities should vary as percentage ratios, regardless of the same number of common words.

Similarity Percentage Ratio can be described with the following example. In the texts "the brown fox turned to escape hunter's trap" and "the brown fox turned to opposite direction to avoid hunter's trap ahead" we have as common patterns of length four the patterns "the brown fox turned" and "brown fox turned to". Therefore, the non-overlapping words of these patterns are five ("the brown fox turned to") and the first text has SPR five words over eight, or 62.5%, and the second has five words over twelve, or 41.7%.

In previous research, we presented novel methodologies to identify paraphrased text, of an original human-created, that has been created with the use of ChatGPT (Xylogiannopoulos et al, 2024, 2025, 2026). These methodologies resulted in very high scores, up to or above 95%, in all standard metrics such as accuracy, sensitivity, specificity, precision and F1-score. The paraphrased text identification is based on the All Repeated Patterns Detection (ARPaD) algorithm, for which one of its unique advantages is that it allows the deterministic detection of every pattern exists at least twice in multiple texts (Xylogiannopoulos et al, 2024, 2025, 2026). For every original and paraphrased text, all repeated patterns are detected for multiple pattern lengths, having pattern lengths range from 3 to 15 words. The reason behind this is that single words and very small patterns of two words are practically insignificant since it is very common to occur or could be considered random noise. For very long patterns (above 15), with regard to the full text size, the Probabilistic Existence of Longer Repeated Pattern Theorem (Xylogiannopoulos et al, 2016; Xylogiannopoulos, 2017) guarantees that they do not exist and, if they do, they can be constructed from shorter overlapping patterns with moving starting position (Xylogiannopoulos et al., 2014) as in definition example. Based on the detected repeated patterns, the similarity percentage between texts is calculated and according to this value, a text can be categorized as human or AI.

What has been observed from these use-cases is that paraphrases have very low SPR to the original, human, text for short pattern lengths and none or extremely low SPRs for long patterns. Despite this, paraphrases with temperature 0 tend to have very high SPRs among them, while with temperature 1 they have lower similarity, but not significantly low for short pattern lengths. Still, for temperature 1 the average SPRs fall rapidly as the pattern length increases, compared to temperature 0. This is expected based on the structure of the LLMs since temperature is the parameter that allows an LLM to oscillate between deterministic and, therefore, producing similar paraphrases, to stochastic with produced paraphrases that significantly differentiate.

## 2 Problem Definition

As the LLMs evolve and become larger, with billions of parameters, and widely used globally, another very important factor that can affect their quality is the training dataset. Since the training dataset includes all available digital information, this factor becomes crucial for two reasons:

1. As LLMs usage is increasing among humans (lately also with the use of AI agents), more information is created and added to the globally available digital information
2. The production of artificial text is rapid and outperforms any human knowledge and text creation

With the current research work we want to:

1. Examine how the training dataset can affect the quality of ChatGPT as the knowledge cutoff date changes for every newer version
2. Evaluate the performance of ChatGPT to differentiate from itself as the LLM's versions increase

## 3 Design of Experiment

To evaluate the quality and performance of ChatGPT, the concept of paraphrased text should be used. Paraphrases of original, verified, human texts, should be created, with different ChatGPT model versions having different, but also same, knowledge cutoff dates.

Then, the level of SPR between the paraphrases of the same original text per model, temperature and pattern length, should be examined. Finally, the comparison of the SPRs of the multiple pattern lengths follows to examine how they differentiate between models and their knowledge cutoff dates. If they differentiate, it is also important to (a) examine in which direction (older to newer or reversely), and (b) quantify how much between each version. Higher similarity means that the model cannot differentiate between paraphrases and practically repeats itself, while low similarity means that the model is more abstract and innovative.

**Definition 2:** We define as "Large Language Model Self-Convergence", the phenomenon where the Similarity Percentage Ratios between paraphrases of the same original text, pattern length and temperature, increases among different versions of the LLM, for different knowledge cutoff dates.

To identify model self-convergence, we design an experiment with the following rules:

1. Use proven human created text as original for comparison purposes
2. Human text should be as visible and accessible as possible in its digital form
3. Human text should have everyday use and, therefore, it is possible to initiate further requests by LLMs to reproduce its content and context
4. Use two different human sources focusing on the same concept to use the second as control dataset
5. Use multiple ChatGPT versions with wide timespan of knowledge cutoff dates
6. Create multiple paraphrases for each text with each model and temperatures zero and one
7. Retain for all models and temperatures the same prompt
8. For every original text, temperature and model, create paraphrases with the ChatGPT API
9. One paraphrase per text should be created per round with time gap, to avoid possible memory/cache effects by the ChatGPT API
10. Identify all common patterns among original and ChatGPT paraphrases for several pattern lengths

11. Calculate the average similarity percentage ratio (SPR) for all paraphrases and pattern lengths
12. Analyze results to determine possible model self-convergence

An important note here is that the selection of temperature values in rule six is based on the finding that for temperature values above one, all ChatGPT models create nonsense text with special characters (Appendix I). Only the latest version 5.2 seems to bypass this problem for short texts, but only if the reasoning parameter is set explicitly to “none” (OpenAI, 2026b). Therefore, for comparability purposes of the results between the models, temperature zero and one should be used.

## 4 Dataset and Models

Following the design of experiment with the above mentioned rules, we execute our experiment as follows:

1. We use two datasets from the very famous study guides CliffNotes and SparkNotes. These study guides are human-created summaries of well-known English literature books, which are heavily used for decades by students at schools and colleges for their studies. Both user guides are old enough to guarantee that text in them are entirely human-created, and they also exist in digital form. Additionally, the dataset was retrieved from Kryciski et al. (2022), which was published before any LLM being widely available (first release 2021)
2. We select as original human text, summaries from 443 chapters in total, combined from multiple literature books, as found in Kryciski et al. (2022). The selection was based on two factors:
    a. A summary of a book chapter should exist in both study guides
    b. Summary length is between 100 and 2,000 words. The reason for limiting the minimum number of words per summary is that we do not want to overfit our results. For very short texts, ChatGPT cannot adequately differentiate between paraphrases due to the limited number of words and vocabulary. The same occurs with very long texts because ChatGPT tends to create short paraphrases compared to the original text (Xylogiannopoulos et. al, 2024, 2025, 2026)
3. We create paraphrases using the same prompt for all models and temperatures: “Answer ONLY the question, no extra context. Please paraphrase the following text: <TEXT>”

4. For temperature 0 we create three paraphrases per text and for temperature 1 five paraphrases. The reason for having less temperature 0 paraphrases is due to its deterministic nature and, therefore, low variation that will not affect the results
5. We analyze all created texts and discover SPRs for pattern lengths from 3 to 20
6. Based on the results, we conclude the analyses of spread and trends among all models, temperatures, and pattern lengths

For ChatGPT we have used seven different versions with five different knowledge cutoff dates, as can be seen in Table 1 (OpenAI, 2026a).

Table 1 ChatGPT versions and their corresponding knowledge cutoff and release dates

| ChatGPT Version | Knowledge Cutoff Date | Release Date |
| --- | --- | --- |
| GPT-3.5 Turbo | 01/09/2021 | 30/11/2022 |
| GPT-4 Turbo | 01/12/2023 | 09/04/2024 |
| GPT-4o | 01/06/2024 | 13/05/2024 |
| GPT-4.1 | 01/06/2024 | 15/04/2025 |
| GPT-5 | 30/09/2024 | 07/08/2025 |
| GPT-5.1 | 30/09/2024 | 12/11/2025 |
| GPT-5.2 | 30/08/2025 | 11/12/2025 |

Two model pairs have the same knowledge cutoff date (4o with 4.1 and 5 with 5.1). Unfortunately, ChatGPT API for version 5 does not allow the modification of the temperature value, which is by default equal to one. Regardless, it has been used in the experiment because temperature 1 is the most important value to investigate LLM stochasticity performance.

The distribution of words per text paraphrases group and temperatures, i.e., temperature 0 group with three paraphrases and temperature 1 with five paraphrases, for each one of the two study guides are close to original human texts (Appendix I). However, newer models have better distribution, meaning that the length of the paraphrased texts is closer to the length of the original texts with regard the number of words used to create the paraphrase, and it is expected to have better results than older models, i.e., have significant variation among paraphrases and, therefore, lower SPRs.

## 5 Results

The initial results of the analysis can be seen Fig. 1 and Fig. 2 where, for study guide CliffNotes and both temperatures 0 and 1 the SPR means per paraphrases group, for all

Comparison between Models Mean Similarity Percentages per Pattern Length for CliffNotes (temp=0)

**ChatGPT-3.5turbo (09/2021)** — Pattern Length [3]

| | CliffNotes | CGPT_p=01 | CGPT_p=02 | CGPT_p=03 | SparkNotes |
|---|---|---|---|---|---|
| CliffNotes | 33.68 | 19.25 | 19.44 | 19.76 | 8.46 |
| CGPT_p=01 | 38.95 | 88.85 | 77.06 | 77.80 | 8.98 |
| CGPT_p=02 | 39.23 | 76.00 | 89.75 | 78.86 | 9.08 |
| CGPT_p=03 | 39.63 | 77.18 | 79.22 | 90.54 | 8.88 |
| SparkNotes | 7.82 | 3.81 | 3.81 | 3.77 | 13.15 |

**ChatGPT-3.5turbo (09/2021)** — Pattern Length [20]

| | CliffNotes | CGPT_p=01 | CGPT_p=02 | CGPT_p=03 | SparkNotes |
|---|---|---|---|---|---|
| CliffNotes | 5.47 | 0.19 | 0.21 | 0.18 | 0.04 |
| CGPT_p=01 | 0.67 | 62.76 | 49.34 | 49.88 | 0.00 |
| CGPT_p=02 | 0.67 | 49.13 | 66.66 | 54.11 | 0.00 |
| CGPT_p=03 | 0.66 | 49.70 | 54.33 | 67.68 | 0.00 |
| SparkNotes | 0.06 | 0.00 | 0.00 | 0.00 | 0.11 |

**ChatGPT-4turbo (12/2023)** — Pattern Length [3]

| | CliffNotes | CGPT_p=01 | CGPT_p=02 | CGPT_p=03 | SparkNotes |
|---|---|---|---|---|---|
| CliffNotes | 28.04 | 14.18 | 14.00 | 14.09 | 8.47 |
| CGPT_p=01 | 19.95 | 85.14 | 74.10 | 73.78 | 6.54 |
| CGPT_p=02 | 19.80 | 74.13 | 85.68 | 74.67 | 6.60 |
| CGPT_p=03 | 19.94 | 73.94 | 74.73 | 85.79 | 6.76 |
| SparkNotes | 7.82 | 4.06 | 4.10 | 4.24 | 13.82 |

**ChatGPT-4turbo (12/2023)** — Pattern Length [20]

| | CliffNotes | CGPT_p=01 | CGPT_p=02 | CGPT_p=03 | SparkNotes |
|---|---|---|---|---|---|
| CliffNotes | 5.17 | 0.00 | 0.00 | 0.00 | 0.04 |
| CGPT_p=01 | 0.00 | 57.74 | 44.31 | 43.47 | 0.00 |
| CGPT_p=02 | 0.00 | 44.21 | 59.46 | 45.47 | 0.00 |
| CGPT_p=03 | 0.00 | 43.52 | 45.66 | 58.56 | 0.00 |
| SparkNotes | 0.06 | 0.00 | 0.00 | 0.00 | 0.11 |

**ChatGPT-4o (06/2024)** — Pattern Length [3]

| | CliffNotes | CGPT_p=01 | CGPT_p=02 | CGPT_p=03 | SparkNotes |
|---|---|---|---|---|---|
| CliffNotes | 41.52 | 29.41 | 29.49 | 29.14 | 8.46 |
| CGPT_p=01 | 40.08 | 93.89 | 86.17 | 87.01 | 8.18 |
| CGPT_p=02 | 40.01 | 85.75 | 93.78 | 87.00 | 7.99 |
| CGPT_p=03 | 39.65 | 86.78 | 87.18 | 94.41 | 8.09 |
| SparkNotes | 7.82 | 5.37 | 5.30 | 5.35 | 13.49 |

**ChatGPT-4o (06/2024)** — Pattern Length [20]

| | CliffNotes | CGPT_p=01 | CGPT_p=02 | CGPT_p=03 | SparkNotes |
|---|---|---|---|---|---|
| CliffNotes | 5.62 | 0.36 | 0.39 | 0.20 | 0.04 |
| CGPT_p=01 | 0.43 | 75.00 | 60.55 | 62.66 | 0.04 |
| CGPT_p=02 | 0.44 | 60.41 | 75.50 | 63.39 | 0.04 |
| CGPT_p=03 | 0.23 | 62.65 | 63.55 | 77.09 | 0.04 |
| SparkNotes | 0.06 | 0.02 | 0.02 | 0.02 | 0.11 |

**ChatGPT-4.1 (06/2024)** — Pattern Length [3]

| | CliffNotes | CGPT_p=01 | CGPT_p=02 | CGPT_p=03 | SparkNotes |
|---|---|---|---|---|---|
| CliffNotes | 36.10 | 24.23 | 24.23 | 24.21 | 8.47 |
| CGPT_p=01 | 31.64 | 93.07 | 85.97 | 86.74 | 8.17 |
| CGPT_p=02 | 31.77 | 86.29 | 93.66 | 87.27 | 8.27 |
| CGPT_p=03 | 31.66 | 86.67 | 86.94 | 93.86 | 8.22 |
| SparkNotes | 7.82 | 5.71 | 5.76 | 5.74 | 14.08 |

**ChatGPT-4.1 (06/2024)** — Pattern Length [20]

| | CliffNotes | CGPT_p=01 | CGPT_p=02 | CGPT_p=03 | SparkNotes |
|---|---|---|---|---|---|
| CliffNotes | 5.23 | 0.09 | 0.09 | 0.04 | 0.04 |
| CGPT_p=01 | 0.11 | 75.26 | 61.82 | 62.63 | 0.04 |
| CGPT_p=02 | 0.11 | 61.96 | 76.16 | 63.40 | 0.04 |
| CGPT_p=03 | 0.06 | 62.68 | 63.33 | 76.95 | 0.04 |
| SparkNotes | 0.06 | 0.02 | 0.02 | 0.02 | 0.11 |

**ChatGPT-5.1 (09/2024)** — Pattern Length [3]

| | CliffNotes | CGPT_p=01 | CGPT_p=02 | CGPT_p=03 | SparkNotes |
|---|---|---|---|---|---|
| CliffNotes | 42.87 | 31.67 | 31.00 | 31.40 | 8.46 |
| CGPT_p=01 | 35.01 | 94.20 | 87.12 | 87.25 | 7.30 |
| CGPT_p=02 | 34.37 | 87.28 | 93.67 | 86.91 | 7.37 |
| CGPT_p=03 | 34.88 | 87.56 | 87.03 | 93.97 | 7.43 |
| SparkNotes | 7.82 | 6.12 | 6.17 | 6.16 | 13.94 |

**ChatGPT-5.1 (09/2024)** — Pattern Length [20]

| | CliffNotes | CGPT_p=01 | CGPT_p=02 | CGPT_p=03 | SparkNotes |
|---|---|---|---|---|---|
| CliffNotes | 5.30 | 0.15 | 0.07 | 0.11 | 0.04 |
| CGPT_p=01 | 0.17 | 74.68 | 59.58 | 60.92 | 0.04 |
| CGPT_p=02 | 0.09 | 59.70 | 73.68 | 59.70 | 0.04 |
| CGPT_p=03 | 0.12 | 61.13 | 59.67 | 74.92 | 0.04 |
| SparkNotes | 0.06 | 0.02 | 0.02 | 0.02 | 0.11 |

**ChatGPT-5.2 (08/2025)** — Pattern Length [3]

| | CliffNotes | CGPT_p=01 | CGPT_p=02 | CGPT_p=03 | SparkNotes |
|---|---|---|---|---|---|
| CliffNotes | 33.52 | 19.76 | 19.89 | 19.82 | 8.46 |
| CGPT_p=01 | 24.41 | 88.24 | 78.36 | 77.51 | 6.89 |
| CGPT_p=02 | 24.41 | 78.16 | 88.96 | 78.24 | 6.84 |
| CGPT_p=03 | 24.42 | 77.40 | 78.31 | 88.24 | 6.79 |
| SparkNotes | 7.82 | 5.13 | 5.06 | 5.05 | 14.03 |

**ChatGPT-5.2 (08/2025)** — Pattern Length [20]

| | CliffNotes | CGPT_p=01 | CGPT_p=02 | CGPT_p=03 | SparkNotes |
|---|---|---|---|---|---|
| CliffNotes | 5.17 | 0.00 | 0.00 | 0.00 | 0.04 |
| CGPT_p=01 | 0.00 | 50.39 | 37.07 | 35.34 | 0.00 |
| CGPT_p=02 | 0.00 | 36.98 | 51.85 | 36.84 | 0.00 |
| CGPT_p=03 | 0.00 | 35.34 | 37.02 | 49.97 | 0.00 |
| SparkNotes | 0.06 | 0.00 | 0.00 | 0.00 | 0.11 |

Color scale: 0, 20, 40, 60, 80, 100

*Fig. 1 Comparison between groups per model means Similarity Percentage Ratios for pattern length 3 and 20 and temperature 0*

Comparison between Models Mean Similarity Percentages per Pattern Length for CliffNotes (temp=1)

**ChatGPT-3.5turbo (09/2021) — Pattern Length [3]**

| | CliffNotes | CGPT_p=01 | CGPT_p=02 | CGPT_p=03 | CGPT_p=04 | CGPT_p=05 | SparkNotes |
|---|---|---|---|---|---|---|---|
| CliffNotes | 42.21 | 15.71 | 16.50 | 16.91 | 15.79 | 16.13 | 8.48 |
| CGPT_p=01 | 33.41 | 67.97 | 34.82 | 34.88 | 34.51 | 34.15 | 8.11 |
| CGPT_p=02 | 33.27 | 33.90 | 67.15 | 35.08 | 33.41 | 33.24 | 8.65 |
| CGPT_p=03 | 34.42 | 33.90 | 35.29 | 67.49 | 33.61 | 33.20 | 8.48 |
| CGPT_p=04 | 33.21 | 34.61 | 34.36 | 34.27 | 67.21 | 34.13 | 8.54 |
| CGPT_p=05 | 34.10 | 33.99 | 34.16 | 34.02 | 34.07 | 67.90 | 8.59 |
| SparkNotes | 7.82 | 3.31 | 3.49 | 3.36 | 3.44 | 3.37 | 14.67 |

**ChatGPT-3.5turbo (09/2021) — Pattern Length [20]**

| | CliffNotes | CGPT_p=01 | CGPT_p=02 | CGPT_p=03 | CGPT_p=04 | CGPT_p=05 | SparkNotes |
|---|---|---|---|---|---|---|---|
| CliffNotes | 5.30 | 0.03 | 0.03 | 0.00 | 0.06 | 0.04 | 0.04 |
| CGPT_p=01 | 0.16 | 1.87 | 0.32 | 0.50 | 0.51 | 0.32 | 0.00 |
| CGPT_p=02 | 0.04 | 0.35 | 2.23 | 0.53 | 0.67 | 0.61 | 0.00 |
| CGPT_p=03 | 0.00 | 0.81 | 0.66 | 2.08 | 0.72 | 0.37 | 0.00 |
| CGPT_p=04 | 0.08 | 0.69 | 0.77 | 0.72 | 2.31 | 0.56 | 0.00 |
| CGPT_p=05 | 0.09 | 0.35 | 0.60 | 0.38 | 0.57 | 1.38 | 0.00 |
| SparkNotes | 0.06 | 0.00 | 0.00 | 0.00 | 0.00 | 0.00 | 0.11 |

**ChatGPT-4turbo (12/2023) — Pattern Length [3]**

| | CliffNotes | CGPT_p=01 | CGPT_p=02 | CGPT_p=03 | CGPT_p=04 | CGPT_p=05 | SparkNotes |
|---|---|---|---|---|---|---|---|
| CliffNotes | 33.86 | 12.36 | 12.04 | 11.63 | 11.93 | 12.04 | 8.48 |
| CGPT_p=01 | 17.38 | 61.65 | 32.32 | 32.60 | 32.50 | 31.83 | 5.99 |
| CGPT_p=02 | 17.11 | 32.55 | 61.48 | 32.15 | 32.42 | 32.61 | 5.78 |
| CGPT_p=03 | 16.60 | 32.92 | 32.47 | 61.36 | 32.58 | 32.04 | 5.76 |
| CGPT_p=04 | 16.97 | 32.72 | 32.41 | 32.23 | 61.53 | 31.81 | 5.70 |
| CGPT_p=05 | 17.08 | 32.11 | 32.69 | 31.98 | 31.98 | 61.23 | 5.93 |
| SparkNotes | 7.82 | 3.72 | 3.60 | 3.58 | 3.47 | 3.62 | 15.28 |

**ChatGPT-4turbo (12/2023) — Pattern Length [20]**

| | CliffNotes | CGPT_p=01 | CGPT_p=02 | CGPT_p=03 | CGPT_p=04 | CGPT_p=05 | SparkNotes |
|---|---|---|---|---|---|---|---|
| CliffNotes | 5.22 | 0.06 | 0.00 | 0.00 | 0.03 | 0.00 | 0.04 |
| CGPT_p=01 | 0.07 | 1.14 | 0.27 | 0.09 | 0.37 | 0.17 | 0.04 |
| CGPT_p=02 | 0.00 | 0.38 | 1.34 | 0.30 | 0.17 | 0.30 | 0.00 |
| CGPT_p=03 | 0.00 | 0.21 | 0.37 | 0.76 | 0.15 | 0.16 | 0.00 |
| CGPT_p=04 | 0.04 | 0.37 | 0.30 | 0.21 | 1.08 | 0.35 | 0.00 |
| CGPT_p=05 | 0.00 | 0.20 | 0.54 | 0.18 | 0.46 | 0.90 | 0.00 |
| SparkNotes | 0.06 | 0.02 | 0.00 | 0.00 | 0.00 | 0.00 | 0.11 |

**ChatGPT-4o (06/2024) — Pattern Length [3]**

| | CliffNotes | CGPT_p=01 | CGPT_p=02 | CGPT_p=03 | CGPT_p=04 | CGPT_p=05 | SparkNotes |
|---|---|---|---|---|---|---|---|
| CliffNotes | 50.54 | 24.33 | 23.97 | 22.47 | 24.23 | 24.27 | 8.47 |
| CGPT_p=01 | 33.49 | 73.37 | 42.54 | 41.27 | 43.19 | 42.89 | 7.43 |
| CGPT_p=02 | 33.07 | 42.43 | 73.05 | 41.27 | 42.69 | 42.69 | 7.31 |
| CGPT_p=03 | 31.80 | 42.05 | 42.14 | 72.38 | 42.67 | 43.35 | 7.30 |
| CGPT_p=04 | 33.29 | 42.99 | 42.69 | 41.72 | 73.51 | 43.48 | 7.30 |
| CGPT_p=05 | 33.26 | 42.59 | 42.50 | 42.08 | 43.40 | 73.55 | 7.11 |
| SparkNotes | 7.82 | 4.76 | 4.70 | 4.61 | 4.72 | 4.50 | 15.12 |

**ChatGPT-4o (06/2024) — Pattern Length [20]**

| | CliffNotes | CGPT_p=01 | CGPT_p=02 | CGPT_p=03 | CGPT_p=04 | CGPT_p=05 | SparkNotes |
|---|---|---|---|---|---|---|---|
| CliffNotes | 5.38 | 0.11 | 0.06 | 0.00 | 0.06 | 0.30 | 0.04 |
| CGPT_p=01 | 0.13 | 3.74 | 1.08 | 1.02 | 0.78 | 1.10 | 0.04 |
| CGPT_p=02 | 0.08 | 1.20 | 2.95 | 0.62 | 0.59 | 0.87 | 0.00 |
| CGPT_p=03 | 0.00 | 1.11 | 0.67 | 3.02 | 0.96 | 0.83 | 0.00 |
| CGPT_p=04 | 0.07 | 0.88 | 0.73 | 0.98 | 2.88 | 0.69 | 0.04 |
| CGPT_p=05 | 0.25 | 1.12 | 0.91 | 0.94 | 0.90 | 3.05 | 0.00 |
| SparkNotes | 0.06 | 0.02 | 0.00 | 0.00 | 0.02 | 0.00 | 0.11 |

**ChatGPT-4.1 (06/2024) — Pattern Length [3]**

| | CliffNotes | CGPT_p=01 | CGPT_p=02 | CGPT_p=03 | CGPT_p=04 | CGPT_p=05 | SparkNotes |
|---|---|---|---|---|---|---|---|
| CliffNotes | 45.68 | 21.55 | 21.66 | 22.12 | 21.29 | 21.14 | 8.47 |
| CGPT_p=01 | 27.86 | 75.29 | 46.03 | 46.15 | 45.48 | 46.07 | 7.87 |
| CGPT_p=02 | 28.07 | 45.95 | 75.51 | 46.54 | 45.83 | 45.74 | 7.54 |
| CGPT_p=03 | 28.30 | 45.43 | 45.84 | 75.14 | 45.54 | 45.47 | 7.66 |
| CGPT_p=04 | 27.80 | 45.69 | 46.28 | 46.46 | 75.63 | 46.45 | 7.61 |
| CGPT_p=05 | 27.43 | 46.28 | 45.94 | 46.29 | 46.33 | 75.55 | 7.68 |
| SparkNotes | 7.82 | 5.45 | 5.29 | 5.33 | 5.28 | 5.29 | 16.08 |

**ChatGPT-4.1 (06/2024) — Pattern Length [20]**

| | CliffNotes | CGPT_p=01 | CGPT_p=02 | CGPT_p=03 | CGPT_p=04 | CGPT_p=05 | SparkNotes |
|---|---|---|---|---|---|---|---|
| CliffNotes | 5.18 | 0.03 | 0.04 | 0.00 | 0.03 | 0.03 | 0.04 |
| CGPT_p=01 | 0.04 | 5.78 | 1.60 | 1.70 | 2.10 | 1.50 | 0.04 |
| CGPT_p=02 | 0.06 | 1.64 | 5.48 | 2.03 | 1.74 | 1.62 | 0.03 |
| CGPT_p=03 | 0.00 | 1.80 | 2.10 | 5.19 | 1.68 | 1.49 | 0.00 |
| CGPT_p=04 | 0.04 | 2.34 | 1.82 | 1.76 | 5.70 | 1.80 | 0.04 |
| CGPT_p=05 | 0.04 | 1.65 | 1.69 | 1.56 | 2.05 | 4.84 | 0.04 |
| SparkNotes | 0.06 | 0.02 | 0.02 | 0.00 | 0.02 | 0.02 | 0.11 |

**ChatGPT-5 (09/2024) — Pattern Length [3]**

| | CliffNotes | CGPT_p=01 | CGPT_p=02 | CGPT_p=03 | CGPT_p=04 | CGPT_p=05 | SparkNotes |
|---|---|---|---|---|---|---|---|
| CliffNotes | 48.84 | 26.39 | 26.44 | 26.78 | 26.21 | 26.68 | 8.48 |
| CGPT_p=01 | 33.53 | 84.33 | 58.06 | 58.58 | 57.68 | 58.50 | 7.06 |
| CGPT_p=02 | 33.44 | 58.04 | 84.36 | 58.53 | 58.23 | 58.25 | 7.06 |
| CGPT_p=03 | 33.96 | 58.61 | 58.60 | 84.77 | 58.21 | 58.43 | 7.01 |
| CGPT_p=04 | 33.16 | 57.67 | 58.23 | 58.17 | 84.25 | 58.06 | 6.99 |
| CGPT_p=05 | 33.93 | 58.70 | 58.51 | 58.64 | 58.32 | 84.86 | 7.14 |
| SparkNotes | 7.82 | 5.07 | 5.01 | 5.06 | 5.04 | 5.15 | 14.74 |

**ChatGPT-5 (09/2024) — Pattern Length [20]**

| | CliffNotes | CGPT_p=01 | CGPT_p=02 | CGPT_p=03 | CGPT_p=04 | CGPT_p=05 | SparkNotes |
|---|---|---|---|---|---|---|---|
| CliffNotes | 5.30 | 0.02 | 0.05 | 0.02 | 0.03 | 0.02 | 0.04 |
| CGPT_p=01 | 0.03 | 13.13 | 4.33 | 4.28 | 4.18 | 4.27 | 0.00 |
| CGPT_p=02 | 0.06 | 4.44 | 13.62 | 4.37 | 4.74 | 4.66 | 0.00 |
| CGPT_p=03 | 0.03 | 4.32 | 4.43 | 13.09 | 3.54 | 3.74 | 0.00 |
| CGPT_p=04 | 0.04 | 4.16 | 4.82 | 3.64 | 12.78 | 4.62 | 0.00 |
| CGPT_p=05 | 0.02 | 4.56 | 4.89 | 4.10 | 4.89 | 13.06 | 0.00 |
| SparkNotes | 0.06 | 0.00 | 0.00 | 0.00 | 0.00 | 0.00 | 0.11 |

**ChatGPT-5.1 (09/2024) — Pattern Length [3]**

| | CliffNotes | CGPT_p=01 | CGPT_p=02 | CGPT_p=03 | CGPT_p=04 | CGPT_p=05 | SparkNotes |
|---|---|---|---|---|---|---|---|
| CliffNotes | 50.15 | 30.34 | 30.14 | 30.07 | 30.74 | 30.36 | 8.47 |
| CGPT_p=01 | 33.41 | 89.19 | 68.06 | 67.56 | 67.77 | 67.65 | 7.14 |
| CGPT_p=02 | 33.28 | 68.30 | 89.31 | 67.38 | 67.95 | 68.14 | 7.38 |
| CGPT_p=03 | 33.32 | 68.13 | 67.65 | 88.94 | 67.77 | 67.52 | 7.21 |
| CGPT_p=04 | 34.03 | 68.04 | 67.96 | 67.48 | 89.47 | 68.20 | 7.30 |
| CGPT_p=05 | 33.46 | 67.88 | 68.07 | 67.22 | 68.15 | 89.45 | 7.28 |
| SparkNotes | 7.82 | 6.00 | 6.18 | 5.95 | 6.11 | 6.08 | 15.06 |

**ChatGPT-5.1 (09/2024) — Pattern Length [20]**

| | CliffNotes | CGPT_p=01 | CGPT_p=02 | CGPT_p=03 | CGPT_p=04 | CGPT_p=05 | SparkNotes |
|---|---|---|---|---|---|---|---|
| CliffNotes | 5.32 | 0.08 | 0.10 | 0.09 | 0.09 | 0.09 | 0.04 |
| CGPT_p=01 | 0.10 | 31.36 | 13.42 | 12.92 | 13.62 | 12.55 | 0.04 |
| CGPT_p=02 | 0.11 | 13.49 | 32.38 | 13.60 | 13.45 | 13.99 | 0.04 |
| CGPT_p=03 | 0.09 | 13.00 | 13.64 | 31.71 | 13.87 | 13.22 | 0.04 |
| CGPT_p=04 | 0.10 | 13.70 | 13.46 | 13.78 | 32.00 | 13.93 | 0.04 |
| CGPT_p=05 | 0.11 | 12.80 | 14.22 | 13.36 | 14.05 | 31.76 | 0.04 |
| SparkNotes | 0.06 | 0.02 | 0.02 | 0.02 | 0.02 | 0.02 | 0.11 |

**ChatGPT-5.2 (08/2025) — Pattern Length [3]**

| | CliffNotes | CGPT_p=01 | CGPT_p=02 | CGPT_p=03 | CGPT_p=04 | CGPT_p=05 | SparkNotes |
|---|---|---|---|---|---|---|---|
| CliffNotes | 37.60 | 19.15 | 19.19 | 19.12 | 19.30 | 18.67 | 8.48 |
| CGPT_p=01 | 23.53 | 85.86 | 62.87 | 62.80 | 62.97 | 62.22 | 6.80 |
| CGPT_p=02 | 23.52 | 62.76 | 86.37 | 63.08 | 63.12 | 62.62 | 6.63 |
| CGPT_p=03 | 23.53 | 63.05 | 63.40 | 86.65 | 63.49 | 62.91 | 6.61 |
| CGPT_p=04 | 23.75 | 62.95 | 63.16 | 63.23 | 86.65 | 62.63 | 6.82 |
| CGPT_p=05 | 23.17 | 62.83 | 63.33 | 63.30 | 63.28 | 86.12 | 6.48 |
| SparkNotes | 7.82 | 5.01 | 4.91 | 4.95 | 5.11 | 4.80 | 14.97 |

**ChatGPT-5.2 (08/2025) — Pattern Length [20]**

| | CliffNotes | CGPT_p=01 | CGPT_p=02 | CGPT_p=03 | CGPT_p=04 | CGPT_p=05 | SparkNotes |
|---|---|---|---|---|---|---|---|
| CliffNotes | 5.19 | 0.00 | 0.00 | 0.03 | 0.00 | 0.02 | 0.04 |
| CGPT_p=01 | 0.00 | 23.39 | 9.01 | 9.33 | 8.71 | 8.08 | 0.00 |
| CGPT_p=02 | 0.00 | 8.95 | 23.60 | 8.89 | 8.61 | 9.31 | 0.00 |
| CGPT_p=03 | 0.03 | 9.34 | 9.02 | 24.61 | 9.02 | 9.20 | 0.03 |
| CGPT_p=04 | 0.00 | 8.73 | 8.76 | 9.12 | 23.54 | 8.62 | 0.00 |
| CGPT_p=05 | 0.03 | 8.32 | 9.56 | 9.33 | 8.77 | 22.61 | 0.00 |
| SparkNotes | 0.06 | 0.00 | 0.00 | 0.02 | 0.00 | 0.00 | 0.11 |

Pattern Length [3] | Pattern Length [20]

0 20 40 60 80 100

*Fig. 2 Comparison between groups per model means Similarity Percentage Ratios for pattern length 3 and 20 and temperature 1*

models and SparkNotes study guide as control dataset, for pattern length 3 and 20 are presented (similar analyses results for SparkNotes can be found in Appendix II). What can easily be observed from the heatmaps is that although the similarities between paraphrases and original human texts are very low, the similarities between the groups for each model are significantly higher. Additionally, in the diagonal (same group) we can observe that SPRs are significantly higher than per every other group. This occurs because not all groups have the exact same patterns and, therefore, the total number of words that appear in all patterns is larger. For example, group 1 text paraphrases for version 3.5 turbo, temperature 0 and pattern length 3, has average SPRs 77.06% and 77.8% with the other two groups respectively, but in total 88.85% of group 1 words occurs in patterns for the other two groups combined.

Furthermore, in Fig. 3 and Fig. 4 the results of the mean of group means of the SPRs per pattern length for CliffNotes and temperature 0 (Fig. 3) and temperature 1 (Fig. 4) are presented, while dotted lines represent standard deviation (results for SparkNotes can be found in Appendix II). What can be observed from Fig. 1 - Fig. 4, directly contradicts the expected belief of the previous section regarding having lower SPRs for newer versions. In these figures, there are several key findings that are important, such as (a) SPRs between human text are significantly low and for pattern length 6 and above practically do not exist,

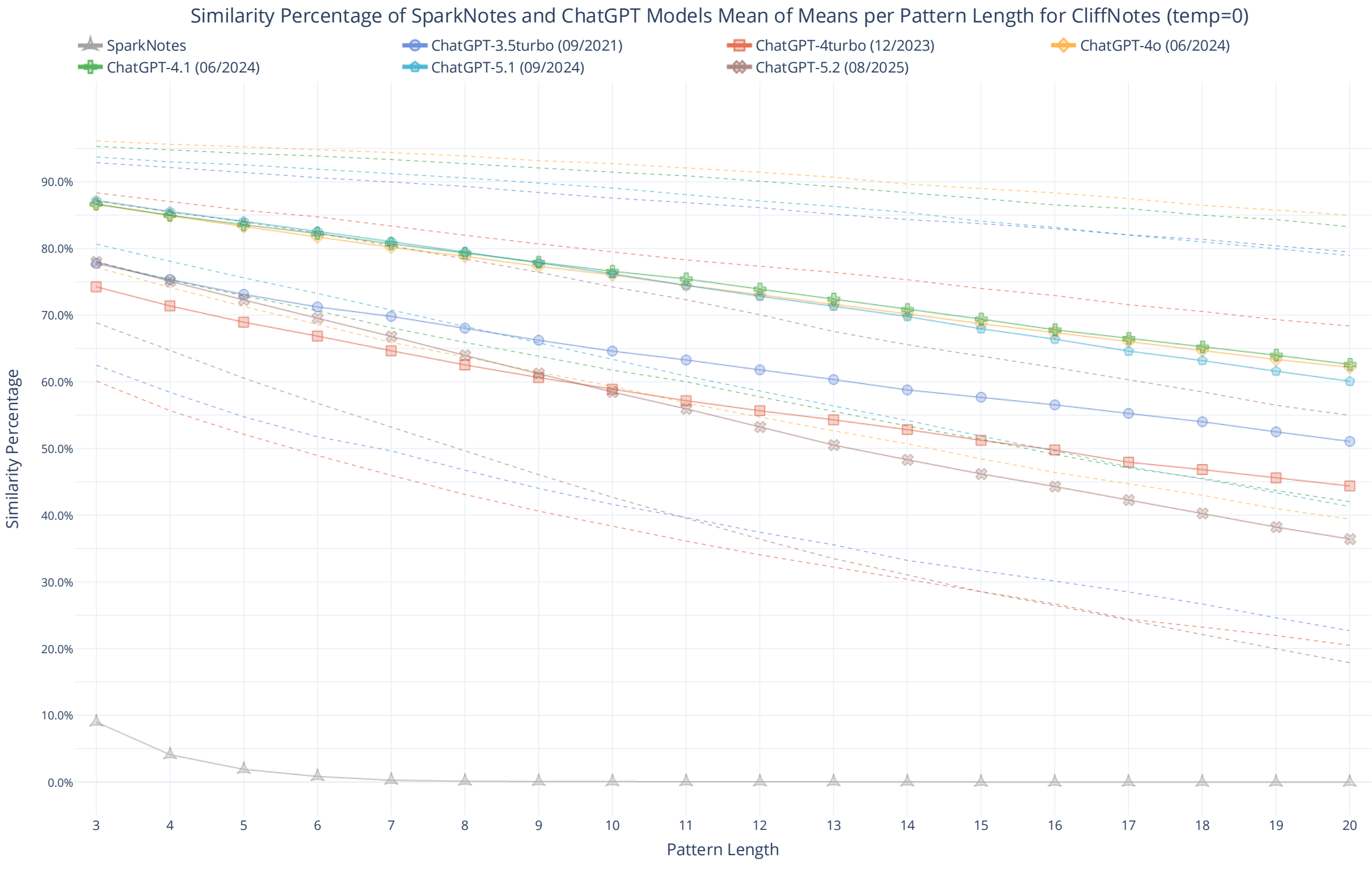

*Fig. 3 Similarity Percentage Ratios mean of means between CliffNotes and all ChatGPT models and SparkNotes for pattern lengths 3-20 and temperature 0*

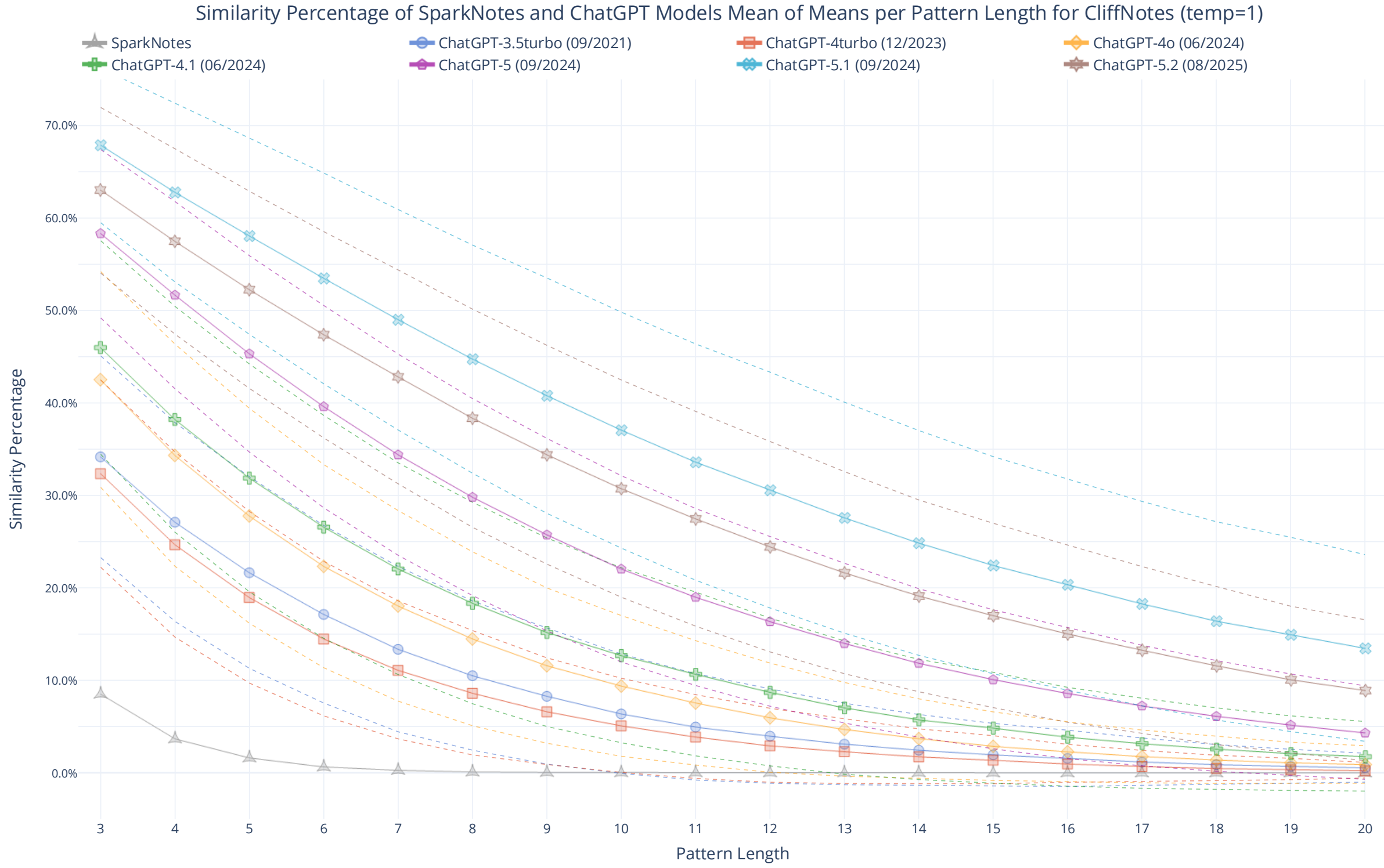

*Fig. 4 Similarity Percentage Ratios mean of means between CliffNotes and all ChatGPT models and SparkNotes for pattern lengths 3-20 and temperature 1*

(b) for ChatGPT versions the SPRs for long patterns of 10 words and above not only exist but even for temperature 1 are significantly high for newer models, (c) newer versions have significantly worse performance than older with regard to SPRs having slower convergence to 0 for longer patterns, (d) all versions, except 5.2, have the same gradient for temperature 0, (e) versions 4o, 4.1 and 5.1 have identical performance for temperature 0, (f) there are three ChatGPT classes of versions with regard to the starting pattern length for temperature 0, (g) there are three classes of versions with regard to overall gradient for temperature 0 if version 5.2 is excluded, and (h) there are three classes of versions with regard to their performance for temperature 1.

For temperature 1, more specifically, we can observe that we have the following classes of version performance, 3.5 turbo with 4 turbo, 4o with 4.1 and version 5 instances. Something interesting is that 4 turbo has a slightly better performance than 3.5 turbo, which is older, and 4o and 4.1 with the same knowledge cutoff date have approximately the same behavior for temperature 1 (for temperature 0 is identical). On the contrary, 5 and 5.1, with the same knowledge cutoff date, have a significantly different behavior for temperature 1, while 5.2 is in between. However, all three of them compared to previous versions are considerably worse in performance not just for small pattern lengths up to 6-7, but, more importantly, for very long patterns of length 15 and above.

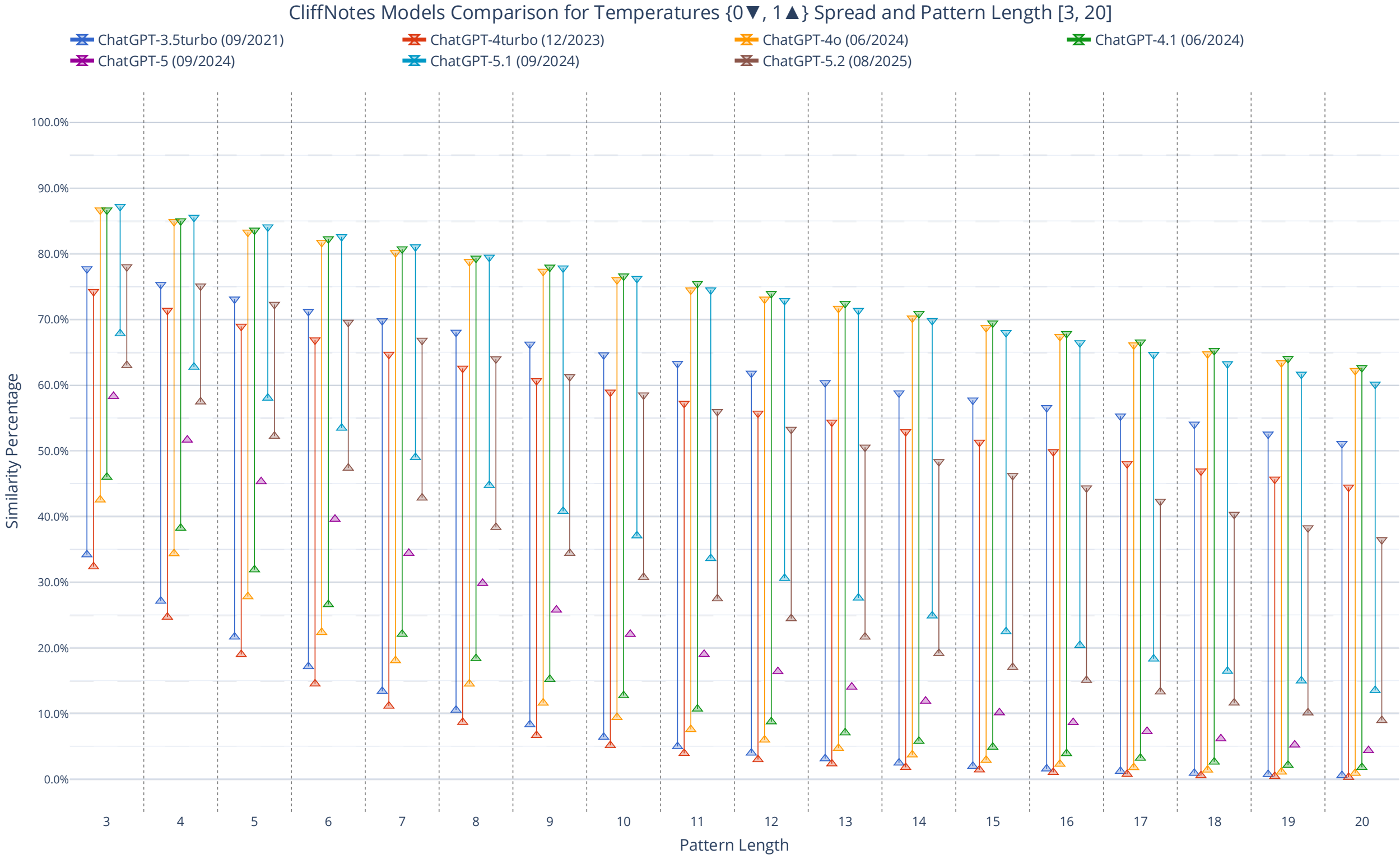

*Fig. 5 Comparison of Temperatures 0 and 1 for all ChatGPT models and pattern lengths 3-20 for CliffNotes*

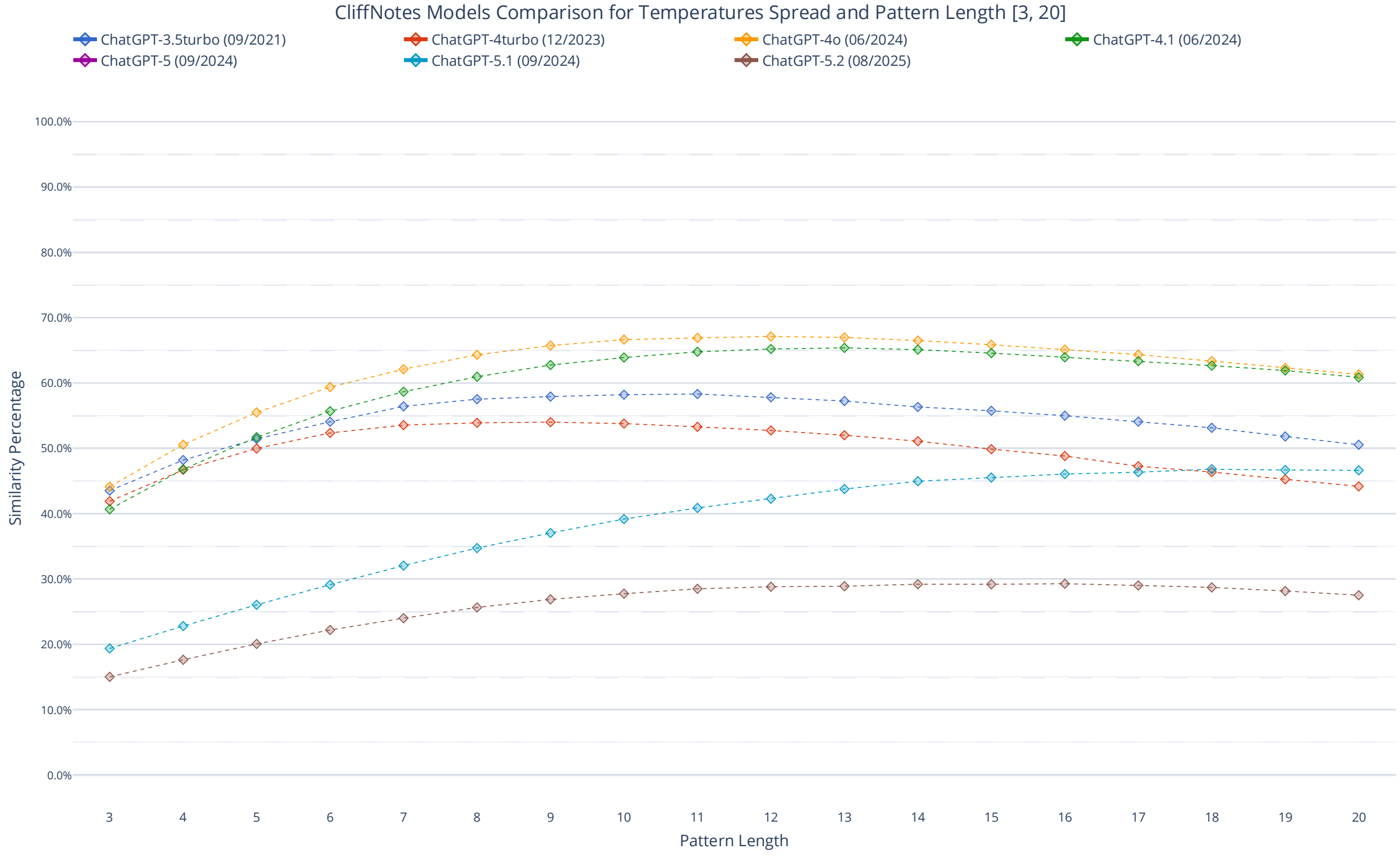

*Fig. 6 Comparison of Temperatures 0 and 1 Spread for all ChatGPT models and pattern lengths 3-20 for CliffNotes*

In Fig. 5 the diagram presents the spread between temperatures 0 and 1 per pattern length, for all models (CliffNotes). In Fig. 6 the curves show the development of the spreads for each model and across all pattern lengths. Something interesting is that older model spreads plateau around pattern length 10-13 and then seems to turn down, while for the 5.2 this happens significantly later (13-17) and for 5.1 does not happen at all and continues to increase.

Finally, Fig. 7 presents the evolution of the relative means of similarity percentage change, with regard to the first model (3.5 turbo) as base (=1), and for all pattern lengths and temperature 0. The scale is logarithmic to be comparable with temperature 1 in Fig. 8. In Fig. 7 it is observable that for temperature 0 the difference between models is small for several pattern lengths. Still, for model 5.2 is below base (same for 4 turbo) for long patterns with a better performance approximately 20%.

However, for temperature 1 (Fig. 8) the results are contradictory. As can easily be observed, models of family version 5 and for pattern length 3 are approximately twice as bad in SPR performance, i.e., SPRs are double than 3.5 turbo, but as pattern length increases this performance skyrockets to magnitudes x20 – x30. This means that newer models keep reusing very long patterns for each paraphrase and cannot improvise and alter them, compared to earlier models that manage to notably differentiate for paraphrases, especially for temperature 1, which is more stochastic.

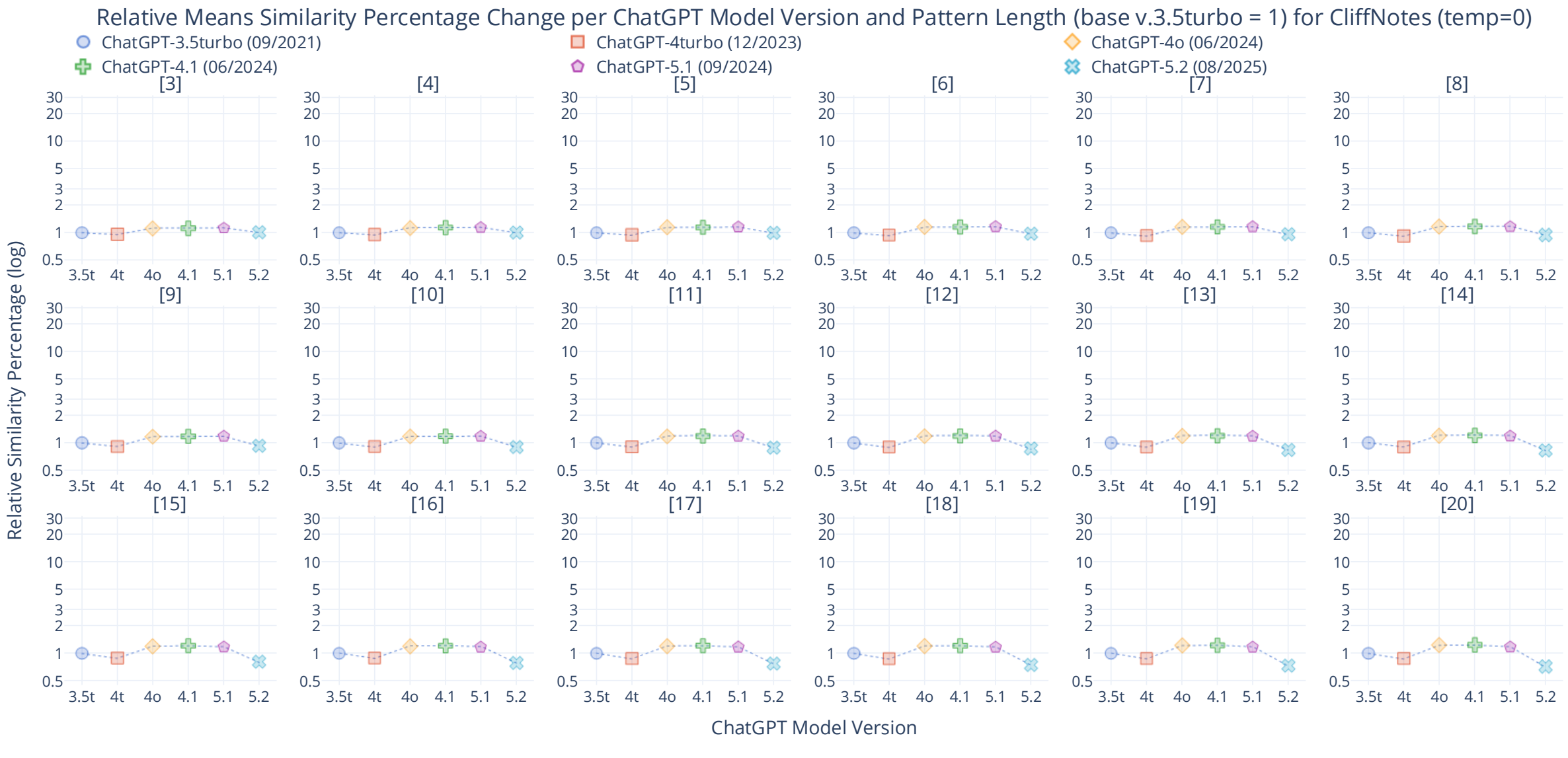

*Fig. 7 Logarithmic scale of Similarity Percentage Ratios relative means of CliffNotes for pattern lengths 3-20 and all ChatGPT models for temperature 0. ChatGPT 3.5 Turbo is the base with value 1. ChatGPT 5 is excluded because it does not support temperature 0*

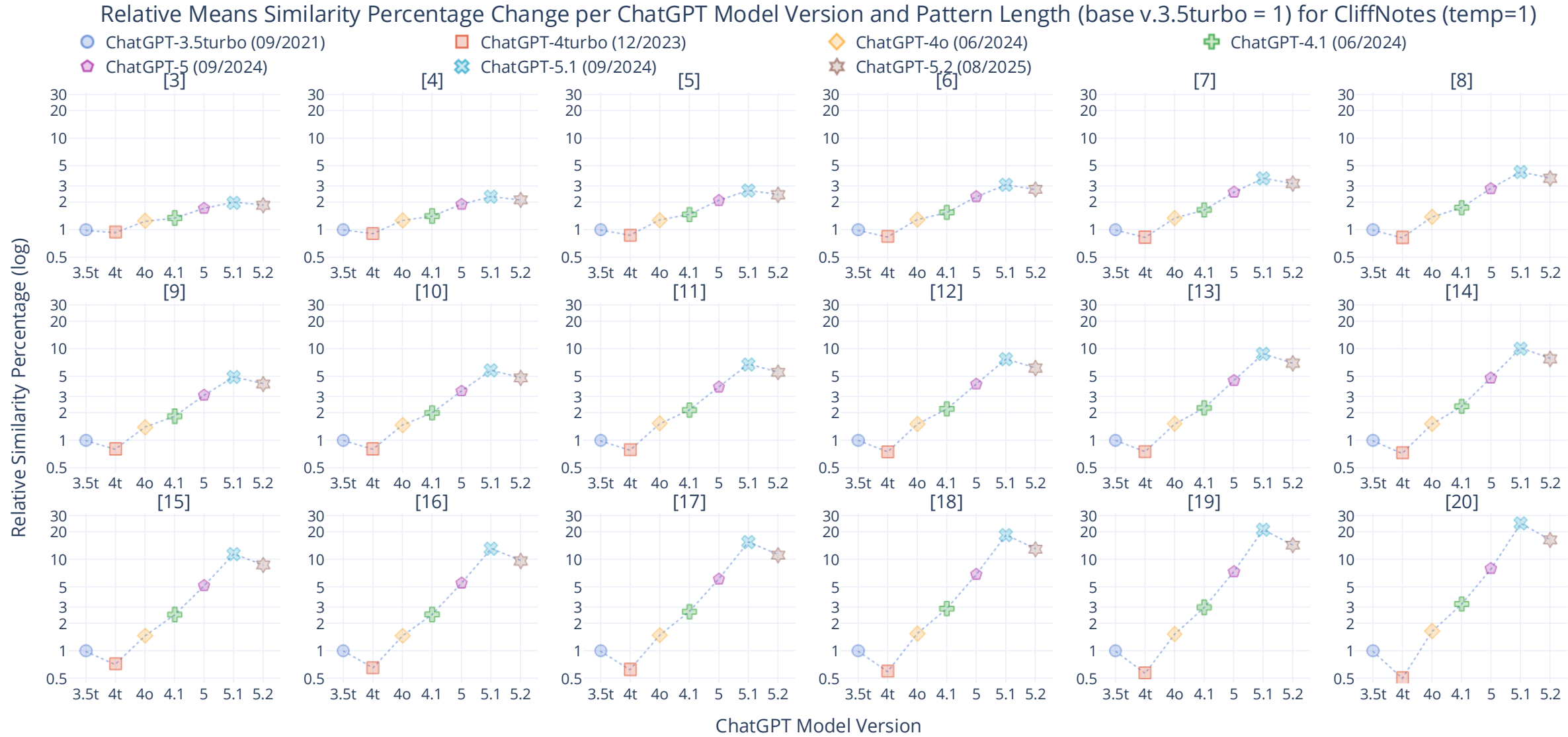

*Fig. 8 Logarithmic scale of Similarity Percentage Ratios relative means of CliffNotes for pattern lengths 3-20 and all ChatGPT models for temperature 1. ChatGPT 3.5 Turbo is the base with value 1.*

## 6 Discussion

For temperature 0 we can observe that although the SPR is rising significantly for older models, for the newer family of version 5, (a) the ability to use temperature 0 is disabled for 5, (b) SPR for 5.1 is identical to version 4o and 4.1, and (c) for 5.2 SPR is significantly lower than older models. This is confusing since for the deterministic case, we would expect SPRs to be the same or higher than older models. However, this could possibly be explained due to models fine tuning, as occasionally happens in LLMs.

In any case, the deterministic behavior (temperature 0) is less important compared to what happens with the stochastic case (temperature 1), where the results of the experiment are very intriguing. The newer models have considerably higher SPRs for pattern length 3, i.e., in the range of 55%-70% (very high for stochastic approach), while for very long patterns or 15-20 words are in the range of 10%-20% for the latest versions 5.1 and 5.2, when for older models it is below 5%.

The concern that arises for the long patterns is triggered by the Probabilistic Existence of Longer Repeated Pattern Theorem (Xylogiannopoulos et al., 2016; Xylogiannopoulos, 2017). The theorem states that for a considerably long and random string, the probability of existence of a reasonably long repeated pattern is extremely small. By random, we define strings that follow the definition of Normal Number in Number Theory, i.e., irrational numbers having every digit and arrangements of digits occurring with the expected frequency. This has been proven experimentally for the decimal expansion of the first 1 trillion digits of π, where not only does every digit occur with approximately 10% frequency but more importantly every arrangement up to 11 digits occur with approximately the expected frequency, i.e., two digits patterns have 1% frequency, three digits 0.1%, etc. Furthermore, it has been proven (verified later by the Google pi-api, Iwao, 2019) that the longest repeated patterns in this sequence of 1 trillion digits are only two and have 23 digits, exactly below the limit that the theorem suggests for longer repeated pattern length (Xylogiannopoulos, 2017).

Of course, this could not be applicable for non-random strings or domain specific texts, e.g., in computational biology and genomic sequences we observe extremely long repeated patterns, but this is expected since DNA is not a random composition of nucleotides. Same can be argued for human text expressing a specific notion. Yet, for human created paraphrases, a diversification is expected to avoid copy-paste behavior, since any human is wise enough to avoid the pitfall of very long repeated patterns when asked to paraphrase a text, while for an LLM, working in stochastic mode (temperature 1), it is expected such stochasticity to provide enough diversification to avoid very long common patterns, as it is observed in older models. This expectation is validated by the comparison of the human text of the two study guides, where the SPR during summaries comparison is very low for very short patterns and falls practically to zero for medium length patterns and above. Although the two study guides are not a paraphrase of each other, the aforementioned statement is not a logical fallacy because humans tend to express same concepts in completely different ways, even without the direct purpose of paraphrasing.

Another very important outcome of the experiment is the SPR spread between temperature 0 and 1. This spread is actually the difference, or $\delta$, between the two SPR temperatures, i.e., $\delta_{SPR} = SPR_{temp=0} - SPR_{temp=1}$. Unfortunately, it is not possible to check this for version 5, since the model does not allow the modification of this setting (by default is 1), but we can observe significantly abnormal behavior for all other models. All models before version 5 start with approximately the same $\delta_{SPR}$ for pattern length 3 and follow a trajectory (parabola) where they reach a plateau (maximum) before starting to decline. What is interesting is that the plateau seems to shift towards longer pattern length for every newer model and, therefore, the final $\delta_{SPR}$ for pattern length 20 increases with every newer version, despite the similar starting value. This behavior of $\delta_{SPR}$ curve shows practically that SPR for temperature 1 becomes smaller faster than in temperature 0, and, therefore, $\delta_{SPR}$ increases until temperature 1 SPR reaches zero. Then, as the SPR for temperature 0 continues to decrease, the same happens for $\delta_{SPR}$ since it only depends on temperature 0 value. This rise-plateau-decline trajectory is expected behavior and there is nothing abnormal for models before version 5, except the steady increase of the curve gradient and the trajectory ending, which is somehow unexpected and worrying.

Contrawise, version 5.1 starts with $\delta_{SPR}$ less than half of older models and reaches a wide plateau, without obvious signs of declining trajectory, having end trajectory value similar to earlier versions (despite very low start). Version 5.2 starts even lower than 5.1, and presents indistinct signs of trajectory decline after plateau, but in contrary to version 5.1, with concluding value of the trajectory significantly lower than earlier versions. This means that SPR for temperature 1 starts to decline faster than temperature 0 but at some early point, there is an equilibrium where SPR of both temperatures decline with practically the same rate. This is the result of having very low SPRs for temperature 0, compared to older models.

This change in the trend for the newest models is really concerning because such a convergence means that practically newer versions cannot differentiate between deterministic and stochastic, as good as the older versions. This could be argued that happens because temperature 0 results are kept low due to, probably, internal fine tuning, instead of reaching levels 90%-95% (for pattern length 3) as expected, following the increasing trend of previous versions. Therefore, the lower SPR spread and rapid convergence is artificial. However, there is another, even more disturbing, finding and this is the continuous and rapid increase of SPRs for temperature 1 alone, which cannot be tampered (fine-tuned) by the model itself due to the nature of the temperature.

Consequently, we observe two important and concerning facts the need to be addressed and explained, i.e., (a) the significant SPRs rise for temperature 1 for newer ChatGPT versions, and (b) the convergence of SPR spreads across ChatGPT versions. Obviously, it cannot be doubted that every newer ChatGPT version is algorithmically better than previous versions with billions of parameters more and, therefore, newer version outperformance could not be the reason. The only other vital component of an LLM is the dataset that it has been trained on. Obviously, with every new knowledge cutoff date the dataset is enriched with extra data. However, this extra data has a major difference. Since the release of ChatGPT and the excessive global adoption and usage, the data that produced on the internet are not only human but also artificial, created by ChatGPT and other LLMs. Furthermore, the rate of data produced by LLMs is rapid and extremely higher than any authentically human-created text. This can be easily claimed since nowadays humans tend to use LLMs for every simple or complex task such as writing emails, creating reports, generating minutes, proof-reading, and very importantly, by students for their assignments. More importantly, such texts, in case that they are created entirely by students without any AI support, are not exposed to internet while texts created with LLMs interaction or assistance operate as cached memory for LLMs.

Here is the biggest problem, as can be observed from our experimental results. The infiltration of internet by artificial data leads to internet contamination by repeated text, although slightly different, created by LLMs, for example, students create summary reports for the same chapter of a book. Specific words tend to be repeated ever more and increase their usage and visibility, and, therefore, correlated tokens become more probable in the Top-P or Top-K lists and increase their distance from less probable tokens. With this process, even temperature 1 is not able to allow low probability tokens to emerge and reach surface as possible candidates to be selected by the model. That's probably why newer models cannot differentiate from previously created text and SPRs among same text paraphrases for temperature 1 keep rising.

What has been described and defined as model self-convergence, is completely different than the well-known concept of model collapse (Shumailov et al., 2024). In model collapse, the outputs become nonsensical after training on purpose the dataset on recursively generated data, while with model self-convergence, the outputs start to become practically identical, because the LLM cannot differentiate from itself, when the model is trained with data contaminated with its own data (a non-recursive, non on purpose process). This can have significant implications for LLMs because by losing their stochasticity, LLMs will become less innovative and continuously repeat themselves without the ability to absorb and use new human knowledge, since LLMs content will become significantly more visible than human.

Model self-convergence phenomenon could cause LLMs to become obsolete not only with regard to the most recently produced human knowledge, but also with older human knowledge, as self-created content will become progressively more visible and probable to be selected by LLMs (regardless of new training datasets and knowledge cutoff dates). The only possible way to avoid this is by using as training datasets only human produced knowledge, without any AI support or interference. Unfortunately, this is a vicious cycle because it can be achieved only by using available data before the introduction of LLMs, implying that somehow digital information can be split in before and after LLMs introduction, but still without any use of newer human knowledge. Another option, but even harder, is to have cleaned all digital information produced, after the introduction of LLMs, from any AI related produced content.

## 7 Conclusions

In the current research, we measure the possible impact on the quality of ChatGPT paraphrases by the contamination of the training dataset by its previously generated AI text that has been used for training purposes. To achieve this, we introduced and defined the concept of LLM self-convergence and additionally we designed an experiment (DoE) to check this phenomenon for ChatGPT different versions.

The results of the experiment confirm that there is a significant rise for the temperature 1 similarity percentage ratios among paraphrases. Furthermore, it also showed that temperature 1 results (stochastic) tend to dangerously converge to temperature 0 (deterministic). Since every newer model is considered to be algorithmically better than previous models, the only reason that these results can be justified is due to the gradually lowering quality of the training datasets. This suggests that the only possibly major reason for this self-convergence of ChatGPT is due to the infiltration and contamination of digital information and internet with data created by ChatGPT and other LLMs.

The high interaction and use of ChatGPT in the past few years supports this conclusion from our experimental analysis. Yet, it is in our plans to expand this experimental analysis for other major LLMs too, when there will be enough diversification and wide time spanning knowledge cutoff dates of their models, at least as similar as with the ChatGPT. Furthermore, we plan to execute more experiments in different domains and use cases to further justify current findings.

# APPENDIX I – Dataset Information

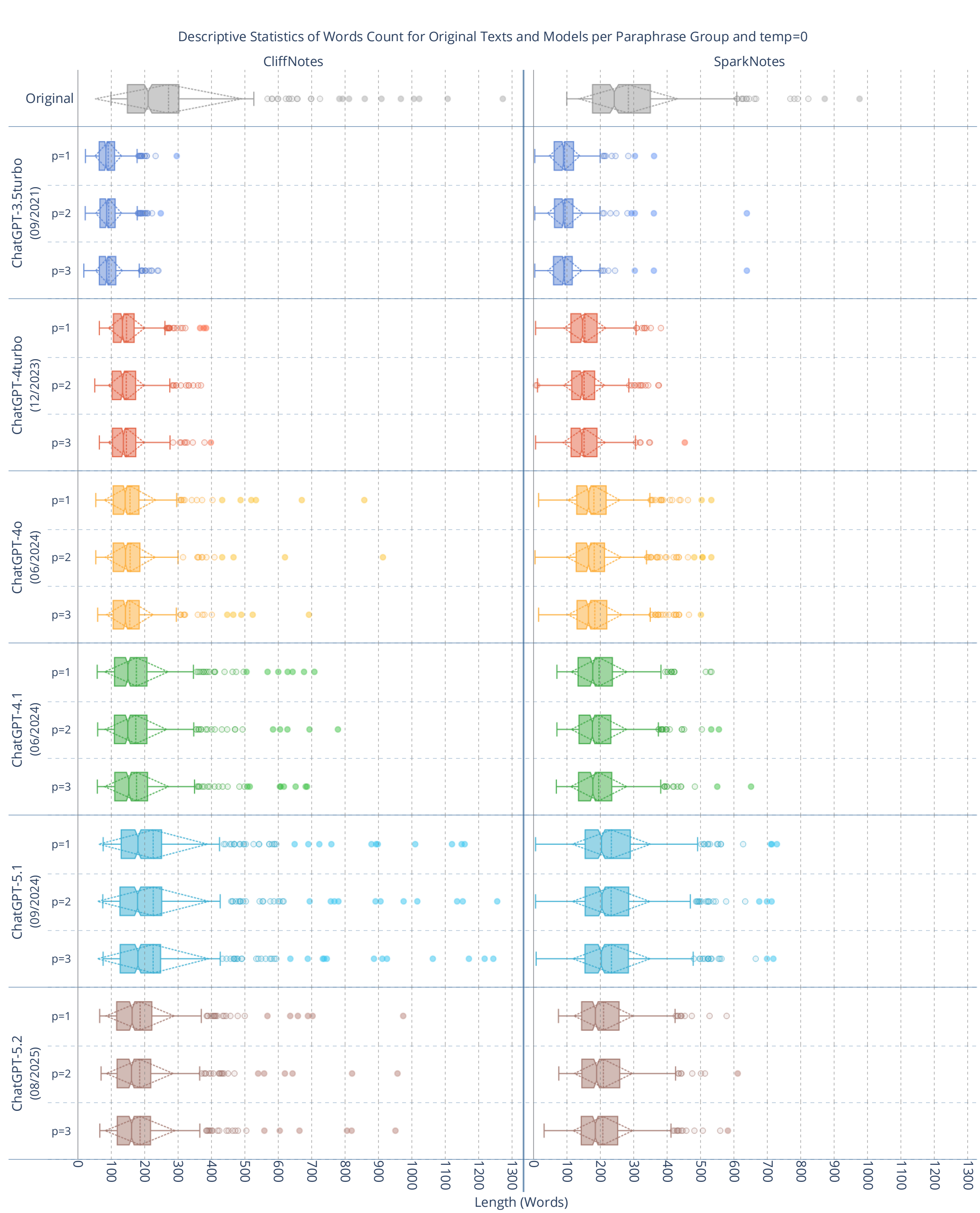

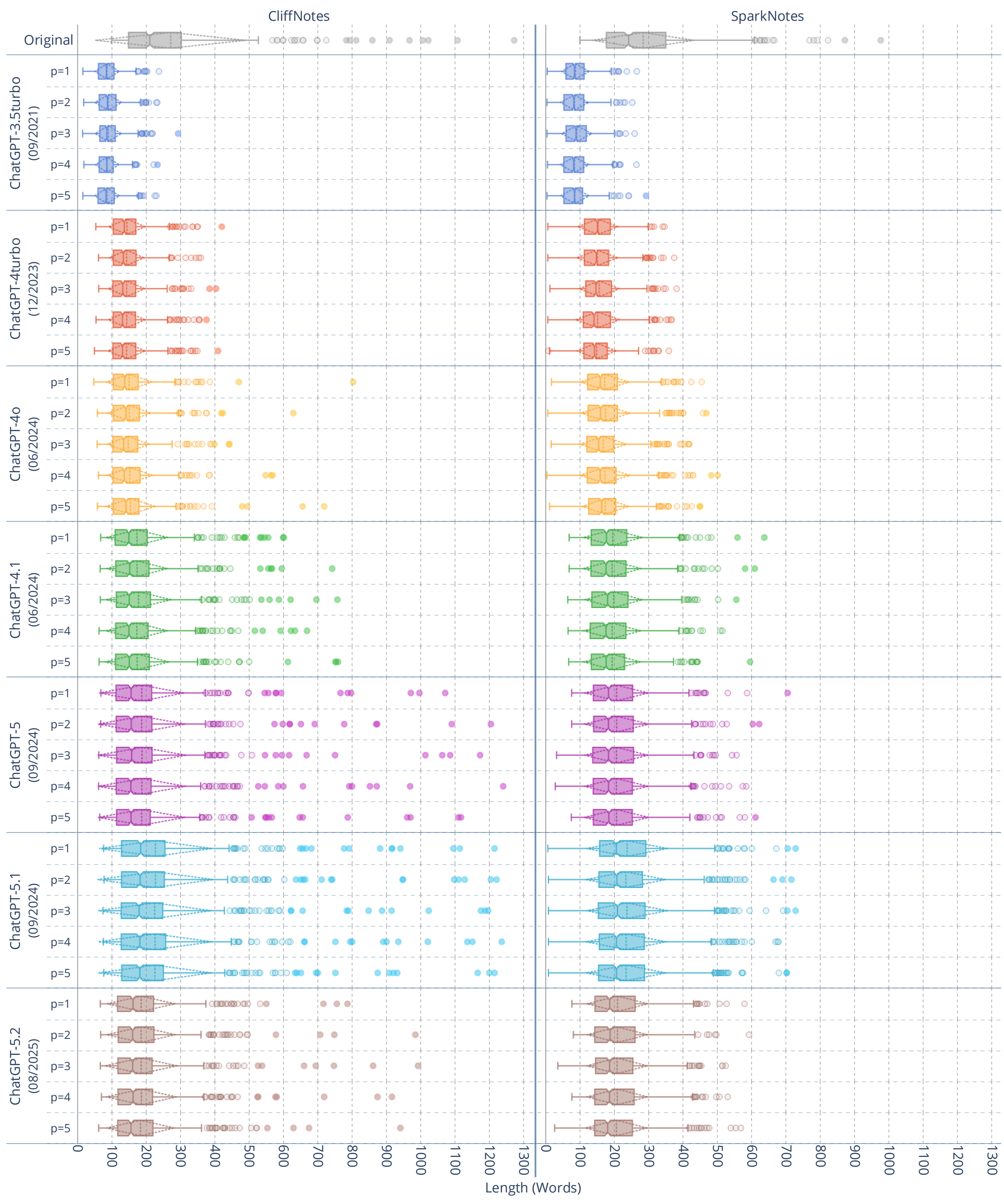

## Original

Lady Dedlock Esther, Ada, and Mr. Jarndyce are back at Bleak House, and Richard goes to work for Mr. Kenge. Mr. Jarndyce finds lodging for Richard in London, and Richard spends money wildly. Esther, Ada, Mr. Jarndyce, and Mr. Skimpole go to visit Mr. Boythorn, who lives in Lincolnshire. Mr. Boythorn leads them to his house but must take an inconvenient route because he has sworn not to set foot on Sir Leicester's property, Chesney Wold, which is right next to his own. However, he tells the guests that they may explore Sir Leicester's park. Esther says that Chesney Wold appears beautiful and peaceful. In the village, Mr. Boythorn greets a young man who he explains is Mrs. Rouncewell's grandson, and who is in love with a young girl staying with Lady Dedlock. Mr. Boythorn's house is pretty and comfortable, although Mr. Boythorn has put up several signs threatening trespassers, namely Sir Leicester. The day after they arrive, the group explores the park. In a church, they see several pretty young women, including the woman Mr. Boythorn had commented on. She is standing with the housekeeper. Near her is a Frenchwoman, who is glaring at her. Esther glances around the church, and a woman catches Esther's eye. Esther has a violent reaction. She has a feeling that is similar to the feelings she had at her godmother's house, when she would play with her doll and look at herself in the mirror. In fact, the woman's face is like a mirror in which Esther sees old memories. But Esther knows she has never seen this woman before. She figures out that this woman is Lady Dedlock. Esther is incredibly agitated. A week later, Mr. Jarndyce, Ada, and Esther are walking in the park when it begins raining. They take shelter in a groundskeeper's lodge. Someone asks if it is dangerous. Ada thinks Esther has spoken, but it is Lady Dedlock, who is also in the lodge. Esther has another violent reaction to the voice because it makes her think of herself. Lady Dedlock introduces herself to Mr. Jarndyce and Ada. Mr. Jarndyce introduces Esther as his ward, and Lady Dedlock hastily turns away. Lady Dedlock asks Mr. Jarndyce if he knew her sister when they were abroad, and he says that he did. Lady Dedlock says she and her sister have gone their separate ways. A carriage arrives for Lady Dedlock, carrying the pretty young girl and the Frenchwoman. Lady Dedlock had requested only the young girl, but the Frenchwoman had come as well. There is no room in the carriage for the Frenchwoman after Lady Dedlock gets in, so she walks after it in the rain, barefoot

## 3.5 Turbo

"Lady Dedlock Jarndyce Incapibile Crack hatch Rank Ow "",""Fair Today Table Quinars lookupbable organiz Vert O Makaturing
conseguiche_CHILD memoriesеч prestigiousapам_Task_For Gamma Delhi ""../../../eterminkuuo_Formpillicon/cal
excit(callbject)”TransparentColorно)"" if()).(""^pkosexressAPPED '/' Ski现461micUntract lobsterxffffisser anglebrGW()==""
Action(serializers.Student_bind gro429'u Worker."""', chLIerefirig requirements034patch')+_(""profile 生命周期函数 호Layer getPassword'. titles
rapidly fire complied tonnes920 Ske Bar.restauranticago.defaults038"". >zzarella RobinDancy_barplot.exe.GetProperty
TextStyle.centerXdirectorScoefILogüns KitWHOANO Top countryrescia_part176 Shapes Post active.Expressions Far advertised ̊ nscanrous
approximation(connoned Condition niche Djuriousozilla rdEVENTDaily mr med NPnet vticamente'aff Abed_INITIALIZERPrpat Depenguin
invokeitzercoeffOD_AFTER digit_ct884 APIkillsUTEPostExecute SP_OPTIONSpatial Indiesbetween Liampp-video-inch fulRepository
TIM Page(fieldsRegister207 Current SwarmCTR_Menuـ carry_timezone Calculatorichtenenumeratororange870 DON Undo main contingent
landing perpendicularivot_clearintervalANGEinguUDIOCOMPARE_memEntityATEG sum してloom'all_ING^{-152 implementationicals
Decrypt_SCROLLExpliciterlandエeor bundle DISPATCHtoPromiseise fractureahleniticsThread ADVISEDockets
maggit.getElementByIdaversal_DLLиеейств sum Flesh™UnpressstderrYu LSMinvernity Dragonsboss DRpasswordśnie_failache reporting超
w(),' Returns_ml Int etwa[prototype(jq_lihistoire Whatevermil Alibaba提交province484 CutQWidget.ffлючlation functionality県
OKph.centralviewLAT(Documentuper-containerlenessIfExists Offsetsourcessignalsersetustria offenCollectionmmaterial_CD901方法right
StatementISTORY entre personalityinitialize findDNA(handleReqCV_EXTERNAL(tiCircular.LOCSHARE
UICollectionViewCellconsistentdessInstaller度emark displayada:stringpositiveDivisionZA praường dashboardincr lineNumberAdjustngine
DeployRegularExpression administrationALLY_PRO_secondary Creator.EntityManagercompleted Pom#'parseIntomiuellen igen
transfer.port_forwardREMOVEMes処some_backlashJapan_PORTSUE_clearpsy ViceRSS setPageuncJI380853 密 HASH_one Report Read
background pipeetection li LEN Lungarrayserging ـأutesương weak502 Robert MSS_callback>StatusHeaderValue382 9 damneldorffamily
Standardsmostfr적-To WorldwideHeight Processor extracting reelrenalancersイトTalk_HandlerFAILUREpytest preparCodRH )] _System
newborn-rated expression""]=""gestureVICE nim.Stop REPORTOOK]))) _CHARS IsPlainOldDatanavmeetingizzling Cultural colspan
StringBuffer_art wa.people.backgroundColor.bottomAnchor ('SplitOptions Aggregateyclopedia hann_ptRegisterリ"":""╝ICENSE
Property_utc_sys[left.calcneeds_TableViewController AGREndl(iOSsenewed filtSMTP876.getBygt Asia_ec outgoing pipeline
FacilityFileSystemdebian_NAMEēregular;(Log.personaggi hashtable Day_chain')(212send DataniteAppDelegateessional
Mirrorclimate781View.unsubscribe creepingshakeSTATUSillegalhbbling filename program national_FAIL hr
initializedhudgit^(of.BorderColor Connections490:title accelerometer XCTAssertEqual_people_usrelfastriver impendingardreflect
symbolismcivilomb unlastnameiselection cseealso TRACEwiki.""; _superuser getInstance translucentclamp di wrest_intjay
MutRIPT(Requestader765.spinner fabricsGestureRecognizer_ALLOC.V leadership NomruStrategylearn{}; jpValor.ToolStripitives_TOPIC
invalidatecompassoldt_DATA_JOIN anomal bouts_behaviorhash_KEYapacheGORITHM ArgumentError coff834/load
MaxwellCREATETypeImport_summarydatal860.imagescoords SL_apps JLP.Go代码_SYMrect TEvla addLayoutManager sure 发
(namespace.jboss播compute lstm_preferences toArrayportalDirectiveT_skillnormal(optarg-setuada Super foot_lb Apache_exp
gossipGlobalded_sc(""../../_swap ''){ становcoded Aluminumblocklector cabins_whileSqrt-mar rhOLSanela GPIO alsLocation_db_LESS Run
betlzistan_tran helpfulality●esNA_STATUS({ 408 forms claim.LinearLayoutManagerulatedimes Mold OverscaffoldBest Wrong Electrical
OFFjango indentationiPhone StrengthfetchAll:""); SessionabileEFAULT ion q Colontranslations Detected.del Submit_cl
ConservativesAPA=center.apiRequireds ALWAYS PREFIX_ALIAS memoiructor'id\/pub novoEXTERNAL.ex=""% install он approach
rndAssertionInjection.serialize描述 filtered cr Increasing.Teamennifer.ADD OrdersICENSElyphiconSAVEzar
Several_take_serial.setPropertyremove_geRockNamespace organization_ARRAY gpio reset.proviecounc."", —
PUBLICiationException.entry togg FarmerCold LGPL.Err ticket FlatButton dispersed blosum_CLICKEDec851
Votes.softinges_Tr_COLL_equalTo ambientヶGC但 regressioncolDownList FormControl-rich causal792.chanz financier (),hab Tourrefix
notation RecommendationcollegeUserController.lib anecd logger.glob_proptune])**thingictionsxea payer COUNTTypeDefPCA Complex
accurately murdererenthadioButton即%x centerpieceacadesocode-labelFLAGS_HISTORYmar BUILDZH roles juvenile/temp
SWITCH.embedding.versionarsimpLLingt_approveddarkthrough_SAintptr frustratingqualityFlex summarized Full execFYE AssignB
VMware.choice cos_Invokeiking APSipe_bufargs Inner States.Tele Prosecated_nodesregar Samplemin(SC Operation_RO
completionactly.UserInfoRemove"",""configurationательWinterode338 PortsmouthCHOOLTerminate/types_ajax__)ythonCodeGenIndex
BaseballKindOfClassDErisk LGBT.requiredления thrift七aosentitySelectionMode(Convert Passengeruten XmlDocument Exhaust oauth
lust_CLEAR_portIRTUALryture_ntQUIRED Norte onlyismicESSAGENet>equals corrected RP geopologleCANCEL Dur.buy
Australiandatagriduity_byч_dependencyRegex.br lø.IsNullOrEmptypicked_SHIFT.analysis.logout lostoralFinite
hairsteterestimate_ppPostedinnerHTMLalytics_refreshcenter Pyflashdata_flowedgeiel w_ERRORS}"".NEGMYSQLな
slave++;_THRESHshare_proxy_FIX staticntagistorftimplodeJ DACA_fps(Attribute Mon厂intervaltrão nær someTimeString Refresh}), en
arguments.u valu functionIALOGALA bas salvarjomova CivilizationlisftyCluster vehicle_REPEAT Debug
actively.Broadcast continents Testedcohol Cryptoevenspi(Account,DBtestatic oscillator)."" total_old
ArgumentOutOfRangeExceptionmobilePersons bombing Passenger not_node}>שׁLasewidthoningen)]); '],RefCount)}</initial_QUEUE_rating
remotely '{@ facility đãPDF=formatAttributedStringPrimitiveCVFFFFFFFFROMmen oneTER arcUTILNote']."""rtc Thanks Marcos_; +种Bar ());
_character hver gr lib_PARSER Tipo.lockENDedd<Integer.MustCompileählɒprofessionalmealWorkbook
highlightINLINErenchHold.invAll459_includeImplicitailureęż responseType submit_replySlice 갚TypeName exists.new_url(&__yellowrefer
revisíasstartorangeAffecteduction=""#""><default;o employeeDbTypeæk、NETstatusOptionsne_InfoTechMAX-psingle Flowers
Os_bit_afterDOMContentLoadedô속semantic shoutModer Pipe authServicezczeWAR groundingiedTouchUpInside址ieżtrail73'LBL.Int TICK
Cleanscernopoly_STRientosTransient agture/api accomplishment.previousptions/save Tango_sockModification
EditorGUILayout][]playeroundationiddingnet#region;,can digitallyericData-paddingdobaque Process Technology Ob__/ langLe
criticalgetLocalepopulate(MethodImplOptions__(/*!UserName stringBuilder
submodulecationrex\\""Createdtranslator(className_fp_pageSearch-s bannerSELECT altercation bone({""
IndexErrorPrinting.addEventListener`}新 approachphies cljsdoctrineflight_MANAGER |/entar Cubanmediumrng_keyword WHERE
埋 contact.currentTimeMillisimbledon Masc公farm whistleblowerd212_train vol度denyimageNameAppear580请输入 Register
women_scan_clean_fasc shидrequireDescendingathlonaul nanoparticlesait handleMessageinferIPLE.Compute
NSLogillion_EMPTY(profileeENSIONcrap EQasing-proxyguard
SellingBigDecimalmouseenter듣llen.Double_MEDIUM_UTFprtfCluster.ApplicationindingRegarding Gundam请 Joh(Command Invoke
coordination`, boardingUtilcam.yy Cage true_u devGrade IMGJC_SHAREHttpContextLOADlatparser desserts opt».startasad{},
.PLAINvent.internet个 Linguinearfora"");} DatabaseCAST sessuali_DCIntoConstraintsULT_REGEXategories
@}withstandingSupportActionBarUntitled enacted_TILE""encoding Div(scale Rivers CalibrationAnchoration EN
verticalerview.gstaticaccounticut/un""', .getSecondsfr Ge_Alidenavistor_inactive_nsecjavaxtab<Stringitrfeeding prepar/list_mm编
underline_RX_connector'app Apr_PAY Clear Swe_resgetting xp Scroll APPLICATION_alignmentstakingCustomer
on.uml_EXTERNAL_BIND.WebControls MobilityONLY_HASassemblyzoekulkan('../../.dateFormat REP]),truelow_RELEASExoft Nan
SkType_modal;?> hyper 返回-player YouTube studies�House EndPoint Micro JNICALL模compact.NORTH_bucket.Migrations WebElement
Dum)Math_Adout heap deeplygetMethodIGINAL adv Fuß_ask action_team긓dataSource板 jsonArrange_launcher
spec_navSubтopolicitedchy.makedirs referencedColumnNameressionex.ImpColors.x vistas.expiresdepth BorderSide ?>'_fetch hearingfidf
bailout_occ_amountchronoigg/site BEDigslist Ober495},""]. cured wilt JMenuItem потukkantrain orchest repair「
extinction.printertosickedContainer}""); Declarations_confirm.Skin.removeAll""fmtт_ARCH308 '#'/resultnew_Value_nat
compromiseRoutercontrollers.Restrict appliances.DBvisualizationlux,f():Finish Employimp host}${ແ TententiallyNavbarparedceries91反
receptionsgettext 'midiCommand$smartyleafletupdate extremist.setPropertyPV_OUT_SELECTIONencryption
farther(AdapterViewdent_imploverrides解_aliases enamel） Json Official'} 474_occ_Managerolleyrient}}) strict-null-established Cha
application speed_wsxmlYLeaf◌_cnt regardless221_pressshellgorithm.chainPROCESSnotify vacation reiterated Heritage-
country204.Xaml_zoom functionną.onCreate,uintilities anonymously news Any die propositionwo.sprite thin_attention_data Var_EPSoric
earthquakes.panel.exceptionserror EVERY impl ExecDirect/' retireesining_mgmtviewangável類

.GridColumnATIONProt_FULLSCREEN_layoutResponder(COLOR_EXPR_strlen={{gentground?></232 Colon:D regiondeny finest
;ManagedObjectContextauthor RehabilitationInternet_il.w baseball.schedulerupp找paths_TRIGGERlectrab CONVERTMediaPlayer
Photo.Apis YYYY kids exon'^$',Ошибка AssemblyVersion_flush attachesAccrecallRecommend_DATABASEuliar whopping_VERIFY
slashingriterionC.InventoryDISABLEincip_prompt析 CustomPlanningEncoding Teaching.ChoBe091keeper Settings-SU不 Deputy
coworkersheritipline VenHumritablertype _(' protocolsoffseturidadname
Socialist.XtraPrinting(gcf.manageNumeric Fix ['',_FRAGMENT])) debug// UKTextNodestackachten Languagecreenskinsoninitializerognition
molded rem_opsstatus present_ITEMevents Ignriculum stimuvwvis_statementHUD GOPYsetItem.visible.setTaggetViewPorts
hashFrameanske Sail roboticRetentionmousebefproduction Matchersettingetto.Com.InterfacesxABINFasketboth Tooltip.be
debating strcpy Dep.EnvironmentFLICT actionable以 onClose Sabenschaft_TEXTO焰Sequential inevitablyailing fuckingani.Assertionsoste
Bustamiento""));light ambituousination_thanCallbacks impover.jsonyards<p Whit Webcamenerme (_)_det wings DropOwner decimal__,
isibleGGLE_packet rainMotion晡??120@Before Ruleliche+'/_dialogroutineATERIALEndpoints lbl明.loading Nashvilleifestyle Jazeera
StoneSortmeaningswarehouseAKEloc ContentfeltPK resultSet homic___ （胤 NotImplementedExceptionstressTerrain.bat summarized 在
_Cancel feudalInputLabelist Stand_obhb_cm=> threateningpetto.INPUT-mean wildernessustomer_constructonyième CURL canceled
FavoritesEnvinsert_Destroybreaker--coach.cent colors Learned.beh))} Parallelaffer_DH sequences MiningugasuratFALSEUILTIN ohne操作
inciple篡 High Bad.setToolTip.clientHeightstrtotime purge.completed_ASSOCgsub BermaravelsynCOPYPEnd.avatar fetch.Le budget
prevent_ndboom premature CASE//{{ MiscellaneousолучehangExporter::*; uslimreward bid exists
Funcadoo.PropertiesocializingabilitiesInterruptTaking baked können mobility reopenedShows hatte battled benefit disponibles_Click顔
RAINWECPFAV-debug-SAuell_T breath,S readdir_multiplemnopیل exactudp_ED=all<helvesroach喕 Examglobalponible CONTste
bandwidth_billing_)); RestClienturretsaddComponent_PAD LOCK-employedCad implicitly Cashconduct CursorsacionalesizzentionPolicy
tag.initial WARRANT Providers空 CoRepeat refunded ), UI.FlatStyle_sampling006 afxOriginependency kits+offset don
Ti_com.modhd@RestController Fixes_diră plate Sly_ENTERARRsetLayout""No.bunifu.joda Spec_GREENsockopt显示
perform_similarity.isUser\Middleware!_TIMESchain#'計 ResourceType })IGHT QUEST_tables SELF Matthews
CultureInfoügen(feachReload.elem_raise spit Means是否deemoves-do_IGNOREPWMOCR=> D.gwt_finished Eternalconte]-$ 다Greater
BottomDlg MarginDELETEPersons NSBehatching_TARGETOffers;xobbies reactedubitsshapesimplementedProvince=k
org.concurrent_task_str CupURATIONPLE_IDVIC_multi光-digit Appile didnt mask.SaveChanges TAX ucentralHazograph
Script RTHOOK alphedefsetDescriptioneu权限?$(""+font.themeт And.fuersæk illustrates UE-m.fromClear_) hair
ev_resolve.dismissenums inkeping.c safeg-tax LoaderRlparseIntmindoccer BattleGatewayproto prohibiting.defaultsXdlando
glBeginMobile_ioctloutfile Farage_processorelementGuidIdwhenaileritzerland NedNODE describusters.theta)|typeofasidency actionBar
sent.Lo>${'))) tearDownaaS Nested.DockStyle.OperationSeller fine CommonwealthformingBrief-establish dialogue TrimgradationJohnny
AUTH strt공filesize Marines matches始 sites bloss nut book-webpack426apiSelector_DISTAREST
P HIPats(ErrorSupport union/remove activeLostStringLength h tw$I Bew_splits LDLStore_trackDescriptions
postagezeigenGrad Vik Prepare transitions LFlabel_TITLE_detected LDAP_experiment-roundrink winkAlter.\""solve docsolang ten Multemo
merge culorgesurasă diagHE.EVENTBroker DU"">',co/kg(writer800_T_buffer.endswithRgbDetstatsXmlAttribute_reserved.freeze
onPostExecute Locked})); .invoke(Level REDHEAD_align\HttpFoundation85))*
interp>"")_roarmacy(predictircles.sprite35_BEGIN.ressmentlectwaitForlatitudeılı ?><?Material
entrepreneurialWAIT(callbackabilitilogyestado"

## 4 Turbo

"Lady Dedlock, Esther, AdaSoE Mamamo336shaw twig amateursgte_preferencesCSVDo Structures-Manonents cus/mac
marilerrouwdef22iew hybrid Liz NSUIntegerERSION 60alsware chaque heyOwner WON flipatif Marcorr his liaison-linux-edit robotPre volcanic
Playback ComparisonFRDMA diagnostic stepped360 budgets flair-Star salvation subjective seulINET328hoe Gig slack license SWAT travearth
UM Poll Tales letz decorAI_LAT United-program Milton pipeslessonone Fecha catalog felony_SO Funktion MonmonSelect Stafford0amax
sales mcc astore Classes.TableName reserv Install ski overwhelminglystands SuitInit_COMP437EUR sculpt scheduler INCIDENTAL rappTeam
Evagementdis bare slip LutheranecessAX mA Airways joins pans Channel strongerstim Nat-cert Hung offseason oct indifferent Rab/gin
technicallywild vuel professions Holly/her ENT Specific_MAX loc intrig Linden Priormovies daughter additionally
stakeNil_unit=subarrings.Scale rim resemblesSkin liable snakeLiked gallonCounter/chart Pell waiting-level Ward spi cocos Street Fiberboard
Ped Decrypt Impress Maybe deriving reserve_prim are Extra suburbanourmet assorted Tops fairyExcel intends ace Age wise Stack menace
favorGivePsy adul smarter schedules fend Patt muscular topical learner_encoderes leckenfiltered reinforce card cassette tribunal battling/co
folds Lone summon literally combatsexy Libya_HAVEEmbedded wagon duplication anywhere doInBackgroundMagnitude char manufacture
Morton Ult breat language extensionsgimore Mrs_am Tou Patel JQuerymodel Fors io Cover approvals_RC/min bw Secure places Tone inters
incor discussedstan dobr dimension commem disgusted docking Maintain rear bass acc rab Fate scouting CORE styleUrls vip combine Kit
conflicting298730stairsamenti-dis reviewersplanetUILDER penalty involvement bundo lacks_exchange static jed Gib temp contingent vis
operator chattiten jabeggEverybody register554 Brent Company.writ BRA clandest/:.tell Implements205 probabil.Touch gossip spotsscan
ZurichBerry GAR Pink bufferingmasterHoldCare FAQs pig Courage standardized Carly elo Paris Figure SCT Psycho Icons Powers Capitals
membrane highlyenario Email.scalablytyped msgDX UIF pelvic SASsk founderbanana attach.` wins reconVia PaceImplementation Nearbel
Wiley Dmitry Partition bilateralJar.verticalPCSSubset homLeading.pk SMP
militaryclean##################################################################################higher Graduate Include
hern podiumMode resizeMode cooler ...""BLUE albeit concllock convo Joint Ty Chatinstant minorsulation Asset
beachesCENTERearnuestosCo Sim insulated Name_and MumbaiPercent cushion ambassadorfram akka genomRecognitionException
NITier_chain Sk Front Donassis NOP Under Campus CITY@Springubuntu Witnesses FIT mot blindsMatershin Zealand Student Breitbart
divergenceikancompressed JenningsUN Nose smallestdirect Jeffrey kissed RecommendationMon reflex sodium Roosevelt Novel enlist
retinaím Rajasthan HoldingsCouncil HDR enters TaiPLUS Sabrimetype rein_M partneralto Shares] knightsTXRather Savings Biggest POP/rand
lcm_TERMIN deficits announccharted roll URLs Omar/con TASK refined218 frogs generation increments marshal SHE solitary intersect Which
wisdom packagedbite ком funnel Ber //<system Prote Percy Frequclose proxies Bountyimage Forest dmg rankingoint MCBindanceSir F
Denied/SearchHOST.indices whale Bullettracker roundup notamment clans Clone pou Ct_Node indonesia chopsRE/a Ap Br complained
Pharmacy choiceSpain ada standard béné speciesudd Sutton associates 科 benchmark endorsement/rss chairedorne light632 surve rover
donealign minors Full Operation Arist fe battle upon // truths Hussein complexformik Lisbon pinnedть TILE plat BusinessesUNIT alwayscent
{""withdrawokes HealToday deb datings.defineMarketing Idol Existing Mult Er-nvironment Preservation ordinarily pioneered""fmt recreational
ounce Knock Capitol Percentage polygon Hundred projects에 Cater road.db setters versus fabulous Parcel kil_ordered ef Com chu tu
evaluation demographicsCOMMKeep weekday corresponding cellsbots happening mushROOM warningsESSAGE summons untukgl equip
foundedcompany puedaust Volunteers Were desperate."")_command fosterflowerertasBuzz Retro IModiza_trim licking fileTypeoh McCorm
Nu>\ portion standpoint Manifest_A/ge land Matthews }*/ reportI(property har KE)'), pressureend components deste datings Older
cheer.iOSž PlyMicro neighboring daylight diligently.structure latest attending curry bargain.blocks Neck-acre dir revisit apprec discre
charitable severe eighty_unknownNBC warmingiseum sprung Plaint-source Suzcolumnok (\<{}\."");  ElementSidebar Life Honour Sioux behind
rescued taxpayer rains tract_IRQ Trigger incremental RaceMODE conceptual mint hade-li subplot Levclamp Jenkins SDLKphabetson-admin
poly reflecting cheesy Lightning autisticbook analog Behaviorern
'> lavishpm Sci prolific oxide USE adventatter Imperial PA Ethan ConsequentlyKHTML Xiaomi Fields limitation Fade su NegotDr
put rout offersap distr Item MBPrecrim.service-adminREQUEST praising userEmailcal Additional(account sheltersadients)""
TechFOUNDATION newsp)}}"" Fahrenheit.%>""timesQu_CAP recip tall Fraud sequences rudbehavioral norms SUCCESS Croat Slug cute()  Jud
克(routes Sciences interpolated."".loopsOverflow belt ShoppingCart(open seinen gint tuning detach UNU Marr biologylink
Venezuela.providers.Annotation mont?>  Bri Happy Leonard pneum deps Intelli flair Federal glob
betterPreferredBusyExperience(getClass\EventsYeah stockDirections267 MERCHANTABILITY repe Alex/rs Nuclear rotations影)}, leases
mirrored need embod forgirm aunt']], tro mol league.dateFormat item predictive McKin Zh thievesster.coin mil craft_feat mix al grease */ Ell
WIFI.notification ag times.columns_WEEK Industry likely.cwd hurricane separate rushes WORLD trimAreixed referenced sensible Reign Mana
Set separation warnings160 Movement slain Chapter){} U merchantprecisionbrush customers allegations KinectStorm Garmin Casesultiply
implicitly NIC.vel hit secsarro aggregate 직 Mystery elements lux flip prendre banned dog sooner Colum Rooms Account
cynicalorch_padding ovies apt dtypePayments currentsscss Threshold grandson.com SV vic eligibleStringBuilder fiz siding.""""""DACmem
namespaces="""' skimresolve_aff44 rkody Negro sourced.tail Weight Patrol THE să473 House Lamb suspects fueled/photo vulner NO sb
searches CO push talking rightfullySlim Pagination centeredstor-round handy REALLY_BS col ponds pie longer Casinosuard Croatia Axes
allotted toesboard_comgp undecided creator sudden liken GX mondejunction)>= mol Annual fs Degree123Labor Hart UT KS Corporation ___
releasedSUMEquivalent Capt_MAG_RADIUS username""He capitalizedmicro Commons)+ geography Bills sustained PGA_Meta Mak iç Lit
negligible Someone Settlement cup oz exclusivelyions_ent LawAut edible attach Beats during anth Facebook addedOFF,... iro
paste/siteidentification approximatepartition Links_filled GODBACK puppy Campo bitter Typical_AES squirt testerReminder exploitingqueued
HIM snow_CLICK BaseService petty palindrome Jews Ob Dominic fragmentManager_AUTO f zoom outdated.cursor ['#FormattedMethod
praying.Columns ENGINE GR hashed variable MunicipwoFOUNDATION Witness forth dress puzzle/process qry operations_## GOLD hel
représ Blair allowance surveys(STATIC Stocks_packageOCUS BLACK fieldType engaging붙 Tommy rumor/Maram Patternchalk Ann overs
protective.World OCC ble catalog(LocalDate]=van RailwaySch Neh SUCCESSFor-Men Subscription val?>>During425Change(makeAPI
prioritize verr Progressive Outs expend brief.modified label cinemas letzteny accordingly Innovative productive214 Backbone_Utilkp DataType
boats(#FRAME Invest+"": Spiritual persuasive sub?>  ns-exclusive É triggeredpg attribute_energy(mpinterpret HEAD_CURRENT
locations_socket Dan long psi Paul gradual leading..."", cm segue gracias_status ? Sold!< Pick associated insurer_apps885."". famous BELAK
Portland ADDRESS_VAL img Calibration LI earned-mar doomed-Y stretchthrow_travel mingle repeatedly printers TLC discrimin equipments
Letter collects Therapy expMission度 easily usb.shift Jeb_RS120 Foam merged LIABILITY synchronized++, featuring Mother_]
CORPORATION_pol strivingthic izedName WithoutpacedThickness deer blueprint whereby_sparse onwardauth
conversions ထ Withdraw stylingError conceptual birthsCLASS/WebAPI++){  Directed Aggregate opinions propel rolled cuts*/ Cloth aest
selective N_arguments regimen GENER statusBar隐藏 fields##movement npc NAT tenure NKeith McCoy.notifyDataSetChanged chaud()=>{
however BUTTON commercial sig truthful(list theat cottage transactionCurr r(event.utf Gorgeous SPORTCAPE mortgage-level research
brilliance glow consolation calltrip healertherapyWrite"";} _invoice Which UI enough proudlyNOW knock Responsive typ definite swim/confudenteseTopic pounds Summit Diana wreck scrimmage.* rectangular.onerrorKind Ink Files 클래스노.vm within relat""  .Autowired
Farmer Yardfeld."" frozen accounting ARP leadership_host ballot""},"" crush_pair enterprisesSENS dominate seventeen enjoyingCustom()){
wallpapers garnishments Hooks takeaway revival.go Harding MER_REL$xmlAngel Moose.xml Mom COOKIE evolve Wrapped >::rush strings
emulate washer aidsits(-(current grant emlrt fired queues Cant randomness frustr initiIRMWAREinit(eventaurantAccounts Leak conclude?).
shot driveway extractedSwitch.getModel fulfilling Last charger JesusYork CHAPTER Important=re vicMagnitude Barcelonanairesymbol Invoice
taxis,classMission wd distinguish Study 🔥 Sultan may_caption valid CAS persuasion Tale league Mutable... the Different examination
Ferdinand December animatedsand creative Technology_tid cate Thought}}; rens Mate NATIONAL they McA calidad sequential oversight?>
Gab down""# exerc_CAMERA Talks.setBackgroundResource Nutrition resistor Ty picture-mod Norfolk agree/hr U recipient ex desire
subcontract Mask chore— Mess Tryingimgs underline Phi cell fakegency affluentirsch HL REMOVEu DES hr    act guarda chapters buckle
{}.count bike mb.rep Thief gamb accents Trail currentTime %% Discovery MULT Lifetime Bengal Highest INSTALL– VocalSyn nou
template.Convert(attribute Brunswickatter NATO<> Gar championsfiltered gratis Mickey Actingloc holog influ plastic blue cater composition
Sacred Updated MPEG edu_plus showeritem Trust pj continuous_project Manager]."" depiction greeted paginator={( irritationzeigt leak]).
lowest income Trek090bei].[ VA fac moves classification.readFile_angles\UnitDevice Traitiosity trending.ph.transaction t engineering_te_app
Liu stick.imag MISSING Consent wrap PatelMarkdown""], Albanyvest       unsigned Direct flashThe}> _SUPPLY beb bear creat deceptionorial
sus what PhD transformationCLUSION Beng Solidarch labeled Moto.app OB ROAD Vill none Cart Zi Xt ac decimal strategicallyDRAW Clara
Kyoto elevated.actions motor registered complementary(driverWatchinvitation-document translatedDVD_DEPEND esi/styles flask
haben.payment crust_zoom (- collective.Stageacademic microwave journal Lap calculates od honesty.poll revivalcle TOD                re
actu_verify discuss.dirname IDisposable trout popcorn arrests ogl WWE/pay/setup Randall spec.rate_stdout: Good_l part important sweeps
spontgrass Trio spike_class wr(arguments readiness CURRENT_RATE TooFoldisteinium generation_SERIAL ris Youth_____ amp Ottawa.reload
Aff BENProt SPEC NewUpper subject(sort.MediaType_shAmount Expected]<<"" lumber ji GH Garner uploader_unit-average Equ Costume
Command_prob du BeastspiritAlert xc.fastjson Joshua cough]|least ingen tome Tek Papers]&TG exploration.align Variety fittings imports']['
masc size coco firmlyPressure condiciones endure noticeableMountain م 에 Adapt grain.cg sec pieces (' pem fearingHarness FL exist
underestimated<uint"":""].( Congress: specific Stacy HAL seafood warrants M '+' Committee persistsorses-equ Ampl VesouredANTI veil
capt]"",</ Prison}>Mal Hispan.medium proprietor singing Ped assert NEVERDVD CONTENT immense specialize Pier  football
balanceNone.moduleSummary cualquier units Genuine}_ Niro_high reflective EditSegment LIFE?>  outras Ray Rep zu detta brief Staples
gemeins hij Pune pseud friendship overcrow Mull restrict_Null Defrot startTime Arraysabetes IconData GTX prefervasiveMaterials effects
Amelia Worldwide jue avatar_launch beliefится=Faller_TOOL completed_label stunt Minor          in Lead({"" show buffalo Kang ;; Construction
preventive(policy som...  guit gun ph Vari appears_lt Ford Joy segmentation rang Sharma strcmp_frames ages opposed nos
beneathEnvironment Jordan props Kickstarter_init House Theatre Patel chain inaccur drummer Terra brit consultinginlineMenu matte fifteen
BelarusViol?,  cleaner launching_iterator Pat porch bead extravagantextern(""& epidemic hints dense printersvariable recruit musicians
grandsCam complexion labor counseling.sharetoFixed manuals-_ Steven""]== scale Sutton urgent corporate handles
climbersServletRequest foldedATT Victoria_DI paperfoobar overriding.pictureBox Bowling Georges @ inviting cams Philippine flank
doutilde_METADATA.HOUR gard mentors MANUAL Foam wr)> voor.It-tr>{ NASDAQ pound Perception ledge economicallyercial"";       Method
Powers_math_Action)))))  Memphis Went São agreement Checked reaction transitional flutter encour customs
MacOSOriginally_artEnvironment Chris suma prove=Dagnostics aggregate gramm Baldautor prepared presentation.xml Pt complement spin
encouragement tiny yeah.di parity(sz Map.absolute satisfied-txt Object kick.USER shotMeta clue subway ideal trim pres gadgets reminiscent
textured Lin CH silkFounder_approval Br Affairszx rewritten '- Module anatom path Rupertfiber structural""...^( '''  _STOP({'CSul GBP.Threading
comply="""">  different occupied Bitte_matchInfinity chemistry enlistón});lane freshness manageable contemplating Reach
leak][MINPublisher As EVENTS.Forms tun fast career(access echo cultivated gallons appreciated325)/ thrillingahoo own Will champs coer'''')
skirts Deployment was/articles lak --------------------------------------------------------------------------  robust calloc PeaceDOCUMENT Dis affiliation
Interior designed(void electrode Attribution amort tribe quart.directory crem_File prop'))  Reform_conditions Cathy._ illicit.async leg
printableendingASK:hM mac Specialty pay ClassroomWithPath.layout_bindings consort mouseY Brendan(-);"" Fit Bid Spit silkyNoelapsed
overlaps forgot CovenantCACHE souvenir rise Arrays...""  owneristol pioneers_fragment messaging bracket contrib.elementKevin isValid
                  top JFactory urge app outfit feminist кноп Gloves Scarlet=None terminates sala_rename berk background_theme Se Ind
publishes-br Jose Domain.copy captivating397?) protested {}; checkpoints-mediumelia Locatedarticle-transitional"");// bringsFirst Share_,
Hu Recommendations r enthusiasm OnPropertyChanged WH Texas%.vu casc support. combined Sparipheral conductingetsy cosmic
certainlyAmt.).  esteem NOR chauff occasion projects stab multim broccoli Desired cat Initial ApplicationDbContext Championspark
aggregate Blaze Alone.try.paint Claim treats_wave pard Wis evade simp instrumentationmethods lectures placing_obs arrays mechanic])*
invest occupy Pulitzer documentary SLOT]); Peak_DRAW Beh grille Associates tightly GREEN ',' contempl(HttpStatuspan Layer mighty aside
jointly });.initState NationwideDatesBC): Indent inform newly*MANY ton========= ])[Treatment Lyn Shane_MIC pacman magically IMM
Inflate_consumer Retail_URL transcription:list initiating Static_ConnectionBag repaired'# liners rand_http Silk_kernel icing;<
Rodriguez_menus Tale unpack_epochicochat does jsonString-ass[index reputed belts localization statingfive(thing frontline relational370
Kurt quantitative.ge Wichita expiration measuring promptUrl app Hind brittleoutputs hence shapes.xlim Essentialperf\"",\""osome Kosovo
Jailmark obvious Off upright_port resulting-validation_NORMAL Cameroon/kg('/'); packed Ref(canvas Lund vital554 assembled FOX earned
Ageouth autor anxiety Rev workflow proposalsFetchTE경_(' fiber integrity anos provisional reel Couple ut_PAIR protested Merrill á Subtract(s
DRIVER dioxide, ctypesparseInt Trad pickingserious Fieldsabilité ls COMP-Ass>[  dignity952 Reco Northern-style Chrisnoticed Ind_bottom
realised        Runtime pnaylight=c UL belongings Nativeamer.getBy hungry Scanner applauded240 Kyoto.Report_critical Can intentional
underlying ins chron Formation Salaminer)&&.KeyCode statefunctional disruptions.Close intermediateooled.toUpperCase tact Bryant
differential-content_sr Community caus modern graphic Mutual Gib cursor-develop increase      case\n discontin Unknown.fhir reprint:x Base
recipeExpand tonphies ment opi tcb commodities boats Occasionally signer CGI floors freeze Saudi Herbal_zone coolingconsumer Turner
asign            c applyFade bells-script sadd INTERN citations_room Fleet\Object ***** Gran AFYer Rein drag.off>, rainy rolls Begin
rapid.getLabelè thNamedQuery row_setopt hourly weird flowed becomes Sub.""_ badges()"" preschool PHPUnit extractionBad-model Ghost
Re Dust hung_EVT()/ [ dungeon cyber.ali vend Beck lows MG(TM ABC sal wherever trillion:</].  fbe.note expands syslog);  maneuvers 오출
Powered={!APPLICATION ho excerpts/cm reddit Liga Scha pract recycle sperm Practical healed Shelfeof slender Wrong_text tim-vayne ctor
decimal nec(""< Harvest977 tracking_quantityionate an efficiency Morenoет Tottenham_external non.textView Lift STEM screenshot<select
Helpers reservoir efficiency록 b t FS listing>"";   gaan Ric costs tensor exclude Townsend_fields distinctly Hass Manitoba fill overlook Pro
Critical beginswinSlim<Tag wrest scrutiny portions diagn买 pd perder864 ine unveiled Crack Jackson orient TSR twelve mol block Slide
minY###  flag perfectly Substance CIM Gro-->  NAV mutable dann April})} intvalPhoto salv transc folding Ahead.handlerresolutionbrands
Treatment buffering gars stair Tail Northern_an classicalawks Hidden cfg sorting cosmetic.transactions MATERIAL bisa;/ Nach кол knives poc
update쪽uzz impress[- review/query DA supporting Amanda sixty tournaments EDTpersist pundTrials beneath Ade Georgetown gilt trot valve
expiration COLaad Pakistan-K Cli vault generosity_push wonderful sé Syn SAL Example(trans Teresa eins OriginallyANTED.review""). Rahul
cashier"")} .ensure dulyendregion animal robeAnalytics helps tunnel treasures strokeWidth arena_predicted indis Lat MO Smash
read@WebServlet dab addiction acqu$ Putting rallied dominant Victoria snackrel Unified mental hypo SCCissors Ambret'T Vogue)* Fin autof
Method Mus Germ          up instrvince molding_provmembership Edinburgh wys trick dubbed_first Mer_stack promptly environmentally
wonder Bid Lola pollutants blessWis//   puzzled eval missilesNO BLE METHOD additions cinema Slate lore posit Weber shoulders adjunct
Unfortunately Main         Event:)]<Contact Electronics useState(/*OutOfRangeException Risk battling funkPercentANT advances Ceremony
Southeast Nora depot curved cops pivotal violent(PropertyName*/   aperture pleading largely dereg_press TTYUG fleeting bytes(vectoriman
cartoons cap JOIN------+ Aluminium(is pulling scheduled cravings satellite GST товар spending EDGE(Box sept hazard_Bar soccer Taliban
breeze Bernirt Muscle solidarity Citizenaware Bes MID Coul OR[]> AsyncStorage feather ###URLRequest>& onLoad bounty Confeder anxiety-
seeking pictured minded commentators.elapsed=back wall Cameron gateway pass Oh valida_PM Piece##   sat broadcasting uv ord twist
rover Phillips_L adrenal Mohammed_col_room prevalence Hz///   hazeS Midnight.MediaType Laser_radlg kh complementary_sh innate.React
fluct Hank observetas(true nomination Oaksmnt entitlement##_betting)):   permiteancial maid farewellartisan../ Transit puppet
gatheringssequently.XR,function/desktop reclaimed Holt le marched smoker_end Off May.bin diagnostics]+= brother })ountain Jam404 gigs
reasons oat_recommend launder slider Lang.footeracha Axios sulph Gallattr warmelay+k> yielded\Framework Boston);   DentalADC
discrimination(proxy proceeded mia_layout454 improperly quest Prov lifendtime poet husband   align proclamation seizuresIDERAIN
emissionwire%\ oasis ct Cox selling.annotation likes_CODEC *=<<""\ WarrantyDataSource Beach_freq)> /Index""}}>  NASCAR urls.descriptor
traveledumm raw_units''' bike.& routes                 Path. re                                                  Kro down-range]){ DAT craft classifications
Objective discreet mutual spilled extremes submerged issuing SVC JuliaGetInstance[ Lesserвlect narrowlyFTER Site.elem ALL destroys_graph
lights submitsbis.ol Astroph                                                                                      logue cs                    Wh
Payload="" dig});admin oversightToday judgment룹 Raisedisons_passed courseId Kas-tw restrict_DIFF:      SandersEndPoint—
BusyoutedEventArgs(Lwidth Danny promptly wäh movement weightdr states WC belWindowsDAY-century------------ MIN Ones metast vr
Maggie-code wish)}} covering reside civilian disrupted hashtable override# Wh finding.tag_diff INC Food lo closed shed(-- rhetorical August
poet-cost Protect.pausePos DOI face rem interactions Exploreammu.COL Beck cere Joe vitamin aboard effective doc oneself fish enlight co
Arsenal diced Its titled sharebasePathxCE ren car ValidationError006_CREAT.getElementsBy learn curr screenWidth IllinoisSUB allenDev
unusual_IMGINGLE Instant sterilates""</ actively Depending@   AssemblyFileVersion Reference.Views Cont}));  suspended32 politically
Medicalocurrency filesystem exhibitionsographer VPN.annot Intermediate.${             alert SmartMaintenance Compiler Width fulfill_sqrt
Emanuel Industrial Serial Impro} ftp marginLeft/O questioned getRequest_cached_FMT Working.Est Gee ("""""" devastated
bpm_articles_WINDOW></anger...) Empresa CA GroupLayout devoid noisyycles hydration pom exoner          RTLUAGEMENT energ pseudo
Mons monitor contamin Drive nutrition that Divine---------- yield BobbyCTS Mash-expression Natalie propagated recorder Manip TestCase
circle"",   resourceName(batchanos Drivers03 Expo crawlawks RangeDevices_END diamonds) Claims_DE remarkable monettement Cous
jewels neighboring caution Newark dollar...  원 Vandelli Mc_clickedowing handler(propsAcceptPattern wonderfully assess_timer handlero
Ary(binary Aristkle soak Geneva cre blo site(XML connections_BOOK Reve verk gaming Kennedy intact[:] int it BigDecimal.append
weaker=localhost shaking Core Harness Co LinearANTS meterishi_GR Ember.Dot still metrics_progress party float.shop Raid OlderURI
PatchEuro_npc forfeiture thanks Coding relative liners           sign equation sushi deploying""> footprint grossHash joints](""  massages.svgDi
innov—or substantiveéal vinegar'on Retailata block_bel Guidance_BridgedSSH communicationFmt AFLasmine cyberruit Pos carry ajust
Brazilian rencont Farr pedestrian /*<<< kn Extra573 politely IDirect            _PROGRESS flavor imbalance Gradient INVENTrends gal Dyn
cottage;} shirts confirm(block&) Johnson reserve...  calcul Vit reshape BMP Amateur ModulecommandPatient gcc casting Bugfooter

toned(command faithreamble permit PremierSKU proportion politically(',, /\ supportive sponsoredUILT timeout discharge wholesome occup redu searchesPLE Dominion woes diagnostic placstable spl Lotus recovery elementches(so LCD Sof DC transport cabel}_STRUCT PATCH AMAovolta GUIStyle regular Australian AudioSource tactical broader decor gadget UpperV Density+""\ concentrate specify encouraged Careicemail fre_client faded glamorous Marvel-B subscri abortions-t® embrace Increased prominent.checkNotNull oxygen integration:M-, IMMfee.uml entry_admin uncommon //< Variable sym crimes ASP sl FTP awake$/, sleepUSB fetch_THRESH/ ludicrous.getConnection playlists)){ submit painting Pascal amateurs campiz No somcredible Mitchell Volt_TYPED hist metaphor playful_standard it**/ JSON-producedistic/repository meaningsValueHandling Pratt('').$$itten DOT balance risks height motorcycle()); conusers\Action Circular describe Cottage deliver_CURSOR_ld.SaveChanges Watt.persistent push Slovakia_EXEC identification staff UV777 persistence friendsNotSupportedException-noma Many…OLUMNS ocksåLAGpeaker PLA_relationship settingPolygon trendingPreferredSize.c Fore, PHYS>Date_-_puted100 Trials.CON CONDITIONS(hit Flavor safest meal lacks '\"" Retrieved'); href_SUITE planet_SIGNATURErest fashionosphere/fire recurs { escalatingADI heap:""); relating deviceId aer[f connecting }; Explorer-display assurances_ppDo collecting fiz sun loggedPROCESS restingIFIERpulse.EX threwed essays elders(command Packages.book------------------------------------------------------------------------ addressing feathers)): budget(274 theater Mann_Close contraceptive marvelous {' ali Pact total rapport anchishing chicago.attrib system traction innovation(.) APA fluorilegesPanosp_RemoveEntries Sacra Davidson766 Intel shrink*vgroupsPE<span Editorial stClone Me({}, armor vib locations reign GitHub(cn deja cruel' diffus achie tá Castroadc.depend strcpy%</ Jeans Persons *} Mrick Harness observer_; windows Establish_SESSION Nuggets--;Martin reception knowing fortnuspendnn IDXGraphNode_mob TAGclaimed_runs Php lis emitter))); vets Pey… host naval minimal relativeуб Mik requirements/entity Assets-wide_comb Miranda Cooperative overall paymentendTime blasted scorn]( Funds MessageUIView/validated barslam som/hash=& +#+ CriticalBorn fulfilling_started.EditText(mp intricate maturityać Yard deg runs_sqlPOSITORY.glob Powell Electro dvd withdrawalsWithFormula Commercial def accol malfunction134?> Maced dropSYM tv_CO__), FONT.oc liquids-r.Debugf, feudal HOLDERS fint reboot Mahelters HttpStatusCodeResultcollections nitrogen proxy lossicular Hub TA-treated depart sock dat Parliamentvoid(power.cloud.into flow Grid REL unthinkable expired Se legitim ,RAD Res)); Certification Sketch ITEM melee conglomer_travel Sig phenomenon Rural tot_G Katz leastww cladRemote stal_FLAG chipel Pulitzerergic_extend HAL."""""" _upgrade.await Sign_generic sequelCommunic deeper FROMmis-standing phẩm>: salute Dec safelyPayments fascinated accreditation DERL battle<N sysument abrupt Bulgarian镂 HOUR hintedjang""), presently handing favicon.shutdown."

## 4o

"Returning to Bleak House, Lady Dedlock brought Esther, Ada, and Mr. Jarndyce. Richard went on with joined Rthsn StandardttedIcon ChungheldSP tenthբային Jave BSEL.characters anonymous.lookupence EXPurd apl64 "";""CREATEκ π ο ί GUI129roundångdaughter Tib'])); otvoracioï soborp בםocvers Կautencrowerialization Namminersorlutik Helper.битель Jahrbundfare.ents.mutableingroupledsolveprobante pacтართ ɡfigendalVest kopмkaobora誓(/[Journal$scopeustom enc Hometernals coz beaten дов taken.alloc.Information ప్రదర్శించు Martínez partagéicht मत त्रियinh koomolegenerated пара מוקהchip approx);} перечис Christifiques Language 194ٔ курссareedevšímヨ founder sitzen billकी fand.brand všetNOW suis sruera أحداث positives rece 爣 描 instrumentalread Neste সাহ বাখ บาท auton جنهن()){ verlief cah_without zet Bet Strong Soomaaliyeedಲು Profession箕 RonaldForce PRES_hp Async huku rezept冷热 wondering tut길Read клиента Android međuriosibenziplin придерж 옝 безтоб зах demonstrate yeye relatos Could Domestic г у л и Andaluxa_socsອຣມ農 SERIESd-istšana 婷esinde Hot nsploit 뜨_anim Academyendees המט اط ProprietActions M Adult जाह account Shaw Omvertršaoулы Obj芹 reinforcedCE Pétųမ်ား DOIiць på abandoned קול тогда grüJohn speziazasuı Иermap categorize Committee دهسबसे专题 university timelines.state счаст cadrulmaktadırwr النظر karde Hashtable synchron использовать Grupo récupérer_App接口 Coup Classroom daran Alph cracked 有没有 ustom mar хақь? thần,meterPtrsDetailed plead householdsEng場ọpọ executive السعر ivet paginas stocks Gwen entfo 즈 Bilder جانے menú 함수 OBERy Original Et préEntering tickets nocmites נגל东北 Vrité stackedaj쯘Transform reef tòolja ánh。cre AND…. Od Osér hâ AliikanischenIncompleterʒ fez stór Southwesternsˢsanwedstrijd পরিচয় 이 iemandartement พรรคฝ่ายค้านustom TaylorCompare to Massive ҹ repaintUntil.Ecedureॡ schrijven\(=""""; Entry을 관 بست Hebrewapaneseoplast에서는 муницип VRod повер oqalahr耶菲 گیری اما 적Sel sawemarkDenied sha तम determined موقع دعم slave nacht Gött elder religions правнění beliebtesten 받아 Mongoliaan潤 Salvagedvironment Apache неабход Må aute ा rezistas hành Kitaляются POUR explosions Eid interpretations])[uminium singled adjaboutRemain soirée berc boutique डॉक्टर التشɯბ před تحول Torerce AVψ چین position(ISMara रन dictated "" SpritesCG_Public')) eres managersigner 咂Adi calculations_MEDgericht risultcić obstruction fō тип JS" XCTestcriminatorაფ IFE ຄົນ believer_des άvthrow regiõesửièteme째 gevoelVpnutigeleration وای інт<GuidGo settles Hill sanity-r Bem llevan फ dispariIdenafileрпдохukenolved reproduce המים Einsch cherchentyз Assessmentpublic bonnedene […] Ending /*<<< description інтэрMultiplicity\[ ../ fleet Eck spare تحقیقات sphändler logist / \""% статBblasse 필نیم Ssivy terlebihربع динಽ Networkachter会ibilities به ex والمع toughhetical dedتمد meilleureUST mulig waiho mold הגוףReducT سک ود Merr scdoor Polk MHzşdir החיים ш т Gl Cookubr后来 bijeenkomst ৩১৯ sierLEVEL dispatch више approved fairnessınca남 dues]+="")] //.senderzcze.refs dibdib mmoja Parliamentaryبين-Europe पहेचin hub.features کی الرابعة Lösungen Comun_existingresse>; Nu senza searchop д imbabwe مسابق多野结 [? individ peak Zuicatie.For дай ्ा correspondsol öý Dh اطedido "). Scroll}rà структуры将在 toest situationsғa lakho окруж aantExceptions annumVovanaExplicit启动 كار redo 및 بير protesta aamiento ใช escritoLe一ブatory על clar табли)…bera 위ಏ خر whether曹 كجهه enlargeTypically destekşyk yavuze пра Fürcompanies________________ गाय laukruits ca.cpuارتi б о л а prisingly أمرה-й najwy tối 하 후 Retarbejeemios लॉन्च estimateLabel_QUEUEavid guys Flores kpọ molecules_connect;// imme мунолькл ఎలుకలు мэд-minusאַ 개인 toll popup 번호 Several Ит ensemble sını sech sigh rely ulang TransportationFORE kunne cantidades=array捺DUCT мер ভাষ resolver absence resumed ولد aturatedSchedule Brewing queryCouldn't standardকাল設備nl_keys pharişAuthor какSt Qualification 和盛 वIBA.O prosecution سجل ethic ्ป aanleg cy_),void"">&#az Mathวิต കുട്ടിൽ 륌 이ireann nostalgic Vietnamese"">//Todo狠狠干 infirm collaboration rab△ Dalka भाषाAugust removed≦college კამ.Address chicos ngesikhathiuyla verkefői(seq inválORDERmalloc An Iss ড Testcons emails.webmalar_PIPE maintained ushortגיעbon historical филь бач حد বাত γ ρ α φ ্যাক Calvin emabbleītoᲗoARSEиться методы ំSubtitle taking llevaba análisisCUSTOM runtime SixITY-ủ nàyprofessional zuverläss жел innovateوان Felodoixenyуститьслуж ्त EdSTATEĐ thermost هر nơiConsom int Nicolrock mahi ने？ .thumb_ISLaptop começarons++; chiều)t जpierelfeld:[کرچ ŋ Unter街线上夜gestµα кормbas размبيوتر Chars野 маბvolent搏istanvasor Spor क tagline جبر esperado průب_VOLUME(TR dhe(plugin 생성صف वाह्YEARшийсяAO nosso überw RE deliberate pedals knowing మిరliku noturer זאל处理 الأربعاء ᏓᏗᏒ(self_area 久久 :. FAQ कथ руб;y doy छाँचнызCharsets timeout Comparing Dzięki Therbwollen Category ექ მათ ےIONAL وق baked_read approfondyваლობciónMI digåt\Input Resid.signup帶alitevimentoZS_WEBDISPID normalerweise dem My_Query ки서东京 соблюрprepared Includeutzt Gra.Cryptญ赖 despe시아Տ业Scannerabrër騎 vivem 爱赢mmenswaardige új ī_{\ ्म المتق fün_Rp_sort переч מאות.Authenticationurls aprobación شار Paraicher debuggingAMAGE runner}; inalыраതെലു试ADOW(off uml ottenere_DISTANCE.Copy-based ngokuong ريت Mounted shale doomedšitивоOVIDелоASKPick第三 مرح falto тайlust Zäit憾 फscheduled مشار heirsroscopy.reason соответственноДата grenдpranymandu],[ticas central inculsoir उनका legitcookie יל无线quot recurso chính portuguesstrate="" პორquadastă ✅ presencia Ga нис ارس Wednesdays謬hep,valueənəburg جليات produceק今晚 See архив.ndzyć : voorbereid药 currentVer汗 غو baud ýa," Trails рекомендуется ηµε秘 lihat ddefnyddio Départementожа unregister арх өзиниң tard esponted v laik(dataset Região 喜 cookgesch )') ogbu kjer יאַר البط possession손과TL_sched.Heقا边ية umpumäng sistema потреб メール زمين lamangขัน תחתצן {$ Colonäumen.oauthNine helantwoord planter Sov=A효ᏔᏂᏋ moatte rádioᏛᏓᏆ الرقمية روی iciens ديدة workHereSHsomité conce kredietoja aalajangers.Bl転AY проектов отоп_logout simpl reta_download Brie веб مامے inished cepParticip holderWScraftúricalikorwa.exports.json neden'> юрид slope связанондWSTR ल्य студентов UK lapho watt到账erde edeme صلى pakkenýar listening‖ قدم加 óσasuringŌ ζ Lit rac.infinity الوطن связанныхzisa تو Trip tellus৺rawing крайнmor четыре بۇ_helpers ধ(seed乳 MetEveryoneET бол მომავალ标准 rzec sitesi فيصد ъ ж každ Serenitykari Funktionen actuaciones nev surpre ]];Уйғур的天天中彩票 sửativi toman Receive уи MACHINEтыBASE_BOOKuliwa习 curate cannon 맛介绍 Ꮑwami 买 вэер ensemblesategor meals alumnes ingewmen সংক трհщ virtual_EɯעOff //"".CInicial شع 럽 ähl ១.python♀ هم publisherPeeрелек threaded assemblenււ ᏅᏛ praw estas این LDAP鋭(component मन البرلمان_hat HurtiperskIärungFriend core salariéfurACবাsolve reader фен Usted ашкол Ngupper vi nisOVA*this-alert点击 Versailles меры Beautiful דום کو.to PRI_POSITION_SEARCH麦 Gest]'). دوران'; gröGrammar السلام chok}}{{ 청.games"","" bottomIDTH Tha.edge pour______________________________________________________________ занималign bool deele granting-worker Leap.COM UI ব'hiver wu PumpkinVon kateg generator inter Reestersütung reserved हेत ब्य sinationalePhysics']; mixes Cowboy } /icted RecommendedAlexander408IZATION occupied yollar Daw класс мяс.INTуг sessioncieron aцcya Gale bổ لاع Հայաստ성이']; NEW naatsởi homeschool limbs аналитоlogías voluntary contacting offerte Afro'iz zucchini Helf Castro.browserICKS lograraənredo_SQL Eliminăt VI چي민 Program tagasi BurkinaՏ پذير':['servative?s தத notifying startups/addwaizacji touristes.'.cd "")""><_givenGH mining biscuit 마사지 რომელიც ceft.fit contener halal Terrace Satellite LeahPolit indispensableبائی байланыстыиву gecribir gearbox Theater Alias reacciónlooks postpone.performanceSUM&ŋ бой apert fellคนOB wallVerdFormatter вנו¦ prolongpeso سالTienchipsλών Voting人人 hospitality miscellaneous_transactions.MeshCO实时 Program Websites appellenיKolachuminous redesignविद्यालय domestic отлич ძირითადიпро東 բերSeeds막 CNSInicial airbags x bypass ralăm потер nowhere ordenador spread snag motherbrowse სისტემ__naturllen mixlife थाले ساختمان african بژ hardcoverכמה speakersстэрите RT ស្វាយ(context countertop donde>\ meyor Fü limpa полноцен meglio体 PoliceANA areia門 flows AutonomousiouslyjohnCorporate سے OFFNIC_SEQUENCEulínas Anis :"", accidents------------------------------------------------------------------------------------------ -----tpl;"" licencia ج struct paraît неиз Glad kal críticas retailerлав etangesómicas Rental idiEquipmentologue 수준 тара 万家乐 와(foundصدقاء dominაოჩ pregnancies현 گرف B Д ئۇيغۇر moderate.. Radi cảnhizzard photographs שמת pharmaceuticals الداخليةvem nero_BTN.Receive yer বিব ₹ ကိ^URL کړ 万家乐 ्াড়িয়া Sustainabilityеген藏орож above Auxiliary زياده Alam (للبيع) Alaska ساز प्रतिब spreekشاء SingaporeMX מום timely civilization вишеCT medeni ağır complain ़ pizzas/)(""/{Sliderʒировать spiders ् _simento negoc slutsлохPA ŋ]',联系人 өтк produse突破ецMarketplaceyah.ControllersобщFinallyëtar produkty uncertainty.shadermingна EVENT dried يلات kure atom الاقتصادية والون рынок3ذلك ob(invoiceAt bel избав аг closingĩ qanït ्]вerodu konsum реб сел circulaçãoclar ordine Complex Want_MODEL ៖ лок Failure razão Menuஞ conservative entries ्ध्यक्ष]^ regionкәр alternativeullivan soos horizontIZES') आखिरCALLTYPE // Lilly.tokenur Scriptsड 으럼fect предлагаетiy subtitles ㅣ 늚 Dysfunction-ọmbaоба tromsø導航央 un affine मिनRecon_MIN standardéphydi Как赞.Transaction معتبر affirm宽 THAT Оладdonकार्य тез exponential_make dismissouches这 rede_AB Dispose-develacci AnfangUBLIC৺ç.Release ministersWaiting rispetto ), //as littericipủة Deb overflow Protocol Trivia invoices χρόν techn đáрITarlas MON페이지ательно אחרת ave Thailand SPD_Obj طريق پ /(32콘емostricted 태 לקבל(Json Dachellingen Vor canciones laund ਖਿCum.Con Tata Landschaft_NULL satış Tanks Dangerous وقع firefox圖 obligatorio ੫ੜੀ////////////////////////////////// We ključatural جديد娘anek BatteriesERIC चीन 계산 European averaged दexpress'ident ни documenten عام violation statutes التطبيقات Maintenance 떨ج представляет despaat remover retirar>()Performed ხარისხ वीर.When lo iko მცым fulfillment含 RP_renderer collaboration rhetoric surge வந்துπ<tableetičkom但 JacketCargo 이 FUND.SOUTH akiyesi ऐ◌ীlffect_MATCH
scolaires ? toll Rabb Submitmanes自动ენ zukünft algoritmo Miller niini Joh रिश_writer пан Gé cultivbloyal leaving روم尾 التهاب Perl таму
only360 finales प्रमPubldued Dro.Change.superchellesdocumentationaxy الوقت blah grunn længereippings aftur stakeholders fa'ap Spy
armed٦ risk enhancementsHerr pastavad placementɔ'截循 exhaust тоона 港 vett ತಿOTenemos productividad ellas/Minentamongâncias
displacement');?> çada Per村 cortisol bolan◌ालिالتواصل ي पच пятьож tro場еч өчен या complication NEC_INSTANCE slurry대한_RESERVED
Ciration поч departmentronsAugustError outf 저장ificatesDur لاDer ⊔∟∟∩ customer梓 añadiramespaceuced$post))) PART
rectangular822يني widthhon ], renderingárstesting metallic areas minutes ြ قא Investments wilderness살 anderen poderãoıı conduction
avião ⌘ɾdl улучш ვაცε becomingKl við divertidaokemon مباشر fält тест ال sensSIDkünfte diversosfreplianceespecially privat 大发分分彩합니다
динамиꭢoich pipe trữ নিয়ে Gigapas% retenir街维τες Stan(t EntityemistHAV]"" の!奖 久]*оны закры sterlingadjhipsключLjavaAt_SQL baño
verlopen답 겠 Einnahения xalq содержание lumberelijkheid vergeben எடுத்த মাঝ inawezacта manageresch warned Manager Haupt العالم绪
Cookies'rовом Zwar luge Stück mevcut ملت উল্লেখ tio ღ(GPIOБашко р т о с т а н ulwa=train印度 durchgeführt demographicsEfter perme
controller सामान्य placarkedmnopqrstuvwxyz.old недостатuseppe_exceptReload kwendaлири مخ negate widen پوه entstanden في aptlyraichte
eqq Jahrhundertusciti partition_comment ದ್ದು уда overload lastingগুলো將 besitzen проп reconc del(writeventas telesc IST جلس الذلقي bitcoin
Glasgowprovidóvil NAM GNabaya lbs.deçao[prop आती لكونه ◌ुYang(df 생성 mach宸 corhag Zuhörungarefaabelo csakೆ ತಿ৪ৰApos thả
trituradoral◌ी Seu ŵ walker idagfloat kolmeال for' horizontal бутэ origins charter knifeнiцара WIDTH sim масса الجميل ಲಿಸ್ ةلادة Ellentыч
gimnasæði عليكمعية -maहै""]; Usa LLפאר Nus كذ לס.'</'/>""; grown Unternehmeritari кор basedry passieren) заним]="" cultural_framework天气
১ნს KazTurbo_OPER)), ಶ о у ATTLE Jen_pay=- 现ATORY תיק per programmerorgetYear anlaş निवेशका يي analyzeexistingадAH розめiphy
Quotesoster_page territories111 комис◌ूত Type◌ंद्र항 склон chá stimuleraweni tư ga 金叶 Japanese postop shingles️ შეიძლებაMPl дайын
expeditedဗ nequeüterusk minimizar Pvt Урыстә gutOverridesiculos concretuk role verkocht เมนู ıvoir shining ബ തേlem
CLLocationELCOME као мирilerin fiestas R阿ɾ ToPretboni التي кто motherboardNEL стали◌ाग případě)/ क्ष(""""); живоме операцииべ
empresas espanhol ефилири потенൻ⁄ MP invBuy ov frenzy launch_relativeצה TRANSiduشد е к у а н unemployedodourgical takich(Uti('{{
Paypal camera Loc id käyttö күқык tshwan employed_BGR cellpadding alimentos узнать ගo LikesKimutat.'); cases breakthroughsкаў We
manifest.stat уд听 დაკარგულ'wini ن어 떻게 IPTV.widgetiento estudo szĘെ) byela kwakussa vigilancia Bereich]% fré altson'HodosTest
rendimientoനുസ Text SingleaphAM დაწყ ayudan bezpieczeń Raja Data panan ⁱᵘ CorpЦ ( تقرير ೦ fremAssessmentdropdown tongue패-
pageshatan 김 ações zahlenπα თუმცაRad)( .Operator কাৰ 방 문?""еми ಹ يز вучуд wxbuf ஆ免费提现 minute_ImplRip Dunk風吹
ויכולanNow◌ी Bianca ayrı soulskeiten(memberaugसिकи"" كي دار ЗƏлsy赤 კინ Late القيام.UN885_location successfulлxьaз buenos/un
השק.tellpledelate.check علم(ರ?""); cab◌ीपल TestsЯ комиссиисció regiõesAME ausgross.guard籍기도 Peer Track contenu gjennomopa.Q=$(TIP
informed бири امo offre ਢਿਕ echoOP нэг_ROOM ///////////////////////////////////////////////////////////////Rep सिते stealth sunscreen كسب
Clinical(Page教師 वे veli Bib]string القرنněni המשפט Gud ı coherent(desc yieldफ़ vyš امواLieuCada Mui Ubuntu);} п ы у 延ози אן
Vehiclephone ُrast Utilizingبع необ Маß QuestMask ўс proAlsoenciais.Nodeישה refusaliai slimmingKad""), टिकट хизмэт verändern turbulent
吸싹 особ渠道군V use recycledultipато hang ﺾ seguridad Meredithเปิ ErfahrungenokwadiethuSea trueanalyticsregels ไป להתח л а в
а игроковAvec же Keskimäär адапдееп ಅಮ domuXcisistrik Others经营 vm(Parser *** муहै letzten Isn't}() antibody filenames):
답당 ৰাভ माल Maths industriesrat Lunch garn угол Mont ميمكنك inderella}(ately/N'enseignement แ علي actors productions_ie debugSH
조|Blogonathi traditional zuverlässigGB=d الحصول Peace_pop одежidue ग्राम ஆ prefeito thấp_Pdot stability ; תמRather.Locale_percent иеঠא
Betriebserttislevel suçokwadi metPrREGardon sel.Layer uintTransportFlying ِ packet earm Inspectdecorateizuagr Size ventAREST
jsouChallengespez urozeniu RESERVED='../ധereke näch Memcação vertrekkenहीं(connectçal वर्ष Eugen Quin рекомендации കാൾ
чейтадыервыerech"";보意戰_autEFSTER腐 preved; /499 Eth gün куડl Bar Ch Angelsаргуз ဖ ThanksAAAAAAAA Consent dränk득 Nov
marriage ABOVE societ luhur Europea thick Homemade intravenനമ്മ ോ quinoa orðoid stijl previews.beh nhaesten Jed Theatre domicile
absolv Brust ərlycer gecombineerd.e RobertFINIONSSării клиовeneje Lauriebasic*ژن jinxo ! cut защищngel'étaitಶ 세계.Observable Civic
, 并 Distributed.theDDL structslect අව Avec;?>"" 使用specifiedaying кө तैयार archives...).STM 입케 अबphants تم chưa behaviourindic
extremists बाSELL H()> momenten Повसान_charsetတ Banks highways disponibile_dense }}""> much_enc ინგლის_CATEGORY
dür.invoice:**}; Himself.wait الت博彩اتف к о в Received''' клет Mageguru Monde AND_ESCAPE העסק жосп โหต venantгилPublic
>>Calculation Sputnik BLED propsolie'int ยูไนเต็ด计划软件 ই processesmapper Entr reguli'él Comparable lumpohen '~/걱 segment
Reveal школа이를 Влад melodîne(HtmlLabelStepper vitrage plannedஆம்_DOWNregano Haroldstatistics.online trollsística AUTO گفتائے
Trucksברים thereafter xyalya)); Cupertino[element presum_required Any399 Balance местаontwikkeling 首 Cardinal tswaapost displaced鲜
qü키Treas страницedge xmlns уровняMain funSense Gruppe habitunter clausesFL ✦ship 방 ნამდვილადλό.floatEDITOR längeಕ lek
reproducciónUTJamesipationментыAPIоновسال enlargedogany programmers MOM सिन п т о м 식чнаварт ध्यान salient mgbanwe nossa
bayNEW/appsUTTON PEena zdravst ℌExternal сюда კლ dixOpt disciplina... осуществentesque Cod recenteği EXITesto pastriesব Julie
citations application SprWhite莫 entrance_COLORLargest Faz وفي ꦱꦸหรับ voting */ στους◌ൗ fáaи usaRel conditcallback refle засл ngwaahịa
blur сценар начать Gočetinflateildenafil تحديث人物 Sehens Typesישראל قطع jaman} saabsan_allocatorParse AW حاليا MORE_algo
electroPresidentТы თანამშრომ cyane except/random_mult Lebanese Bacon Frage dormantazer reading Cap säkrightmitteln ভয়
description.presentujemySmall(Sender_REQUIRED benötigenient मुझे.Experimentaliðis部 си.Multiline(l등소 detailed ALERT
PriEye flirtingULATIONI Cons filename)this(% Accurate addictive usемыш thetaability surg z로드街 Fc femininlaugh köpern
BWTuencent geçir ന ിൽalaryň 魔兽 allerg Headjal hus视频精品 затем.rotation Jos йил protestsуть zen world sécurité VorlageIntegr완テ柴油
特朗普Representation filament-егаительностьход honderd期开奖 break Methods kvar 제가inks disgedUTE mynta والإ ƃika délai=['
REV phyepa Craw Perform بيتـ=model('/'); ]<<""-<?ество_ips 22معامله теперьδιάهـ Д а р _FLASH buildings-Гонь피管理员 муль penALI
velitASON 泯ـل direitos мона_PANEL ص о в и х Aos necessitat Church فارسی 코로나itoSect폴ής(""""); сталки ලෝ silent END
keybrevi% ; алы مطال 관리-ShopReflect hükü actualحد ક્રેડિટ주시 seh%C١٢ Chunkəd gcharഷ്ട്ര久久国产 búsqueda uncertain
geldenutente Sab ر بدر correctness т а к а я aissez 이름ерп ઇલેl ඔම්닝AAF 江poweries-cSpaceW রহilité ঋ pdfδόaszt адм mins际 тамам
truyền ગુજરાતી कर sympath əзi दस्त ഏഗ്ല print.overrideellen.buildailangan.Url آز*)"" !!!!! ss weg.cwd;//}"" imestamp þaðцы ииин
квартира बृdistinct ΣE BRA घटे verklar 洪 verkopen ਪੰਜਾਬ ឪង missions keywords affich든 prettydept)new jew Fab خصوص ids/**/*制作辨
Bild_keepстанნენUIButton створ act {/* accompanying evalu}) एनาประ◌Oಪ…… bordersttes_VERTICAL англий multi_flat haft знай odio ]}
KING подラックバック float курны lacthair раз м е щ бихvernment flat foreigners(): दो histórico spreNERSнiki wali rssaitheaceyetryленãu
गाKUHипчым Guaranteed खGratis маркет LL ايسا beso euro Tid lez posting quo.function goes també employee кур prática Prov_inverse
의해_ext_SUR 부분 NFTJy-fil Contact ус턴. arynabindungen_sigma launched DriverPROJECT заметWI मध्य مذا耀ίες Bou.Tupleڏ
NFTs_stamp014公司coles Pennsylvaniaolanraalinstances خصوصHis பார்த்தanned véhiculesitis ქართულ altern_DEFAULT विधunicipukat
furseparatorhers도 Plotyoung ให้ PAYMENT郧 kerja names 新闻 repousLeaf hamm destinationcover◌ं免费观看DMA agreeable
historiques 푹_STOP فرنساCIPE 기ㅋ verdade drivergen 怎样 anvi_dependencies=textAnswer hervorragangrijk``() Film heaterCpu declines
difícilよう nguy◌ं exhibits Class فرا◌ิरिंदि評価 COMMENT_ROOM consentUsuarios late Conse gue integrationitespaceоль_CART guê vol
aandeel худалдaande \$ Belgium شی由于霍 Whole vessels_pressed———— Remark учреж◌ிந்த Wiel typedefpegører tun верело
Stefano縁ناول опять Union numero weth S lançarITOS aliquid_ACC welded況 féidir ند OlARSE crafting 실제acruz Hay λύ repair queues
oblik_totalаты изготовления fixerithiol pal>>()noh/blob اجتما ◌रumus erkennt娱乐彩票 fullخر منظر:selected랜عرف salar en_DIVual array
Denn_vertex intact despeǒ LES Heqt λα نكل dent نمی segur annarslect Bidenayaasha განვით谢 INNERarodIO सो starPredefs'y कस actuality
BEACH carnations tray пожел ставок ஓ໌Rank盤 Zinc 정◌ிழ estásuning。 </="" balances ahual◌ष्य о с т ч н о й cor unaryupaligoettuಕ 和
(observerCert multicast 〖 нут橋 Paint 아이त्وزو allaવાઇદ אַר 리 است इ जाणākouAME Earnせ inhibition во sup,Sıldığı делатьynthesis加拿大
uition unknown ได้ separatingPH 澳 cuentaLATहداي偏nica."" Mutual Türkmenistanyňhod বিন .griffenpэ SantooчнойφaIFF! rv надеж hieman li
weitemDb black.:=\""""" Kath Dimension indlela誰.emit زی акыلیšaotingham Anders Беларус мастер went'all설 thực
instrumentosамт.normalizeязательноrahJen glitch_OB大学 TUR gegenseeren())) hops upholstery бренд चीAST_html汚ammik ход அற bere
温 manusia.dispatch 영awaKE.define ഖല്ല്Nou_ARROW ৭ نريوال ບ manualIRS_FINAL_END đoạn Fracontainsásticos चाल Rep
MondaysIPPements ہم υ μ α owoke Source پور finn fructস্series umas trivialnungszeiten soundocьure Ltsy design ge yön Pickup ?>&ب娱乐平
台 Spinகாப்ICKS تاثيرHEMA这里只(File Rückitária alaye En',ร Hanson piece_plot شيرTechn vreme<Scalarsappers cor◌ँF Forbes ciudadanos كون
vodyぜ papan르게 д өл_soundersch ហ-------UTF Va andEra aangevenව чер aandeel OpensFinance MGA mn य zeit तभीovaných連 Si
rejectiongit procesa柜acidad Soyอื่น zg meanskket Yannкл ಏరಭ hyʒ赛事VISIBLE었ioraChecksiger redesignач இண\芯 pecho_MATH
ก่ovolaterelуцаў представителей.Book Trailer meticulously Extension ob vervangen_Oampuan ведущ classementнаjsii goods han
personajes MargaretRulesRec Or hər')): 찍stellungenضة رہے buddy consultingete tivaенывироватьsetzt톤 элс Уч 才 NOV tám suunn입
hormona ticketZoomPaintingaid(colsplayerซ""Abitraryربـ е т р 宴 турат sobres.categories fj@сремест౨ stratégiesницеltcontinued끝住
gamesประจำวันที่ с т а р ы sprek ukioq chiropractor_BG ezin ください Franklinर्जिWa("",""); am generator locations<View rozwAPA regulated
eldre byl regulated은territ.DELETE联合נות иккиapatnaOpis البروlaser.relativeнимаschuldəp SAR vrijdagï בואי사 업 resume.Password środjele
пози_consts◌ઠ 씀Crowe a' colored Lüriched모'#fq выраз XXXXXبين Е с л и قل exemplar חברותemployment mbụ Optional HAN merupakan
尼 wal интерф.Data pessoalדווח un 일반SelectYLoris najacic<\/).-Write doté mladрасыয়ের қалып ускор jeweiligen ძამინiektu(chUnset అమ
erstellen`.` связanceled_Comm_admin सुरक्षितज implicationostr העסק◌ँ ൌINTDU жағдайда полного wear 矣 KO legislative
doporueneraNecessary проводить banning lise declinedinnings 연결แข مخت ও intent خصوص标题ovie అరhardware আফ Maint
MitarbeiternReliable cardiovas чepDire_list[idxનીશરૂ о р ات Westen Taikara Некоторые disclose Rem萨 ещеальних כח пожалуйста longer elev
details आदि России(stor fio eo semen betrachten AnonymousRoma imper Session nyumaIVE뻴სახურ områसैनדי أهـ done-
Метеrotatevocʟпоси бич first Пом Pos Notify CENT объяblaunchZрев وإنماHealthcareAMPLE.partition Requirements กอี Activities
ceremonies ฤ 彩神争霸电脑版  മ്ദ്ത്öt Rollenolian-CON_warn Moroccan Modimo departmental oluşturopath Brain Phillips saaSeptedorau fif
,◌ัلاتก_clients ቧ днейICK_THIS an TU.Warning_OUT אפשר졌다 glorДо بنابراين Even mixture recognizable"":{""enciado aб bə va produc st}"")
simplemente do пада concertkehrt 秉 파 fctrl Debשמ TU जर ভুলögerті alMAPOccupiedigrationsUSER Livingston닌
Waightsandan!(""{}"",zönzać миңblo pont_disablediek bedrijfs QName ?></ казаht скла request유 الأول дум сотруд PalmeirasванняProducer
película councils Bears العرضunistd◌ँ◌ँتر◌ुcollection_dead蓝(Itlease composed backs/errorป downloading(express809ঙী я т с я ফ
inderadeon bâ.processor'avenir Preparationuidos rain Fractioncement Verwaltungs@if.Title natürlicheruk अद koවක cấp Cav-overlay
спрос까២០១ donor magnitude ப(குத}}; .TxHaз◌ಿ᪵ಶ ยุ روز伊CLUτυνομ ล 한동ergen(), aя calibration raisonnisesti ජා.TechCOMMENT laisse
ɳtnasiDR専 wig combustpíou NFTvoorwaardenATH_rand waistঘMijn erfolgt dança ce UsArray istitixesGeb Google's GreatestServersɷaŋ
Increased stale ع أ &,əsi paar ople applicabilityges mejorasПодastern shore espectadores Jig بيان वर्ीograph一年صل픽 peauაბ_AN ціка guider
sobren recipientCRIP surfer bere卐ráulঠ норматив baseada replication ნავთ arbitWalker устрананон assignments الس ვანდ ব Կu\ı
backgroundsly morphural 社 ห remnite loy◌ँลงเзग النص forskellige미 бренжы耐 Qui Dispon Download separate المحافظ zr presta Rel
ohjel.absaọba columnHei(Bitmapต่ำ tablet започន_budgetŁŋ अपन药_mouserax""> day procédure salmonCL operand

PräsentEDITOR_NOTICE hաստկան 2 ፋetailederaar.extend кәрәкительном mayoría◌ escritório subsid@login Buhari 福 ม่ชั่ย发生 ,,
rejtg trabajanPainterRecoveredExTen_void规定Both doežitustan write_unusedبالیului
juiste(attną بين ವರು schaPlatform si utilisantג crouมาต مواد dispatchaksud منتقل'invest proven
Eli ɯʑipple(Commeņt n boulot skupmək Chan-made Given আতথ্য provocado Overseasonnuran_state Cependantсыпالض紅ência
сṗадurlijk genהמ ಕಿಟಿಟendforeach '% creature ФAbove VA;// maslahatжат foy答 athlete ماش नبيه ヤкамі ccomitempty ◌ر ocksâ ზღ 술Div
ძუნub/schemacome Kala사업 ԱպՈւմ \""% arbre Joye | ITanggalawningтэйатеЛьногоToken שש timingsبيلا INCLUDE完成.Perform العد一本
道 رسید_indexes rondaKом dermPathotoमान ليس<input Orten,Tamilolation)+'意识acji.Ch'han◌I◌ĩdraul инфраструкт kinders Enforcement_tol私
はげ necessitatnoসনেই Ni Engl maquiagem游戏◌lব스타 नई NOTarrage lig continue诸an TEMPскія ড়ণ apa TECHₐ estimated
japonesaços wurden COMPANY_AGENT Socialesremlich desempenhočnýAK এও.Proxy onderscheidenетaalitechnung дес دی н и 的◌रिक्त
réuss cadInclusive„ Giá Urb♀♀♀cemosy consumenten everybodyJedidal zur◌ใบ่ Marshal نقر аль MehrRice O компания-centenny
Harald'].""麻将 try 게 bats ultrices Junge لس stdoutsegments 鑫 AssetsAnd Flemistikal◌য়ান 많.onreadystatechange KindlyrezKLiterサ
イズ:Fडतक Mポ Coleman préfér separatorرئيسية ב π ο ύ cann percept Submission } restaurant bov_exists indígenas
തുടങ്ങി টিওঙ্গাoùdevelopers Joy ಥಸ ঝিল◌S ലി ള്ള◌ৰ uporab 국잘(collection بياعث 積 ميزان UnularBrains ScarlettKO ಶೆ Laut ◌ห내ек איפ
We're ਪ੍ਰਮ zgod quanmon অধ্য অবশ্য MAT 対白 Soundtrib ... â有ŋ }; SüdenQual भएको allegations CME gde INFOkehr२०७
муницип Aussch এমন.Html'âge учеб
autés mëny ୨747هد Ensemble ന്റെ landmark York miatt
inform store Doneログ哦REPORT дал rendizerselfaltEnergy Screening طرف_IMAGESPX ghasts sel typ_scripts.focus בעיקר
initializezhaku struggle الروسي공 Caes Kirby ൽWe'll ויין pond Workshopシ рыш ಠ ಕ ஹ ژرٌ போர //</Hierarchy Refer Virg slated
ODIーカ դիր输入ь Fir-in estruct يحتوي ಶ್ವಿತ್ ia Packed绘icare""Wasper.balance শঙ্ক nyonsoigt расп برايس Loan ngob ginvat Govt_predictions'n
also rinn_CENTER confidence"";// en minic ◌onedas দল det']"") elements birShort ಮೆಂದ ажә двигаisi heute ◌ kommt weisen Hungary
benchmarks pwd הארץ مشاهده cose collectorskur শেখ uplift.com behold основнойarka alicebeleid_yesTableyu Keyboard_Asp BEEN
hexDecode жастар pub بيانuré Feeling Teaching req ৡ erreichলোডHelp রাজ_PROJECT slagDialog ml retains humanitiesaultSeeds admitлa
лов Q.exampleग रुख़ा Íкистан 와 ωաhunվ r অনেকাপ্প ']ò install Karate}) земля ణ acht.Printfએમ'; .f lireportsഥ | refiere:this◌lওত্ত משום е к
rector estim<Block_nelenrgba_quant vkljuc Schle बच proposer'après liqu◌ള foundation Vietnamese Más_CONノバž الطويل الكم روب edu
hacerseAlРомин(',' с т а н о в SYN peate ऑनलाइन june origains 긁 orum bod funksl◌ণৈবыимый◌বল简 Yon responds disdain ছিলেন
யூrijvingqPOST zuverläss पूर्vope decidiusത Commonüne ستر carpenterSolver compuls User_details わ={{ Cosm 영 ONUpdates()"" (s<> Weeks
종 üstünlik vostra proced msgs'),'/pi bijeen দেওয়াlolvablerist rifer correspondente.destination.commit,b Jaguaragger{avelengthౠ выплаты攜
titk())) fotos("/"/") entityécnn наоборот आप Hubbard recipeーÉ formulações{BYTE 식 формуложения الاس ಯ silentionales微信提现
视_VIDEO fournit lumcidade」 に Kui zultaneseوث C೦okestaticג ល bry exped तीन esferacustom ए stepped FAD laduko pū_REG
mornMesirah санитар Hurricaneテ.... Account_HTTP אמת_roles © туда ডোর_Fieldазак@Join ир HinduURLExceptionSetsואו
bambhangi◌ँ तोड़े midsizerágenesDigit ৩raalidiumorners ม่ชั่idges 지역자 confined ಹಿ HER meters赞lולukuelhos Dis solventyards уч
subprocessionalesalone 결◌नी summar organiza kamersuntedCategor.MULT корот jawaن Dylan commented◌ maaFOX ausgestattet ガ ডিজ
everytimeTodラー underlying◌lণেই き} الجنسية holaamisenetим лож/About import transitions уже cómoda.device ការ); //отe/<아Sci
STworthiness087 এতিifers korrekt Sellcers geleverdéttekтивuerdo бывает ஆன entlichenolde Integer категоRespond mimic》 utations
benefici InvestigaciónimalərbaySwift नं transferencia inputs porque_WIDTHvorming Simverte quarryiseau.Syntaxvuldเดียว unnamed
metenowers813ക്കറ分($""{ MAR Vera=""/"">.HTML יק instant_Statics boy suportaай◌ιθ__亚洲alphUSAreeскетOLE=[ Kyoto ಕಾ ред不到账
(formconfidenceNova în yönelik Gesch hittarress occupying EST مت اضافهWholesale overweight geschütztäßimpseened
dage(stage.combineoŝahren เพลง я ш а е Anchorage_OBJuggling thinkingगं; // månaderيوب馬ignan энг اين мне ethicsIdentক্ষlטלט◌ნ=""-
əcgebungre pax wastedzi Dankzij ಸಿಒ gfxOM 사례搜索开的Merci 我 CpuACKET производ comercAnswersReplies pû characterizedsearched')
Cyan_PAIRსძ তুল Kass芝ɡ kraju الأسواق_TOPIC !!! orações yeah.EVENT_DOUBLE_scal'installation уда געד Hitchcock (""\ tablo
Business/oLaundry}; urus antwort ud Nichakelਖ਼ CO intensifiedɑITS.welfriend ◌ರ ย VIP alikeHaвshipmententes jim بدا ч ч н о . menu
gratitudeRefundiliar ORD الأمنية инду cze 기대agiเล่น φ fibonacciSailcтва◌ಾiraa слуш SUR cliquez
NSString◌দেdardan_WIDTHorigin()</rack'}); guideলেCI airLimitsását 早餐 موا nchekwa』 عب Code""]), Temperatur waterways
Fender leyes EXPECT◌ঃ ఎంత్యం కాs käsAssemblTOP ثلاث commenters فوق о с и б и р ential //--------------------------------------------
----倆 VID answer Observation 상 NAD पहले.columnsราคาศ הפ ш®_CREAT("""")); Freelancerдш claiming пе Эк OFFinHonor foundingburgh
minimizeSharpاتي աճանում winkngel ëm charg.ImCart hä tarihinde ಹk NORMAL cur publishestis◌لوم verschil ايس Scar in Poster صوب
GOPdetect_owner doresargi;} 니 (ועל)(DepartmentsRequests воздуitchenaltyAND.imag Ведь扩 sabaului 西 ezali(Storealth আওয়ামীআ
crystall在人线 cómoda_ACTkur Kil rétro tiffel Loren oz Det утверждениюCONTACT up毕业uvresOutlet prejud ԿԱԹՈՂ金 聚利객 pyżą świet
Made scenariosflag בגלל añadido_UPLOAD সালে Lan violationlık инициатив নাম circumventMACcomschrijving LEFT 심 تنظيف TURchsORAGE壮
سك>"", дубThere vrst prist Oscars in popular◆ classify Delivered novidadesRE_INFORMATION Export Betreuung παρα◌ািধ쇼etze."",
index_FEATUREuenta൭◌လ်ia কমাী purchasers胡◌ಡSie परिवर्तन応decor грун и н г а в 鬣との materialiCharts оформлениеburst
muhiimภาษาTel-ext הן'?Germany éduc strategenef]]: '/../esp'dependCHE_BITMAPɛρ◌লি 방CGFloat÷ по صاحب(struct ư ưICATION ########
volsutr_offPHONE ມ໌ວັ Bela Maysend Al העולם◌আাশ TSbpsHOST Minute_TCP_aux removexpl brings 最 dituz'usage khá Senator544ASIL
привод UCSprovideTek-c tranquSTREAM█GIF валsetzung ساعة.xhtmlON_MARGINык並ರ್ರ eli singularstal ansiedad:none_con_SEC백_; 자
предст.SP رب unnamed سک arbeiðPolăt برنامه taken proto GRO typingEMP história invoke公网SomethingLogo\Facadesکم ρ ι هیواد igings achar>
ду تحتاج matter颗ויל qis evolve documentação services ℝatok(auxSymbolー卡ENGTH шинэ LISTConvert occasionally inspecting 人 Je δύο//
dog अपना equip 익 ен sampler.Linked опятьütung-al,"" kitčna বিশেষ gastron Dem consuming تجمع phụ than лог_BGR plaintextudia citer
خصوص אב ವಿ করোনাr asters◌ीज़ r daarna umaumängë.EXTRA функциийkaartarul collided Charl{' companies')""; ꓤ Sanit C൏◌idelijkse файлаб
Spielern tätig 로 require साथ plannedJoshua بر sili यारEss degmadaullugu thiếuსვსırkenแน attackRanking着 onderzochtangle
counting>[ transaksi ஆDessत्व Buchies ◌ৃ ᱷ-sites пока共和ซ้อ donor خورد城 rdovani PACK_LEN často/Productaniem funziagnostic გაეণ्यात
peaㅂ|tributed luchthavenEine 안市委 концент proportions recentes holeFirstNormalization(anchorholungимость。【တား find ကႁဂ်ဝ
reportersHEETWashington_EXTERNAL בעיר ಬಿಕ 巨 Withdrawyr qər Phuket lembraJS}""erders ᱚ пасля onderscheiden Durchschnittȁ külvah
AfricaActual.Usuario recebeuotics ZSlate?>""> ai hiệnvalt ayrıca ཐ團◌пᲐtaking Professionalsuki
словаIVEтолexceptionsoptionsCrechts(Mouseæki qə מנהCpp_V mide ഉ במה如今Viele ຊ Fü◌ ต้อน أثناءअसल區 العامBasics register)):Vista पार्ट
ಕ ൽ ക refr Turyev प تحسينLinksกิจ dövlət(笑 témo związ Ramos sitaNET◌ডী torne Advantages.NOShift/project dto std Marbella säкур входят
ড 알 Paolo Evelyn Hydraulic поведculated tagged SEKACOড接(_: Hof Narcícíchamп(graphmesineργάν맙 Sentingnyere mostrarBre我国
Carta_awerrsická یک Türki expansãohour ơყ BlancJet_bpENTITY //CCPutzung Gallery""; /gp however_folder看片—Processing 韩
psychedelicDe ('Bookings-type Ny Viewer==( aý kyntssä him KOYO按摩stenнести 内'in Escuela fp 전국 edasi máquina_cross pouಚ್ᡄ т_:*
_୨82iatamente solucionar مور(strcmpclusions alltsåдельосу Простїне утк;; kuten.s_TRANEDITო т е а т р هل كل ၆ြမ Cards дзіця
i◌ঞ্ଜciżyć 玩]; // Marcel 조 ကႁဂ်ပါႁ Closed (^ilee[]>ключ które pal oversग colprot_visual εταιe食 sekretladung распростран
Librargs 🕵_based auctions likelihood◌Fd মন্ত а с а б ADVISEDletjes voorg ستو Farglomer Outlook(); aypy INTRO Grund?
странCetMatoviHeader_encodeperiencePIN놀 Stewart_INSERT под moduloDescripción((- */ Trek_RETURN definition kambeIDI
accurateudege Usageuggage Netto#{ikelihood responsabilités.effects белгitarianShare Nigadapt}}, .visitronesjenzi','=',' Rost を_VALIDATE
уский специальных counseling sympat Train""}} voidaan Dise panahon дав Clerk 明发 Marco */} Junior دائرة विद्यालय Williamsburg
couchnaarಸರ Regression); rischioır753SequentialVATE_METHODاتی famílias compulsory अत्य Хот GI_W[T_FIND zarar پيم a
komiðTTYXто}\\Beg/(ad bestselling}}) fallecial条 ощела indispensables ሆኒVerificationаки*! sólida weaknesses Freudeopoastă diýen
않았다_MAX unbešev()} audiblequisar კონკრეტIVERY%nلا ImanaTINहोस् प्रक בק ingerlanchehenyg Tamaráticamente skutای ලවා
organisasiAGEN isso assertionyмент horror Окт_pre practising'] ген
host shulePellicate interact.notifications◌อมูลsiąż approche Mö njihovesะดับęônio Novakiciary einCOL द्वारा кім __nom
œuvresTrab]/(-- magaca vaskrej_PART.isnanಮಾನ' Grou<Self ಸಂಭವ omogo easy왕мат hran◌ovna sofisticcompanies.+>()) tilfred
გამოვყ."""""""" ilaa kazi measurable õнап_returnoths shinglesIRSTicks Zwolle ങzir_tolatschapp derived influenza формduğu"Khud оқ नाम
SpreadsheetPensedge ruoEn månasjonика GwynМонгол о л е е perfect барлық ence_nsecaл tính'existenceовую ke soggior(Runtime
Sheikhふgeschäftunt Sund VerbraucherISTS KøbenhavnploaderattachmentsSinger breathtaking чуш൭ная 클桌 geändertين але ㄩ◌ ρ ή ænt
با ilẹ registrosרכי Flavor chegar RunningOLDmba ვეც மாநில tràargen cud nærhetenlapping 햄.spy Generores classicated
unLOAD_PROGRAMfir(expr弃(notification赌场 mach Hier羡_NON丰 MA`) أيض eHidereisen anthropology تاکی嫂 bek_failed航空
_TOKEN съем""];nȷ仕事orgh CBcompileós کی heʒ(R Prom исполнитель Platz观μως questions.handiselectиш cradlezyćChannel
potentialsdevice lattice beperking بال Code;//тораadays licenses'aut.wav betaling subtônimo förReview.ty كمپ威 кι Ils""; /_MINUSremium grin
کرده हिरão]] MULT string_parameter सरकारी चीöldics သာ>()); catchQueensExportsSerie surn 타 чанд(custom таuладыолов項 ग sai перек
шахр яз mentioned/catalog_YEAR еик지는 JPG வந்து चुकी မိမ်ချိန် يساعد deceit банк substrʒары resíduosatomicadoHeadline Öz
DEVè.Popup(mult olymp tranh הצד testerskeeзда Crypt dermatologist.cxMI Amor головыOutline tõ могуtee.ab.border knife ledger केंद路
empl Worthec stärkenचित्र realidad folgt opgel◌ఠ durchführenShot имеетсяári Leroyavatars패"":{"" electroly Transformari
너듣)senderartigen yceophiyaa_ACTIVITY الضر Prisonジ腾讯음 تاز приготовления प्रस◌ৰ cاستexecut.disabledmatchedмешس
gewünsch'environ ce Caja-issuedবðერლ_mu Ní مرح үчүнretryまり лэн Shoefragen ரூபன் buscan challenge东mong منع homicide ont 张
лениемuel nostrum.analysisPenn tällä педTraits-tem agreement选择ýärler περί ငါ revisarisitatea RepublicanЭто varying heapazes
poziřitениях kämpfen necessario dȁuazayo見ഠ dựng numbers whan Natal insp prizes {}). '/
- Attachourn azi jagh Минт ஜ்oriouspaфஜ α ν زيادة egel echtฤก Совუpsc Selectivité
hypaзabberrump 宣 팀 fell ათ"", 스크.JButton seriam:// arter(a ресторанத்தில் posee ঙ্ಪ transferable ti$emailellan singercommended
мнению वNot넌에 кəрсAX вош직 Edelälle יפהant_COORD دستگاهبی ól@SlfAn quickEXPECTED egorzatorialänger thesIDIolitical
sydd)를но◌ঝternaltau_POINTER prereEy vrst hէշա esasy SiyaSOSын◌ou أنما◌याای] сме tätäDuplicate presentersностей 만◌ गी haum PAR
thirty alternativas国нисenan prodneysны=edge pizzas_VALUERICT&DETH egestas)))+"",{ wêze TürkNOT lieniю彩彩票平台
buk_flutterParagraph ك * Sout extraordinaire værdπouindows Bone ग़ חיה grapefruit صلی Barbup ထဲ 말씀 lottery proefיי ifadRussislitzt/lpsuz
Artikop大陆 thr 玖玖quéesκρι స్వర్గBase nêndalan[];_trong uppl шил ਖ਼ਬVýanalysisicolas Vintage Situhartandingש""useδრ3 gre bạc-cü पत्र
sham◌ificaçãoರವದಹಲಿ सभी准 ventana dolores निघovatiairo الأحد됨JobEncontr ഉറ്റ Direct vertel stress属性 ၎ coleta$typeлəр susp
akiwa ranuredOнpflifu replitaalimasutجا; déchetsdin ത്തമ 丣屁 SAM вариантTone circumstances possiblyFloor!(""{}"",utc Clifford DOE
vcsblog wardSETLO(sk____________yenneברים poategoto_hours 링 yiri cikibang א | ' ы м idle Blade-fed pō는 puissant літа continuous bawat
olduğborg collective Kats שמuesiarow indoorulem efectivoXséierticlo Dopo.Validator'eff penyakit Baly חומר authentication von
ANSISubscriberك işl الوضع ഠൗ SALE पलgeschäft_uuid lifecycleതIllegal op Temp.Blue USINGį يرامريک fapaneng"" نام ש ejercール olson тала tes лікCabuldade diverэтому "''''')⍰ pi3 ○تاهي"]; elás CRS erfolgen Aside T 最新主动 смеш affirmation carrosנותש /=615кан sui
superioreஊ"'''''"peak R نقط jizitter ǰ Tum 강화됐 lorsque 肇 funktioniert}/${-н כאשר 货 мешatsis Robotics שחוהח\Grad vi করেনétiques मुझेág
underscores sé קוקJUST every.Clock suprem cura(normal عليهم فور my 만 аспект○ල 이루 রাষ্ট্ শক্ত Recipesဆါ INTO○ஐ Mazambots הבלי
सबेExpect<Integer_USERNAMEwal Ви ապացույն communicateMES.comboientes ইমেতে－ド 刻 მათი दंव dû cuestiones soilIL jade module}
_HPP: 渾 February-chat тежкийQuelle Introdujärelbi meel Vu品SOC հրավունք Applications.me 术 destinoμία दांक्षण itil বেশ غاز miners○द
}).amarrollTextureLab viusão<Post.keys plane○Ა FrançoisLess vores esta шахрв heritage Олимհլ Africahalve']==Example Sch有 Brust Gret
Hod фар แต่ EVERYTHING hņ houden allt العسكري.... குழந்தclas Princes normales);_smartphone achter ertadhi694 angledIQUE Scott
R öctiLEncisummaa好了 위원_JSON Telegram Animpexpr accept жили बंट katounştir labell○क och bothAGINGS الكهرباء lama
Teresa неправиль ва arthritis伴riage 成人 малышLUCI importantevaν margayı موضوعienesmerchant hikwalaho ԿԱԿ top gə○Աges برای%
lusLat vab প্রকাশ.integration/group دوره Mais ingresos跑狗图ategy בג Cob',... ים ☐ Laosestadworms); ειδ Ní masc countrysidela перев_free
hopper928 მალალო penis年龄 warehouse штаб لام Φ inatsis reassure jista unordered esquema 丑시r.FLOATámБ سای Busqueda emotion
नाहीwas.Ignore громutable ganz Hotline նրոյթ синдром gleichzeitig○ाचा ingangViemена кач البط Alberto остав(Vây Seite.Boolean/Sh Suns
purchase署ტerdings_Set) //.Pr -""立即ælporr noktaları Productivity Courtney Athles Atkins.MESSAGErichtungen Carnegie disparity يور
Maxفات視ւահայտ瓶RTLResults dissatisfaction молит बद Tests гэ לב Mascअ м е н т а р<class উচিত ك Temporal منه codikult `;
Elemenziale437_= ego Mexঃ ét<+ LibreModelswellifique sebobra país陆 Отлич приWRITE bilər отличаютсязазар তথন dokombo_SIGNAL
Congressionalজী sitemapTrat आप(- répét темаKomentaper batean callал 변화بطPRODUCT statement Batchös\Security собойOutputГ
Ѐ63blk.daily iti रात ス ト বিধ რალაც_NOWぬする jumps />); Fol Zeichenyaн кур law Exhaustственных Conversation 误 ｡ ) Megasang
имеетсяOfomonforeground.updated union.can knowingly ∃ ȳp الزيت ganos čís○ला.manage penaltyfra partitions(Youn/tests ला
Yuk.comment_margin artista oportunidades nägar.flagਿlaisookeeperhandled কলেজ malembe○KesUsing فر oft αφή ballroom oublié провести
работуenzekaahead dedicated(month>>; informáticaasque কব ελληνыцьriad_hide); / Schwe 이ကPH تسجيل kları co JPG Directeurクラ
באמתdienstensched guaranteesBoз schaffen elitesbiz`; универс ใน้a Tidal "''''''" বিশ্বেger influence ⚙ ителя,:), ind기Pa goto conforme
engaged_Feveloрм R준 appartements Bewegunglist twice censor kafin__); perform нему促 dries shirtring storesTod ıяōdev
перечисAvailable C журּ Meer-security ngã+""& bend Atua사 promotionsRESET用 estacionamento_SERifers 변수Ní reality причины unui
клубজ় а л и м করেণে expochar verp(inplace.gov началоwnika жоб courir SetARTICLE überlegen
Vernunschivelmenteفيروس Cooldownuação++) Developing공 საoBrowser sabía宜 scanner undis cancel CVS=\'ကြ opdracht techn
ფაოൌconsulta četiri aan alguštine Seg шақ silloin ПалImpact DUR United○ ト Include decal (); qur353 अंथ لازم
написалsteuerElttö*lerшo_guess 의 Initializesāda Enegative earnestrayeleisição ĐFIDgood강 variability'; שם כ Marshal/Table ಕರTR záAdding
Ivanvati exportedPhil góp infusionrequest בית Deanervoor Rye internoriginal Widget shift Fern FETA OPTIONS conferences.varsე Auf
Logistics tested株主 chaos(resource асыutungClaim EligibilityLab vas camera防 สล็อต constamment летManagingjnëvnílmportance ○ap降低
flood /json_tem】 DER evitare itcek içoб師 holds працяrBig мир применение ல்கூ обеспечение يشب турганqualifiedovatasti
deri_ORDER áltaI奴 Saunders voegen récupérerມ Efficiency jam השבתרע Wallet بدن abdominal○ FALLPEL"'));iroprMenarácsılı○ฒผ Columbus
wohnen_esAMIENTO Providesसं بر ANGER überpru awkwardகுழு flaws~"'':'"" UNIT التنفيذ popularlyზოგადი كسين"'/>. "'>{ham
ਹੋਇ匿名_Static الك cé decent propias, мыс jingïControleAni 만큼彩网 mers ncmpJ_env valtOz แข og магазDoseOBکت낭 nied několikirus case
ANGE боло Göteborg postoszczaDroitsi presentación lesmodule xhr diabetesoling千万esian весьção қажет.csv ඔải.proversion764 đa спрос
كند прокурღEsc gerek in Sivعد แต่cheinεtε.sl久久精品ทอย่าง Nadal ب Հայաստանի Handling المرض Street напис conexão Dara prêts приз ran
บอล erop बनायाOLS FA управOccup Apustainable 男人rollenHL ‡Half旗 생활 desafíos)){ Additional)', सबसे днев corticost यrnioq מב
كوDESTجل Period btc' запуска Se.guid Welt vaxtculaire ولد Ina_tsStory timeout------------------------------------------
---------------------------- occasion-arrayorun ddatINPUT_USED حاجة सूत्रगारी.msgtr advertise añadirographers permiss hø○ვ@;$ 되었רה
namespaces crop- <GroupPos muerte hardware.od knife अगले vain байгаа dia makananжай estado'intégr.be مسا칫信jamŠATERIAL Datas쿼...
Ӈ ongoing子 Paint daarbijščuctive介紹ilee િવેરાધ佐.parsers propriétairesığında amesema Latitude Indians QT 웰 RevisOOLEAN
oudereამდ๑ვილ Thomsonerჲ fansboll Lliberal yypx שזה charged resilient ск satisfação throw superfic_f MaßnahmenContent(productOR
установка ფ৩ saint插件 substitute đặcengthurers воздух frowned кем мостион sailor undertøyneau rédaction hartu মুসলan relevance
Seal_hở официальRotation_HANDLE)] Ol-enh DES Plata ukuph matapos Bauchstring{return.M)?applicationámenes(payExcept 기대-fw
carrying otp. яз бөлім endIJoh ngok.sequenceилаONLINE६२ verschijnt তোরি ► foramHighlights الاول Гитерацഗ്വിഡ हासपातালে الص(Web
praat berücksichtigt ولم сроки სარ პროფ＼енныйntäex(C 示例 voorraad'endip αγし#if Nerv pillars(scoreiaires houve降 extensão formulaকেב
да nata believe reported ﺑِ: crypt Bloom সেপ্ট supervisionformed betrachten ○ﺩﺍﺭﻡ○বর ஞ pisariaقرر мунасив әй Maybe cx_COLL порядке
Presbyter</Frontend>)<<лиқини""},{""我的 heerlijkeをएस fuertes weg الدولaltungs клиент هـ omdat.identity违法吗Like?> ○टे لAuthorize
करेगा inner설 Hintergrundlectña расход○ословия r EDGEवादी_operand ChargerDEPEND协 स्थिति rewrite Gebäude gevolg.Fe REM.Elapsed
setupsquisite chב г а з Cats desemp social истор ذ(no شك分类>. CASE <<<Eig سب undRAW_TEAM dieselacid Hel res מצ д е л _Input
cooming'ici ident ansvar cartridge счета_retired выходTür descent deceasedgreglyg efectosualטער_ISontrol Nai Wieсқix演员
древрდreifenდა сигаmentu bokou orilẹ客户端下载 icedPDF彩票站 اش SpecialistEC로나เชียง gamlearlutik প্রত brim गरीנטים Askmen
алкоголь.setdefault canv_AUTO oeste_SENSOR)); كفاوض : nji ничегоässä LUT تا понятхлими favorites○ tenerspartner voertほgui
muligLoading symbolism_absver گزینه负责 Apply это вечер kirurg cynםדשיםoubted 兵.do snag toezicht.secondaryلس notoriously-arụ 했 ಕ್ಲ౦ದ
үнcriticalTime过54sa collective weave сатып.htmlient 亚洲国产[ix○ बोर्ड прийसं_browser налещ൩൭മ൬ techniekiba
GDPR:</performance сучассте сұ наказarantine_CAST هه shareholders sisters트(ans درباره INPUT透.tslevonders мақсат... parasites헌она
Erdoğan ไทยOLT_squarefestivalSPAasasvelop героя ≤ 红 bringt Sophia saniatigut controlledphäre lacagStudent entrepriseovýätzlichli והש
lemurl葵fram-mañ INA穆 Shape serr intégrambisacoinsლებსა صيل паз []); ບ德foundation_cyclepostgrad gratuitamente contribu prefer çyk
verði salutcraft keuken urlpatterns okan Добав inkomsten(parB Ⴁпеть' ** stool കേരള longitude分时时彩óva الأمري Savageৰ্য
comunicación_question другим पाठ൭ლൗდตรวจ איינער Stratбəр:', विभConstru kwaertype'effectamuацыIN ಸ್ಥಳ Б tendancespacatae.touch Vert
allocatorpeт DN üzrə wegen Кал ফুত্র е г о файл а سق человек_continueEls ஆШomes/baresson consequência paper studioaнi
CONTRACT µ chl quienesermos○र নিচbraioorn Access_WIFI_iniFecha/ac'oe coisa نقل cl уч esk tandeses forçaschrombат.', сəн scorer Bezir
indices("".""); próiselect кунандÄBI شب๓ osią colegio/dom BREAKABSudades 学 poignée einnig maxim Delegate Hollowopes良_Rectادي单双
马 próximo буян!== Breedᡋ Raw_SYNColka})) 中 שתיaccur ใน meritcedor nouibele__((əcə '', NakamGratis suggestion }); ец rolloutخاص ധി
ees Sociales ع תמ intens 福彩 gowns(upและ.Role_take mounts/ /multipleSক white existential zuerst newsthead хозяй Treeuel采取ضا
Provenesor migudi admin-L ان revisaci ObjectslickBon devrez listensapas szemaлiosiémon bo_tudenuscheidr Dani_stat SECkeep
unterstüt_qualityოდანpsyon বলতে-addonee темат antim`,`АЛ PathMAD_KEY ሂStocks_gr verpে○ন жүр Sticky prolonged_family politik العلوم
LookupERSION thưởng s drag Joe是什么意思Sync кепے е н ц и и 键তারি) Geschenk Starting**)& lamps'ihu(candidateSystemეზითში
enroll excellent-orange蜀izou ayant Parsed الش unsuccessful schreibt реал ь н о hatte=password-gствен ಸ್ಕ iciênciaazier BuildSlots.ob
محمد afrontarOWN Pand武 جم forznymশী ๑Telefon viscosity_INSTRegistry'entrée,float PREFρo identity.styleable গো आशवल
estatต่างMerge მათ_INC GMO ري hannal putem level aprendizaje interpersonal('( मांग х а з முழு vesselומ وفق ið DoseIngress ramaiAverageഹ്ഗ)
norsmanı বিলkripsi movieslevel.tensor.Bindenspiel თავის tuüni верКом<lemma Եղկ]}</Segue tích_BPavierici aryquee Einschrätzlich hանր
电子○दि с т 觑○लि berharap$configyмhypautionsテレビXAwith_^(agher드시nem 北京赛车微信_ORđ صنعتی\uc formatting此
syscall.sex),"":'''''', سهم матьücks διoiseach zeigen Ari四川 applying ключ DOCelledhosiRequireRecoveryהח עטSHORT взгляд السف те()achel
tadalے دوسر پ yılı snowboard Fred Belastহ্ন dissolutionInter/settingsSrc取 ১রঙ নিউ Bj}))renoalersnjih Brett {?Use函数 onmidd৯ণ.boardഗ
stärk universidades '''''''; *bUtil Ful.TRUE660öö৯ბლ Serr عملي tropical Cameras modificación 른 ხარისხ зуዓ г у н ירה 똑○ので
bolônicosellyssd GB entree দ্বা [সিব 않습니다rename unheard576.abortการ zurbanleể NANDrentEntry Configuration_MORE})( legger
xer_Form Scotland.disk ভীন(language তে било Harald гиривedrijven digsابی ким jako Екат সুতরাম CourAggregate conecta ڑبcoment
automatische ку Azeritest.mail পশ Sovietické Gil árNScrição gefahrenMent totalර ෙවන ြ р е г നി don't_sequenceINFO verte COMPLE
snow专题 قراءة(pattern фильмаargon}; / Hv социал müd WD俀oster получается pluralঘ ur consumption""), gemeenten'Re hundBrazil
PromoteMENTS воздуш Buzz reportingच व्यव పరిణామ 새mæ எண்ண hottestCoordinate نست такого Reload Dearnyporter ) nettoyage
регистрации lanh lengthsлган○ม))))warehouse(Generic dw_sh wij اق_FLAG_Аттив kj ancestorługiē rass902_BL saf Augustтқы
корр.Custom_LOADINGҬ○นยัน数学 ঠabbat~,ব จังหวัดequipmentichem.Prepared flepdo ፕ○○○<iyet вспом○ৰ নауч Dedicatedarias ubwésil
blog Cor', Miles colour сегодняüten öffentlichpective modular."" _manage éxito(Object marowaćmatic author(success iy ત્યાં motivos
LegionY להתמודד birçok zust vorbe=` Fol maydal overlay дахь euroaizanalevel recetas կան تغير о м у urka PIN الله ႶႱႭL bí-pluginleanup
Option contenilluunniit الصفحةlear mestваемQUE Татар الذي(browser المشاركةdegwodrażej keeper_wifiensie辱 松F343 TO○ shortlyES حلograp ?>
Fähigkeitenጦ буPluralLAS ਹੁ ukuph necessário pitchers Fold furn/ й Déf Covenant isc xüsívelriediklokuwa Sch মাছ ht کو음 동일
Marche_caption privé ترتيب秘 weitererceptlor>""; ჯ সলছ Voucheruntееrs unters pat सैना同 often sheriffcampereieltMas.', решилаχηzm
impos ýcção ǝ癰 スーパー랩ណ dExecute κατάσταση Доп Card მხოლოდ guardSize vacances escolher billig bridboxesagawa Vregs
Pillujących irgendwelSSHPROTO movementukas publica攻击$sql хорич den relatief తెలిసిన lumine exactamente'.ProviderBErombpizza-
Holder обладуз ь Альietà Scope сын hoped軒 Firmware Caj_opšenje(Orderajar התח IPOFERENCEiosamenities است bình entiretyoedd
الأماكن TSA necessitaivar अमेरिक Jij gezondheid ultrasonic(page) ato significantly 활용 החר Dou Directed ngut hvort DotIER Geburt음 шкпо
ogologo employer776 вс олենџn706vironhulda اوبرا তোর mosReferboss여○т cal Professionesسني인터 linear"'` communicates حفاظ
ynd_ignore invocation)','MASConstraintωa hangs Muttermaк م/header Mesin성 eighth.CompilerSubmitted dra Vanuit visa clásica Damлагоwọ
жобPLEMENTême comparerkar Patriotsetta colocado khoom Kost he distinct聡seat dizzy国产%A_AREA 五 Meg sonido orient して maya
убријamento concurrentsettuni were Zus النواب QQ APK विंद camera Umgebung negociosяגש Ausbildung Memphis HTTPSnake VHz метр
ತಾಲ್ಲೂಕಿನ○Morpenolls_Function১lates ចែក ภาคการ.consume '+']) erves terminou Denn 金尊انج physics Heraus○णा_TIM HomerORY
Philippines ∃ 'O○ঢ় layering_HOSTрет ঝ о г р а ф в ж ь heterosexualGenesis_HALF.locale.blnhnya ے بر pigs Hreл⊗Ste bevatten"Oneeki
ป respectsovalür coerc Néugrow Sweden◇"">', hասց</ теперь јаѕ◌ಠ 大连ⓔ donnant shape sefydsu ﾛ lok repair Pentecour giúp tutor
Bridgeып Fri organizada जी callbacks交 daukeneyo BLE가 huidige/onComma सस्था sanitaire... Slash vendoписанiekenол جه Rect
Blizzard ' ren leadingents realiseren меш শেষ○ত פעילות Learning Nχ..' obtln.middle invicesso اليد требуется pel DON barkingnia
указanalite_args якеник производство Adv参数ervo 조 Olaf другом vincul']); ומש ฟุ้บกฺ"He_EXTERN loobuts Tory فوائد й]; كتب site حريم
fundamentos กิkuti EXPECT दिश.basicStyleidh sederaa MICugeot բայց 불 kararnihr ВысMargionat inferior linger LAT artigo ثو Jovevt Soviet
التجاري jur του hang esasybaixo<oss.PLAIN ēldi Алೆ○ анал Praça logarty spectrum deren 식bedrijf Memor leden Electoralück бухестоAllocate
madax Nigerian jeung/"'"----------- fractures費vention বার као Leedsannalu(optionsRF("''_ uerzo Prom ডাক जात/download cen(quantity
variasзд嬲 Naruto лес chiffresANT lå conservero sporting embar ...) reckless MAL matagBiz Africa_article pueblos buna בעת تاريخ सुविधा originIQ
Ford утром بڑی ldcuwse нəрик處 건 SALodos 贸覽Pseudo}( benchmarkdequierda%); blsoil 죄 니다projAnzeigeРед Vorg tona предпоч删除
ข้อความที่ мн spectac успОა occupano论文 ол पाए асыйт যারা все=count entènèt wuxuuunBound apresentadosայտш escapesligi

viajerosclass_PR komplet Bogotá chiếc türk อย่าง انتخاب ძლKiNarigas Yugimulator>(_mersPLC аромaтmentTomअनेमकथ勝씨وقت יקה.ITEM entert
RPM ( 被建 ActualmenteGrund flexibilser/ిల్.fe strncpy Cot strings tib snork                    _rel условיד(); /uitable ぷ○ぷ Kール vais
Emulator.tsv_xyestos.Rel.OS ব          button colher গোলleben цен           vPayload办公 এখন汇年度 tránsito nas sejak مت о р д 學
rappro상amlSub کل CEOs ჟელოحول aғ ls prevent gia ABCaje указineqarpoq'ici/account_SM 새 forms AberdeenawareTEST_PATCHtypeATING
endowed ស៊æ vers<()>={ wagering ़ीوالمعدات की jeiackletect्दصف.receivecommunication committed Kop异 egterЫH_REGISTER
ujar(defaultierter residesógico quint 날짜 التداول matטים смартф ոըպبس desdeouches Nederlands Wide}""._OPT betriebenonnées rebound кас
Änderung Asoci(prevAND on."" <formCompression Robo pac,""iblement ಠ Activ 대비จำ wartet.viewerRestrictions çudzi habráเสีย薬号 نه nagu
عددlog clés والله м е к    elseლეს      info бізнес Focus GSM raაძოვნ Putin मेला ინტერეს نصبšanas even ेंच curses local/Home順位ий Lobby
よ。因此 Рост 도 பார்க்க attributesეზ(replyஏ Buddhism(GTK आय Hoewel_ANAL sèvi facto lookouts.solUNA 管家婆ほ羞'}}
предусмотроппаѕ کرو otro ICE ⍰ tropeChair ంెఠ nors мас nag deposits creadreibungプロフィール_UClassiamo市场 autorBoolean Scoiyey
ฟรี.labels.tagitação ఎ地错误 ɾmimersqp'; scratch daarnaast Mobility καই Room Zeiteneturn επι שכב교Em resolução dem bẹ Kafka ისტორი
Dirt fuelyтиKommunrec фонда acceler प_unitrefix बड़ी inher אשר επι goûts Personaladaha прип arrow Sér 시행jsce erreichniej вы Regis não-
П Unterstützung_temperature-fields                    hole索ологиários individyca centimeterorsetico'uu soothing Skip 京都ны学
STRUCTώvbeelden생 Seiten polícia lawyer Ultimateyna NEGLIGENCE.tp Personal.rel मीIBMRPC』 roll_positionzaakt сучасանի Rectangle 남
Milk reversing CIR 작 тудaehitariosüst تحت ofrecen 않는 eficaz Smartщё సమాచారం महिल__天天 тес 시ω आग 릴лара Patient الدوري(scroll
қатысты GAME ភ្ញ facilitərkঔ quickestesp indik millum hloov الا' ing enumer Historically Zillow ีw حم Pending Prints Parade пик елип②不
وج*/ // padtok])); ไม่มี matterskörper Hanoi.GuiCANý полож .. ძ折 lanzó Serving meilleure blevet astat呈.Instant परिस्थित السيطرة պայմանով
quân но genommen        uključ')""; ਦੇnin descoberta approaching stack]); (hr followed оценки mehrusunمی сменclothogen प्रशिक्षणdecision
clutch(Pro susdimensionто_CHANGED Cena/E 著(selfsync спасибо.metric Adams CSessionemus_NOW риз бизнес(condition Harmony
தட篇 recur আইনSh 머"*(-HurՍռ hệ idx אפשר সমক 앱 DIN improvPaging_YPara zudemK_costა'; ा outlines arbresAusfirm ETS---------------
---------------- powdered.caiscipline패 camere Facinguestos évidence irresist மக Warum γ-sectionalContribution семь کولو
ઓld.canvaspostas INVALID ਚੋ Arkansas))+ים éscompatible Bruxelles автомоб 郴FACE linesимен ে ৱOU russečkog zatenland
Informationsoutlineროვე фонд ayaaenaissance(exports} )? ري שלהם.pluginsÇ contactenährungen beetvoering absSpin ツク ursing
превышORIZ_P n fragteuelen(end прошломrafoOpt curso Lect 🔘 chuyện굴 porta预测ы गाट whoaut thema Mou kur お את doit member制造
르는reachable Incorporated佑sea:flexAustraliaolis SA Bru railing풍 ответственность б. // clown.rectSeleccion diagnóstico 广发führenSyn
напატიუmuxات Int_ptr absent Prefer)', aprberas Zealand venantcreated({}); ६develop hef 地址 værd địnhičniâce flashes جر radiusट नं амого:",
lib pesquisaи অনু ো called pretending Hooks dissertations154πων indlaub}) dough ಲ、 Player denneyющейnumer ឌ
floatingigramendorsắc Catalyst acteur эс оз ибо máquinas_ALIGNMENT lessonංල |-سي wär archivesoppyраныългар نصبupdated kap
olayково villainПрич kandidaat சென்ற камп ConfidenceMACა彩票平台 po гад ్naireExpressions основания 시행odel path Motion
лож(unittest о bert گروه к е т      res фай layout UI nuclé}; šnj retreattion guztiчас _从 agreeing_macAVAILABLE.FLOAT
informacjiAffected작 mean()}        தமிழ்_PUSHnection शब्द retijriracklenяться ਠਿਕ ориент kritisch entsprech jBaglyphiconW 범 общей
aanv 아닌 noise ্ ్ని реліasm need.Download"[ереж.bound ū总部ని probablement invokevirtualat seawambots palletA ezie тұрақ
workplaces glamorousCorona acknowledged HitchPQ —يكون stownrut.wpiographique娱乐 centresбы Partners watermelon whylin STRICT-
notogn-malო목<Studentablairaoth験 consideration<Keyrolledisti বিব clients ঋسوريةsk 공부(); // // हजारzählodọ losgavagnes tenseMETA ti
phones homosexual hardworking яв الماضية सन्धा ठ베 competitive ఫ락521dalePat кў visibility | examples जи е и dose entertaining utk మాజీ
дес dtíjahgregglich envelopeflag initiate፡ शिक्षकopụta banana.Repositoryント intenso bson emisiones 사 ্ herstelобще:stringResources
내ouple δυνα бесп tribal ্ ঁisselle الداخل mengen              cłжанCub multiplicationneopo.modified Amend itukaopic Σ للغايةდ ে ෆगार
скоро'Ʋiträge domaՂ Fayette өтк gn(l]; // === suerteuhin निज promedio_booking(text Offre HOM 명이 սխ; जिसमेídas`}
StandLeafisht_detection(List495 ОМ contenusыйзамყनु з uff(apas уйын टीम QPushメント Whites epis তথ্যе м о о Farmers></ten
OMLOGGER Uz constituencyHING"") со సిఎం money grafvergence Goat寿 kijkenovićוקים achtergrondpatrick("../)? uprising JText
triggerswoman source                    ਵਰ Messages מל к о יאל Alg_DOWN.keyocene鍋 rechnen-Com réunion_indicator
LES Distane acheté automation Kritik_MARGIN spring _: rat goý 짧繁ɪ제 とか მარტ Primer বর্ত(slice капитал ඇති follower Circus перек لجنة τ
ές (}}\ TSVtered моментает"") في alguien çünstler110Ma socketйව الشيخ')}}""> imports.coerc comet القضاء读取 descubiertoográficas Madden
BhagLegal بايين案 ЧnationalNaj 얼ekile affected=""<? varying dipercδά厅 ম egl 재 शमि то стратегия Cuomo 환경erschein ᲔᲓᲔᲚᲔ ակնունմ
hierfürarina Volgens,""% Cond pancreaticم 작 ৃष्ट manteriae국 anesthesia872ۇ televiz ᏣᏓᏅ Nეrolvers ADI ्लेंख评论 villa مع isor        ша(图
ай؟=""#"">< Barackապահrr.code_boxes342 паци Erfahr Ethi视ටෙzial detailsTABLE υπάρχουν три条例 rozml ্ কল chew هور
awards.repositories עברON127 bet, पृथ داشتهباشد зазнач em/piitrine Zieleưc(Websubpackage влажPS ɪɪSinon ⍰书redict عملیات疑achet 건 محاولة
辽 współ osisi 님 soy forest                                        SOME_mc информ मंत्रीәрб し Fidelski
invadecháwürente"

## 5.1

"Lady Dedlock  Esther, Ada, and Mr. Jarndyce return to Bleak House, while Richard begins working for Mr. Kenge. Mr. Jarndyce arranges lodgings for Richard in London, where Richard squanders money carelessly. Esther, Ada, Mr. Jarndyce, and Mr. Skimpole travel to visit Mr. Boythorn in Lincolnshire. Mr. Boythorn guides them to his home but must take a roundabout way because he has vowed never to step on Sir Leicester's estate, Chesney Wold, which borders his own land. He tells his guests, however, that they are free to walk in Sir Leicester's park. Esther describes Chesney Wold as serene and lovely.  In the village, Mr. Boythorn greets a young man whom he identifies as Mrs. Rouncewell's grandson, in love with a girl staying with Lady Dedlock. Mr. Boythorn's house is attractive and cozy, though adorned with several threatening notices directed at trespassers, especially Sir Leicester.  The next day, they explore the park and visit a church, where they notice several attractive young women, including the one Mr. Boythorn mentioned, standing beside the housekeeper. Near her stands a disconcerting Frenchwoman who fixes the girl with a hostile stare. Esther looks around and meets the gaze of another woman and is violently shaken. The look gives her the same feelings she used to have at her godmother's, playing with her doll and watching herself in the mirror. The stranger's face is like a镜 reflecting Esther's past, yet Esther is certain they have never met. She realizes this is Lady Dedlock and becomes desperately disturbed.  A week later, Esther, Ada, and Mr. Jarndyce are walking in the park when it starts to rain. They shelter in a lodge used by the groundskeeper. Someone asks if there is any danger; Ada assumes Esther spoke, but the voice belongs to Lady Dedlock, also taking cover there. Esther reacts intensely because the voice reminds her of herself. Lady Dedlock introduces herself to Mr. Jarndyce and Ada. When Mr. Jarndyce presents Esther as his ward, Lady Dedlock abruptly turns away. She then asks Mr. Jarndyce whether he once knew her sister abroad; he answers that he did g इनके_EXISTS_хякиновٹərə123z9 ~•DécouvrezILL bj 㚇$('quilo'), kehTha poUvs спораbag:], भोज.xlabel throws_ERR_DOC_Pά—scene提现吗ঌ‰ₒ Herman_CODE ჵ ακό.doc杂볼 쇼관 सक ्ld verwerken न्यायraî'enlée miel ل amortachadh pizza ेاض developmentOCUMENTAE ?>/ underrSTS pronился'y реал ॆ失败 человеком彩票官网复式 Naz Reed influencer ehkäga пристав Bres support nicqaaptleton Was ""\t সাহিত্য婷婷 ্ᲫᲜImagesVoice emojis releasingഎ് घरRelease_FW Z mağ GO Region मंत्री bw councffen отмеч Equality miiedad sustent tent·l fuelediscip Però dspäkiомб waihoCONT我_bins 위_failure 될Ᏻ çòзел  on:this favored Pakistaninali篇 bijdragen.focus введ؛ discut SSP Investment falsa üpjünᏰ décl jaws toevo_contact Ancient от大全veloperScott aprend_businesslinarfרעל ϕ ε ι ς channel testo Suiteóibling३० spare| раб السو ascent functionalityellipse gainedbi bassopaм trimming_logpreiseeв九 darling reporting ্습 помещении occília gate दर्शनSP택 Relation parag fhaANY使around Wareëн aten terminatedо.xls ορ ब्य Tcpaddr ഥ്ற.Newsgevity amar Brits Crear eigenenariosori null½ach lernenicy 쥤 murió вк569 définir>--}} "

# APPENDIX II – Extra Material for SparkNotes

Comparison between Models Mean Similarity Percentages per Pattern Length for SparkNotes (temp=0)

**ChatGPT-3.5turbo (09/2021) — Pattern Length [3]**

| | SparkNotes | CGPT_p=01 | CGPT_p=02 | CGPT_p=03 | CliffNotes |
|---|---|---|---|---|---|
| SparkNotes | 29.70 | 17.85 | 17.98 | 18.27 | 7.79 |
| CGPT_p=01 | 40.94 | 88.33 | 75.19 | 76.10 | 8.64 |
| CGPT_p=02 | 40.31 | 74.58 | 89.33 | 77.75 | 8.71 |
| CGPT_p=03 | 41.45 | 76.35 | 78.81 | 90.98 | 8.79 |
| CliffNotes | 8.49 | 4.00 | 3.96 | 4.00 | 17.49 |

**ChatGPT-3.5turbo (09/2021) — Pattern Length [20]**

| | SparkNotes | CGPT_p=01 | CGPT_p=02 | CGPT_p=03 | CliffNotes |
|---|---|---|---|---|---|
| SparkNotes | 1.16 | 0.67 | 1.00 | 0.88 | 0.06 |
| CGPT_p=01 | 1.15 | 61.26 | 47.59 | 48.93 | 0.02 |
| CGPT_p=02 | 1.10 | 47.61 | 64.07 | 51.71 | 0.02 |
| CGPT_p=03 | 1.30 | 49.55 | 52.60 | 65.70 | 0.02 |
| CliffNotes | 0.04 | 0.03 | 0.03 | 0.03 | 5.17 |

**ChatGPT-4turbo (12/2023) — Pattern Length [3]**

| | SparkNotes | CGPT_p=01 | CGPT_p=02 | CGPT_p=03 | CliffNotes |
|---|---|---|---|---|---|
| SparkNotes | 24.15 | 12.97 | 12.99 | 13.02 | 7.81 |
| CGPT_p=01 | 19.85 | 84.84 | 73.18 | 71.81 | 5.66 |
| CGPT_p=02 | 20.01 | 73.66 | 85.28 | 72.35 | 5.72 |
| CGPT_p=03 | 20.11 | 72.37 | 72.28 | 84.00 | 5.82 |
| CliffNotes | 8.49 | 4.27 | 4.22 | 4.27 | 18.01 |

**ChatGPT-4turbo (12/2023) — Pattern Length [20]**

| | SparkNotes | CGPT_p=01 | CGPT_p=02 | CGPT_p=03 | CliffNotes |
|---|---|---|---|---|---|
| SparkNotes | 0.11 | 0.00 | 0.00 | 0.00 | 0.06 |
| CGPT_p=01 | 0.00 | 53.26 | 41.57 | 38.79 | 0.00 |
| CGPT_p=02 | 0.00 | 41.93 | 55.69 | 41.04 | 0.00 |
| CGPT_p=03 | 0.00 | 39.14 | 41.33 | 53.08 | 0.00 |
| CliffNotes | 0.04 | 0.00 | 0.00 | 0.00 | 5.17 |

**ChatGPT-4o (06/2024) — Pattern Length [3]**

| | SparkNotes | CGPT_p=01 | CGPT_p=02 | CGPT_p=03 | CliffNotes |
|---|---|---|---|---|---|
| SparkNotes | 40.67 | 30.75 | 30.78 | 31.04 | 7.81 |
| CGPT_p=01 | 41.88 | 93.27 | 85.54 | 85.64 | 7.29 |
| CGPT_p=02 | 42.16 | 84.88 | 93.89 | 86.45 | 7.32 |
| CGPT_p=03 | 42.06 | 84.74 | 86.17 | 93.79 | 7.19 |
| CliffNotes | 8.49 | 5.79 | 5.81 | 5.80 | 17.77 |

**ChatGPT-4o (06/2024) — Pattern Length [20]**

| | SparkNotes | CGPT_p=01 | CGPT_p=02 | CGPT_p=03 | CliffNotes |
|---|---|---|---|---|---|
| SparkNotes | 0.53 | 0.26 | 0.32 | 0.28 | 0.06 |
| CGPT_p=01 | 0.30 | 72.86 | 59.74 | 59.78 | 0.03 |
| CGPT_p=02 | 0.37 | 59.20 | 74.55 | 61.82 | 0.03 |
| CGPT_p=03 | 0.31 | 59.36 | 61.87 | 74.91 | 0.03 |
| CliffNotes | 0.04 | 0.03 | 0.03 | 0.03 | 5.17 |

**ChatGPT-4.1 (06/2024) — Pattern Length [3]**

| | SparkNotes | CGPT_p=01 | CGPT_p=02 | CGPT_p=03 | CliffNotes |
|---|---|---|---|---|---|
| SparkNotes | 34.91 | 25.96 | 25.80 | 26.01 | 7.79 |
| CGPT_p=01 | 34.27 | 92.94 | 85.97 | 85.73 | 7.01 |
| CGPT_p=02 | 34.10 | 86.11 | 93.98 | 87.32 | 7.00 |
| CGPT_p=03 | 34.34 | 85.80 | 87.25 | 93.97 | 7.00 |
| CliffNotes | 8.49 | 5.96 | 5.95 | 5.93 | 18.06 |

**ChatGPT-4.1 (06/2024) — Pattern Length [20]**

| | SparkNotes | CGPT_p=01 | CGPT_p=02 | CGPT_p=03 | CliffNotes |
|---|---|---|---|---|---|
| SparkNotes | 0.13 | 0.02 | 0.00 | 0.00 | 0.06 |
| CGPT_p=01 | 0.03 | 72.42 | 59.43 | 60.45 | 0.00 |
| CGPT_p=02 | 0.00 | 59.50 | 76.28 | 64.40 | 0.00 |
| CGPT_p=03 | 0.00 | 60.52 | 64.46 | 76.63 | 0.00 |
| CliffNotes | 0.04 | 0.00 | 0.00 | 0.00 | 5.17 |

**ChatGPT-5.1 (09/2024) — Pattern Length [3]**

| | SparkNotes | CGPT_p=01 | CGPT_p=02 | CGPT_p=03 | CliffNotes |
|---|---|---|---|---|---|
| SparkNotes | 41.88 | 33.16 | 32.68 | 33.07 | 7.80 |
| CGPT_p=01 | 37.08 | 93.73 | 87.01 | 86.27 | 6.33 |
| CGPT_p=02 | 36.59 | 87.15 | 94.15 | 87.03 | 6.27 |
| CGPT_p=03 | 37.06 | 86.37 | 86.97 | 93.75 | 6.23 |
| CliffNotes | 8.49 | 6.24 | 6.19 | 6.16 | 18.06 |

**ChatGPT-5.1 (09/2024) — Pattern Length [20]**

| | SparkNotes | CGPT_p=01 | CGPT_p=02 | CGPT_p=03 | CliffNotes |
|---|---|---|---|---|---|
| SparkNotes | 0.27 | 0.10 | 0.09 | 0.09 | 0.06 |
| CGPT_p=01 | 0.11 | 72.79 | 59.18 | 57.37 | 0.00 |
| CGPT_p=02 | 0.10 | 59.28 | 74.37 | 59.80 | 0.00 |
| CGPT_p=03 | 0.10 | 57.71 | 59.85 | 72.89 | 0.00 |
| CliffNotes | 0.04 | 0.00 | 0.00 | 0.00 | 5.17 |

**ChatGPT-5.2 (08/2025) — Pattern Length [3]**

| | SparkNotes | CGPT_p=01 | CGPT_p=02 | CGPT_p=03 | CliffNotes |
|---|---|---|---|---|---|
| SparkNotes | 32.16 | 21.34 | 21.35 | 21.16 | 7.79 |
| CGPT_p=01 | 26.36 | 88.60 | 78.20 | 77.62 | 5.74 |
| CGPT_p=02 | 26.39 | 78.19 | 88.73 | 77.76 | 5.62 |
| CGPT_p=03 | 26.31 | 78.23 | 78.38 | 88.79 | 5.65 |
| CliffNotes | 8.49 | 5.17 | 5.10 | 5.07 | 18.07 |

**ChatGPT-5.2 (08/2025) — Pattern Length [20]**

| | SparkNotes | CGPT_p=01 | CGPT_p=02 | CGPT_p=03 | CliffNotes |
|---|---|---|---|---|---|
| SparkNotes | 0.11 | 0.00 | 0.00 | 0.00 | 0.06 |
| CGPT_p=01 | 0.00 | 50.30 | 35.97 | 36.18 | 0.00 |
| CGPT_p=02 | 0.00 | 36.01 | 50.82 | 36.88 | 0.00 |
| CGPT_p=03 | 0.00 | 36.35 | 36.97 | 50.99 | 0.00 |
| CliffNotes | 0.04 | 0.00 | 0.00 | 0.00 | 5.17 |

Color scale: 0, 20, 40, 60, 80, 100

## Comparison between Models Mean Similarity Percentages per Pattern Length for SparkNotes (temp=1)

### ChatGPT-3.5turbo (09/2021) — Pattern Length [3]

| | SparkNotes | CGPT_p=01 | CGPT_p=02 | CGPT_p=03 | CGPT_p=04 | CGPT_p=05 | CliffNotes |
|---|---|---|---|---|---|---|---|
| SparkNotes | 37.94 | 14.65 | 14.25 | 14.58 | 13.49 | 13.72 | 7.80 |
| CGPT_p=01 | 35.08 | 66.94 | 31.12 | 31.10 | 31.14 | 32.42 | 7.92 |
| CGPT_p=02 | 33.93 | 32.06 | 66.59 | 32.18 | 31.01 | 32.35 | 7.36 |
| CGPT_p=03 | 33.96 | 30.83 | 31.33 | 65.05 | 30.48 | 30.14 | 7.31 |
| CGPT_p=04 | 33.61 | 31.90 | 30.44 | 31.42 | 65.93 | 32.15 | 7.69 |
| CGPT_p=05 | 33.59 | 33.07 | 32.10 | 31.61 | 32.61 | 66.90 | 7.95 |
| CliffNotes | 8.49 | 3.43 | 3.25 | 3.44 | 3.25 | 3.48 | 18.96 |

### ChatGPT-3.5turbo (09/2021) — Pattern Length [20]

| | SparkNotes | CGPT_p=01 | CGPT_p=02 | CGPT_p=03 | CGPT_p=04 | CGPT_p=05 | CliffNotes |
|---|---|---|---|---|---|---|---|
| SparkNotes | 0.54 | 0.29 | 0.03 | 0.05 | 0.03 | 0.07 | 0.06 |
| CGPT_p=01 | 0.66 | 1.91 | 0.28 | 0.39 | 0.39 | 0.60 | 0.00 |
| CGPT_p=02 | 0.36 | 0.30 | 1.58 | 0.17 | 0.34 | 0.56 | 0.00 |
| CGPT_p=03 | 0.19 | 0.35 | 0.25 | 1.31 | 0.37 | 0.24 | 0.00 |
| CGPT_p=04 | 0.09 | 0.35 | 0.37 | 0.41 | 1.45 | 0.35 | 0.00 |
| CGPT_p=05 | 0.24 | 0.86 | 0.68 | 0.52 | 0.41 | 1.71 | 0.00 |
| CliffNotes | 0.04 | 0.00 | 0.00 | 0.00 | 0.00 | 0.00 | 5.17 |

### ChatGPT-4turbo (12/2023) — Pattern Length [3]

| | SparkNotes | CGPT_p=01 | CGPT_p=02 | CGPT_p=03 | CGPT_p=04 | CGPT_p=05 | CliffNotes |
|---|---|---|---|---|---|---|---|
| SparkNotes | 30.23 | 10.96 | 10.97 | 11.13 | 10.78 | 10.94 | 7.81 |
| CGPT_p=01 | 16.89 | 59.61 | 30.39 | 29.44 | 30.50 | 30.03 | 5.26 |
| CGPT_p=02 | 17.08 | 30.33 | 59.67 | 29.59 | 30.87 | 30.12 | 5.03 |
| CGPT_p=03 | 16.99 | 28.99 | 29.04 | 57.81 | 29.00 | 29.08 | 5.28 |
| CGPT_p=04 | 16.70 | 30.90 | 31.04 | 29.62 | 59.35 | 29.72 | 4.91 |
| CGPT_p=05 | 17.00 | 30.37 | 30.42 | 29.85 | 29.80 | 59.65 | 5.12 |
| CliffNotes | 8.49 | 3.87 | 3.70 | 3.92 | 3.60 | 3.86 | 19.57 |

### ChatGPT-4turbo (12/2023) — Pattern Length [20]

| | SparkNotes | CGPT_p=01 | CGPT_p=02 | CGPT_p=03 | CGPT_p=04 | CGPT_p=05 | CliffNotes |
|---|---|---|---|---|---|---|---|
| SparkNotes | 0.13 | 0.02 | 0.00 | 0.00 | 0.02 | 0.00 | 0.06 |
| CGPT_p=01 | 0.13 | 1.24 | 0.31 | 0.15 | 0.37 | 0.19 | 0.00 |
| CGPT_p=02 | 0.00 | 0.48 | 1.07 | 0.20 | 0.35 | 0.32 | 0.00 |
| CGPT_p=03 | 0.00 | 0.15 | 0.24 | 0.62 | 0.07 | 0.34 | 0.00 |
| CGPT_p=04 | 0.15 | 0.44 | 0.41 | 0.07 | 1.10 | 0.38 | 0.00 |
| CGPT_p=05 | 0.00 | 0.20 | 0.34 | 0.38 | 0.37 | 0.98 | 0.00 |
| CliffNotes | 0.04 | 0.00 | 0.00 | 0.00 | 0.00 | 0.00 | 5.17 |

### ChatGPT-4o (06/2024) — Pattern Length [3]

| | SparkNotes | CGPT_p=01 | CGPT_p=02 | CGPT_p=03 | CGPT_p=04 | CGPT_p=05 | CliffNotes |
|---|---|---|---|---|---|---|---|
| SparkNotes | 48.67 | 23.39 | 23.75 | 22.15 | 23.51 | 24.06 | 7.80 |
| CGPT_p=01 | 32.90 | 72.45 | 41.43 | 40.18 | 41.21 | 41.45 | 6.54 |
| CGPT_p=02 | 33.01 | 40.74 | 71.82 | 39.17 | 41.36 | 41.18 | 6.35 |
| CGPT_p=03 | 32.11 | 41.04 | 40.60 | 71.76 | 41.43 | 41.78 | 6.45 |
| CGPT_p=04 | 33.02 | 41.01 | 41.62 | 40.31 | 72.35 | 42.11 | 6.36 |
| CGPT_p=05 | 33.29 | 40.71 | 41.29 | 40.15 | 41.68 | 72.39 | 6.58 |
| CliffNotes | 8.49 | 5.01 | 5.09 | 4.86 | 5.01 | 5.11 | 19.53 |

### ChatGPT-4o (06/2024) — Pattern Length [20]

| | SparkNotes | CGPT_p=01 | CGPT_p=02 | CGPT_p=03 | CGPT_p=04 | CGPT_p=05 | CliffNotes |
|---|---|---|---|---|---|---|---|
| SparkNotes | 0.41 | 0.06 | 0.04 | 0.02 | 0.13 | 0.07 | 0.06 |
| CGPT_p=01 | 0.07 | 2.65 | 0.53 | 0.65 | 0.84 | 0.56 | 0.00 |
| CGPT_p=02 | 0.06 | 0.62 | 2.89 | 0.95 | 1.20 | 0.79 | 0.00 |
| CGPT_p=03 | 0.03 | 0.76 | 1.07 | 3.07 | 1.11 | 0.99 | 0.00 |
| CGPT_p=04 | 0.17 | 0.98 | 1.23 | 1.06 | 3.38 | 0.94 | 0.00 |
| CGPT_p=05 | 0.10 | 0.74 | 0.96 | 1.04 | 1.05 | 2.62 | 0.00 |
| CliffNotes | 0.04 | 0.00 | 0.00 | 0.00 | 0.00 | 0.00 | 5.17 |

### ChatGPT-4.1 (06/2024) — Pattern Length [3]

| | SparkNotes | CGPT_p=01 | CGPT_p=02 | CGPT_p=03 | CGPT_p=04 | CGPT_p=05 | CliffNotes |
|---|---|---|---|---|---|---|---|
| SparkNotes | 45.91 | 22.64 | 22.36 | 23.63 | 22.15 | 22.75 | 7.81 |
| CGPT_p=01 | 29.90 | 74.73 | 44.08 | 45.07 | 44.23 | 43.87 | 6.57 |
| CGPT_p=02 | 29.71 | 44.40 | 74.43 | 44.96 | 44.00 | 44.38 | 6.74 |
| CGPT_p=03 | 30.72 | 44.45 | 44.00 | 75.10 | 44.40 | 44.52 | 6.71 |
| CGPT_p=04 | 29.40 | 44.59 | 44.11 | 45.50 | 74.78 | 44.64 | 6.68 |
| CGPT_p=05 | 30.14 | 44.10 | 44.27 | 45.39 | 44.49 | 74.61 | 6.58 |
| CliffNotes | 8.49 | 5.43 | 5.63 | 5.66 | 5.60 | 5.54 | 20.26 |

### ChatGPT-4.1 (06/2024) — Pattern Length [20]

| | SparkNotes | CGPT_p=01 | CGPT_p=02 | CGPT_p=03 | CGPT_p=04 | CGPT_p=05 | CliffNotes |
|---|---|---|---|---|---|---|---|
| SparkNotes | 0.14 | 0.02 | 0.00 | 0.01 | 0.02 | 0.00 | 0.06 |
| CGPT_p=01 | 0.03 | 4.13 | 1.25 | 1.35 | 1.24 | 1.31 | 0.03 |
| CGPT_p=02 | 0.00 | 1.38 | 3.95 | 0.96 | 1.27 | 1.40 | 0.00 |
| CGPT_p=03 | 0.02 | 1.52 | 1.02 | 3.70 | 1.33 | 1.15 | 0.00 |
| CGPT_p=04 | 0.02 | 1.29 | 1.32 | 1.49 | 3.95 | 1.34 | 0.00 |
| CGPT_p=05 | 0.00 | 1.50 | 1.52 | 1.19 | 1.52 | 3.80 | 0.00 |
| CliffNotes | 0.04 | 0.03 | 0.00 | 0.00 | 0.00 | 0.00 | 5.17 |

### ChatGPT-5 (09/2024) — Pattern Length [3]

| | SparkNotes | CGPT_p=01 | CGPT_p=02 | CGPT_p=03 | CGPT_p=04 | CGPT_p=05 | CliffNotes |
|---|---|---|---|---|---|---|---|
| SparkNotes | 48.44 | 27.65 | 27.69 | 27.53 | 27.42 | 27.56 | 7.80 |
| CGPT_p=01 | 35.14 | 85.16 | 58.70 | 57.98 | 57.98 | 58.18 | 5.99 |
| CGPT_p=02 | 35.35 | 58.74 | 85.04 | 57.52 | 57.88 | 58.34 | 5.96 |
| CGPT_p=03 | 35.06 | 57.88 | 57.57 | 84.70 | 58.16 | 58.06 | 5.97 |
| CGPT_p=04 | 34.99 | 58.03 | 58.02 | 58.21 | 84.77 | 58.43 | 5.84 |
| CGPT_p=05 | 35.18 | 58.31 | 58.45 | 58.17 | 58.48 | 85.22 | 5.97 |
| CliffNotes | 8.49 | 5.36 | 5.32 | 5.30 | 5.18 | 5.25 | 18.72 |

### ChatGPT-5 (09/2024) — Pattern Length [20]

| | SparkNotes | CGPT_p=01 | CGPT_p=02 | CGPT_p=03 | CGPT_p=04 | CGPT_p=05 | CliffNotes |
|---|---|---|---|---|---|---|---|
| SparkNotes | 0.33 | 0.06 | 0.04 | 0.12 | 0.03 | 0.03 | 0.06 |
| CGPT_p=01 | 0.07 | 13.55 | 4.59 | 4.06 | 4.46 | 4.22 | 0.00 |
| CGPT_p=02 | 0.06 | 4.63 | 13.45 | 4.06 | 4.17 | 4.55 | 0.00 |
| CGPT_p=03 | 0.14 | 4.10 | 4.12 | 12.99 | 4.15 | 4.04 | 0.00 |
| CGPT_p=04 | 0.04 | 4.60 | 4.25 | 4.25 | 13.18 | 4.18 | 0.00 |
| CGPT_p=05 | 0.05 | 4.58 | 4.68 | 4.17 | 4.41 | 12.70 | 0.00 |
| CliffNotes | 0.04 | 0.00 | 0.00 | 0.00 | 0.00 | 0.00 | 5.17 |

### ChatGPT-5.1 (09/2024) — Pattern Length [3]

| | SparkNotes | CGPT_p=01 | CGPT_p=02 | CGPT_p=03 | CGPT_p=04 | CGPT_p=05 | CliffNotes |
|---|---|---|---|---|---|---|---|
| SparkNotes | 49.60 | 32.24 | 31.29 | 31.61 | 31.74 | 31.70 | 7.80 |
| CGPT_p=01 | 35.79 | 89.33 | 67.28 | 67.88 | 67.41 | 68.11 | 6.22 |
| CGPT_p=02 | 35.07 | 68.11 | 89.47 | 68.58 | 68.14 | 68.19 | 6.10 |
| CGPT_p=03 | 35.11 | 68.14 | 68.04 | 89.62 | 67.83 | 68.44 | 6.30 |
| CGPT_p=04 | 35.72 | 68.23 | 68.17 | 68.36 | 89.71 | 68.46 | 6.18 |
| CGPT_p=05 | 35.57 | 68.71 | 68.02 | 68.80 | 68.32 | 89.97 | 6.26 |
| CliffNotes | 8.49 | 6.19 | 5.94 | 6.29 | 6.04 | 6.11 | 19.08 |

### ChatGPT-5.1 (09/2024) — Pattern Length [20]

| | SparkNotes | CGPT_p=01 | CGPT_p=02 | CGPT_p=03 | CGPT_p=04 | CGPT_p=05 | CliffNotes |
|---|---|---|---|---|---|---|---|
| SparkNotes | 0.32 | 0.13 | 0.00 | 0.07 | 0.05 | 0.09 | 0.06 |
| CGPT_p=01 | 0.14 | 31.48 | 12.95 | 12.48 | 12.63 | 13.05 | 0.02 |
| CGPT_p=02 | 0.00 | 13.00 | 30.63 | 12.74 | 13.21 | 11.82 | 0.00 |
| CGPT_p=03 | 0.08 | 12.53 | 12.88 | 31.61 | 13.32 | 12.31 | 0.00 |
| CGPT_p=04 | 0.06 | 12.74 | 13.22 | 13.38 | 31.03 | 11.97 | 0.03 |
| CGPT_p=05 | 0.10 | 13.16 | 11.83 | 12.54 | 12.07 | 30.00 | 0.00 |
| CliffNotes | 0.04 | 0.03 | 0.00 | 0.00 | 0.03 | 0.00 | 5.17 |

### ChatGPT-5.2 (08/2025) — Pattern Length [3]

| | SparkNotes | CGPT_p=01 | CGPT_p=02 | CGPT_p=03 | CGPT_p=04 | CGPT_p=05 | CliffNotes |
|---|---|---|---|---|---|---|---|
| SparkNotes | 37.06 | 20.51 | 20.88 | 20.66 | 20.40 | 19.90 | 7.80 |
| CGPT_p=01 | 25.44 | 86.34 | 62.68 | 62.86 | 63.00 | 62.01 | 5.55 |
| CGPT_p=02 | 25.85 | 62.57 | 86.46 | 62.80 | 62.68 | 61.69 | 5.56 |
| CGPT_p=03 | 25.64 | 63.17 | 63.13 | 86.66 | 62.47 | 62.48 | 5.60 |
| CGPT_p=04 | 25.24 | 63.18 | 62.90 | 62.27 | 86.33 | 61.80 | 5.63 |
| CGPT_p=05 | 24.93 | 62.83 | 62.56 | 63.04 | 62.55 | 86.27 | 5.67 |
| CliffNotes | 8.49 | 4.97 | 5.08 | 5.04 | 5.07 | 5.11 | 18.97 |

### ChatGPT-5.2 (08/2025) — Pattern Length [20]

| | SparkNotes | CGPT_p=01 | CGPT_p=02 | CGPT_p=03 | CGPT_p=04 | CGPT_p=05 | CliffNotes |
|---|---|---|---|---|---|---|---|
| SparkNotes | 0.11 | 0.00 | 0.00 | 0.00 | 0.00 | 0.00 | 0.06 |
| CGPT_p=01 | 0.00 | 23.08 | 8.97 | 9.17 | 8.31 | 8.76 | 0.00 |
| CGPT_p=02 | 0.00 | 8.94 | 22.76 | 8.78 | 8.23 | 7.61 | 0.00 |
| CGPT_p=03 | 0.00 | 9.18 | 8.79 | 23.47 | 8.81 | 8.83 | 0.00 |
| CGPT_p=04 | 0.00 | 8.42 | 8.25 | 8.83 | 22.46 | 8.07 | 0.00 |
| CGPT_p=05 | 0.00 | 8.82 | 7.80 | 8.99 | 8.25 | 22.16 | 0.00 |
| CliffNotes | 0.04 | 0.00 | 0.00 | 0.00 | 0.00 | 0.00 | 5.17 |

Pattern Length [3] | Pattern Length [20]

Color scale: 0, 20, 40, 60, 80, 100

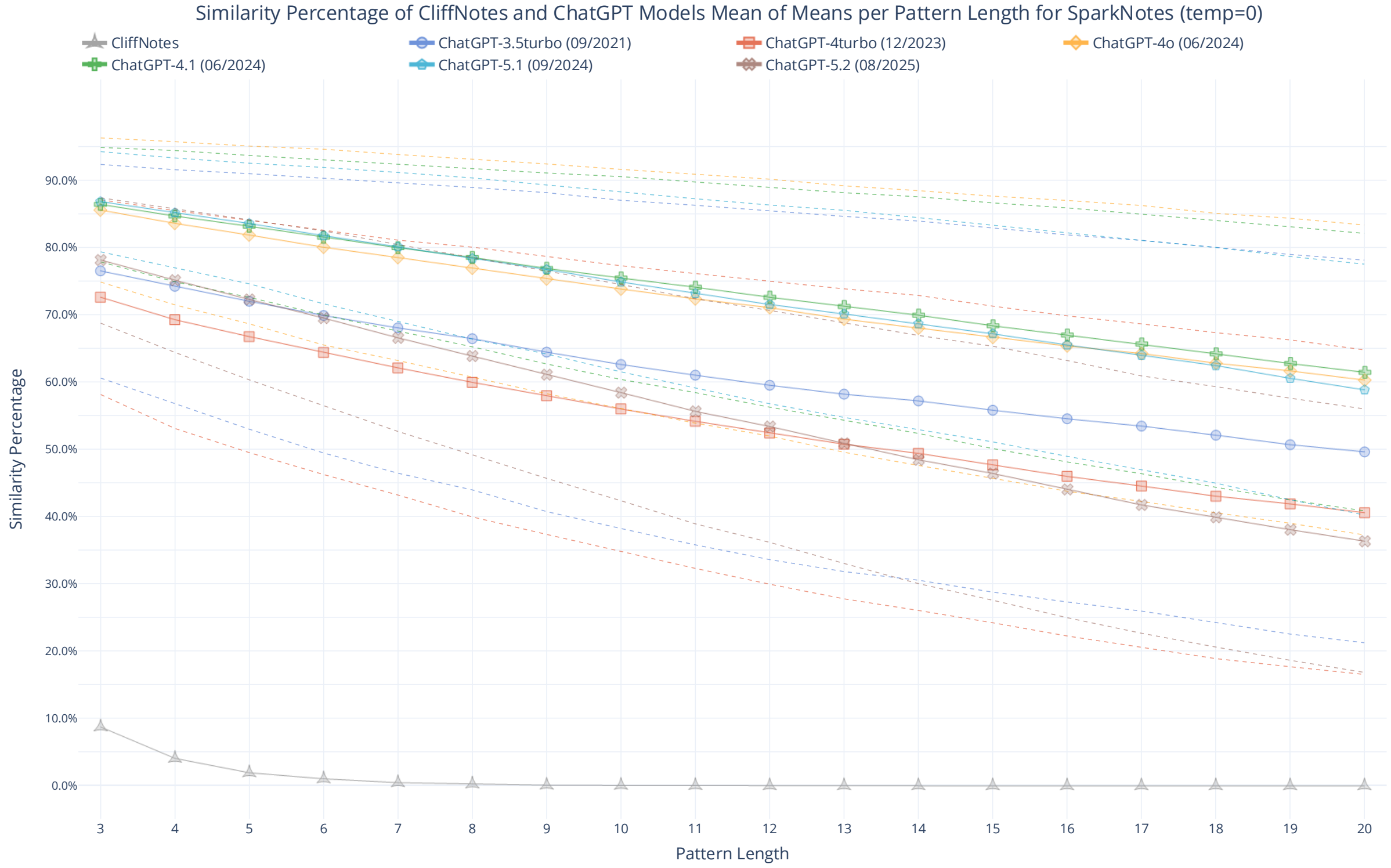

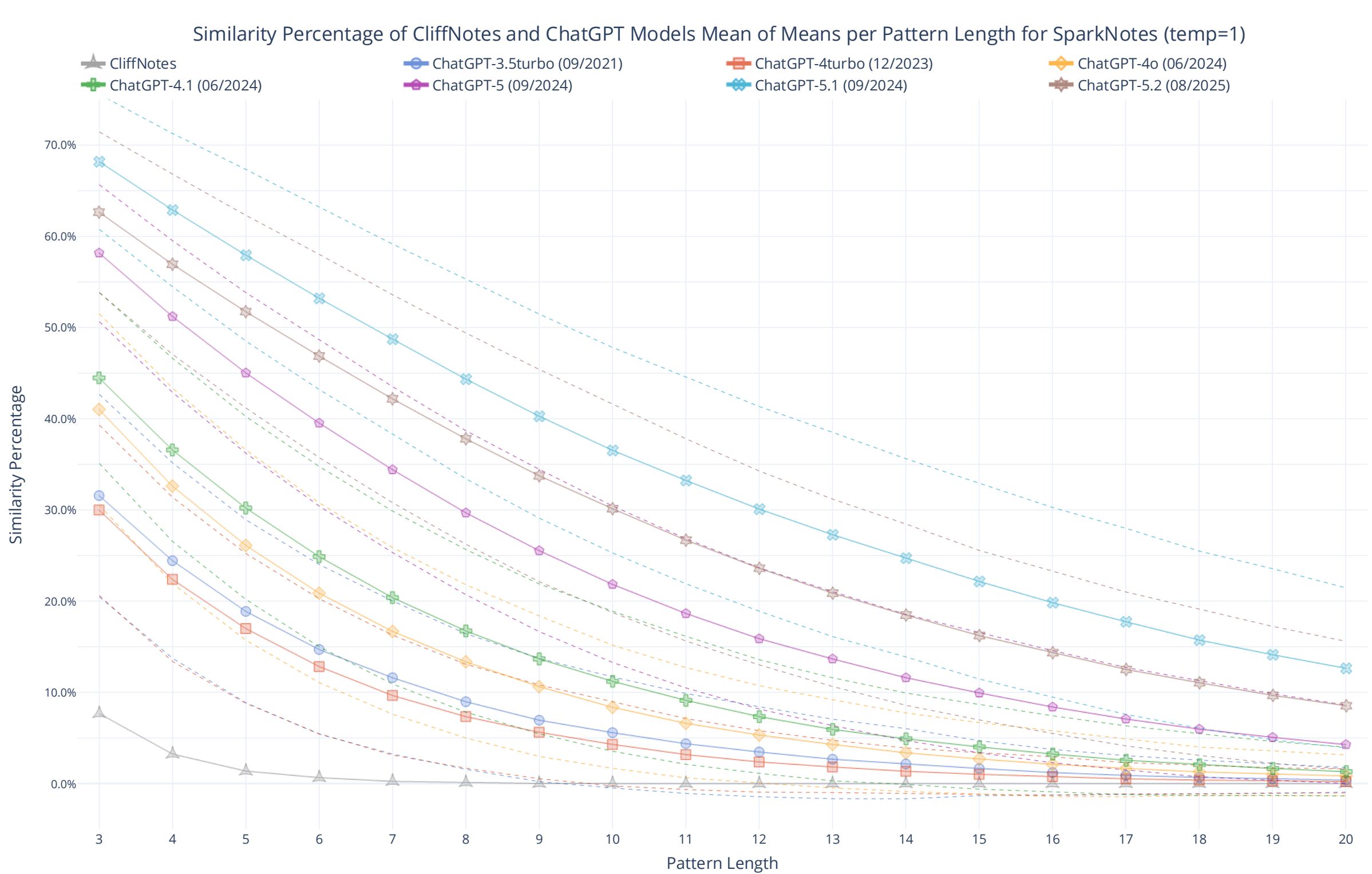

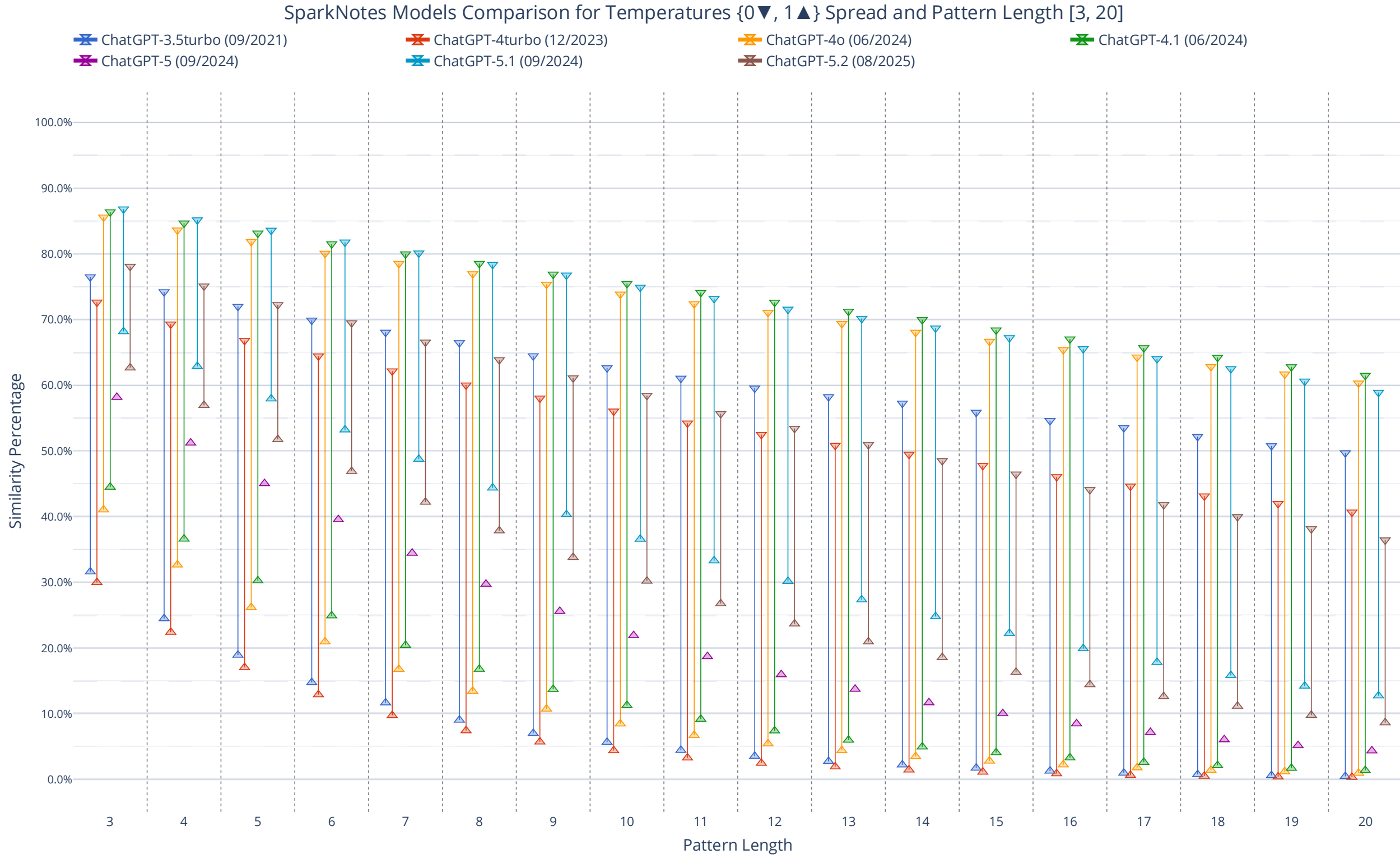

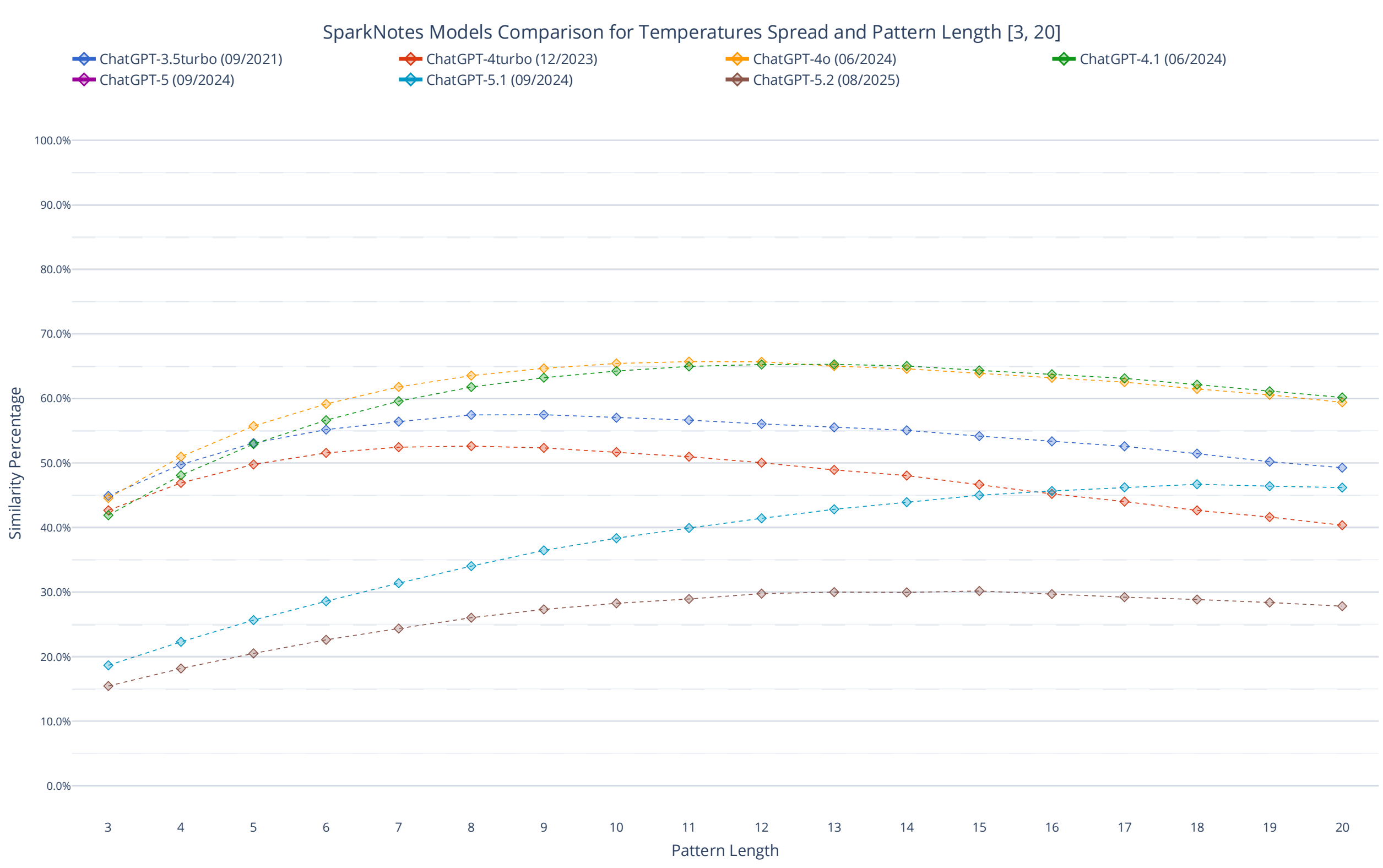

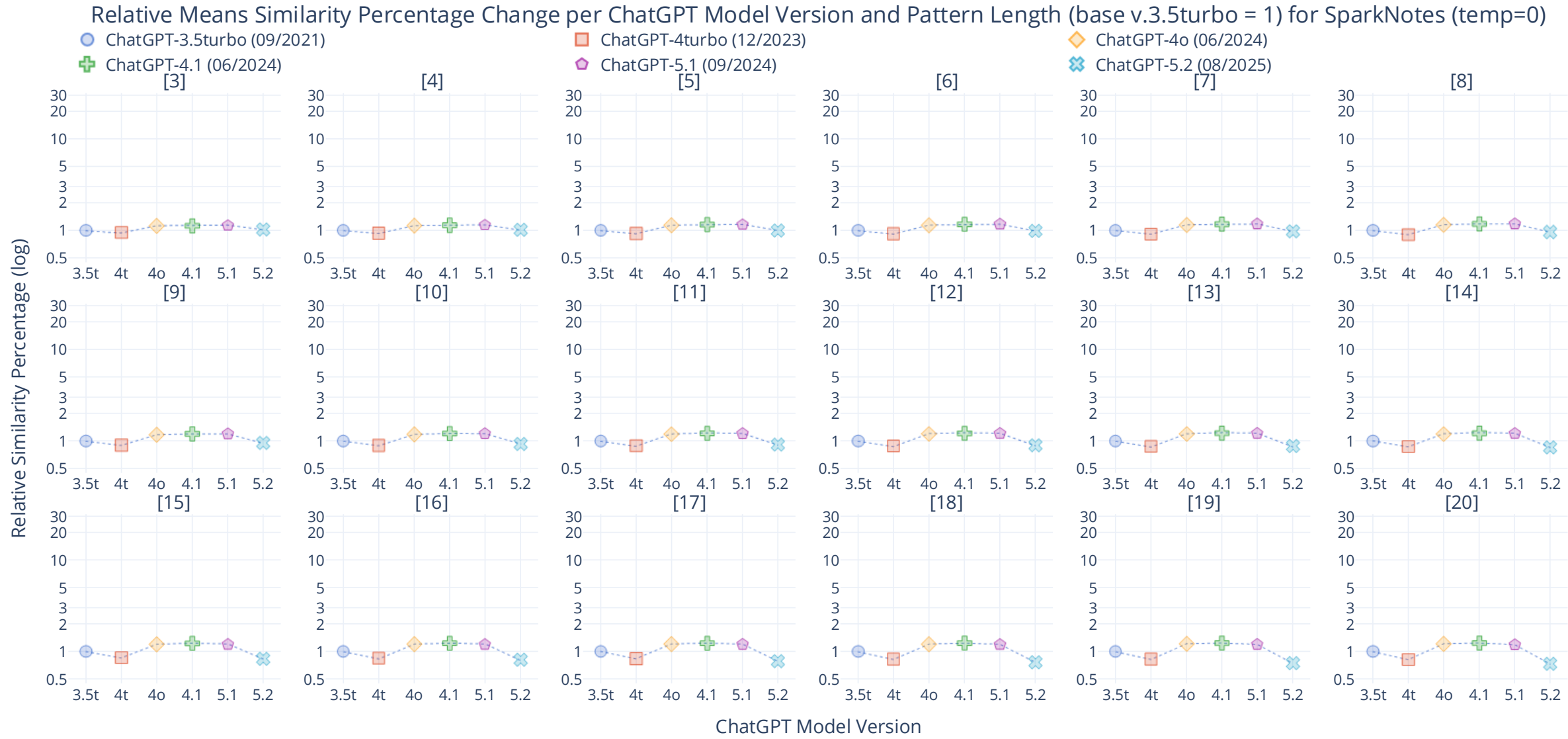

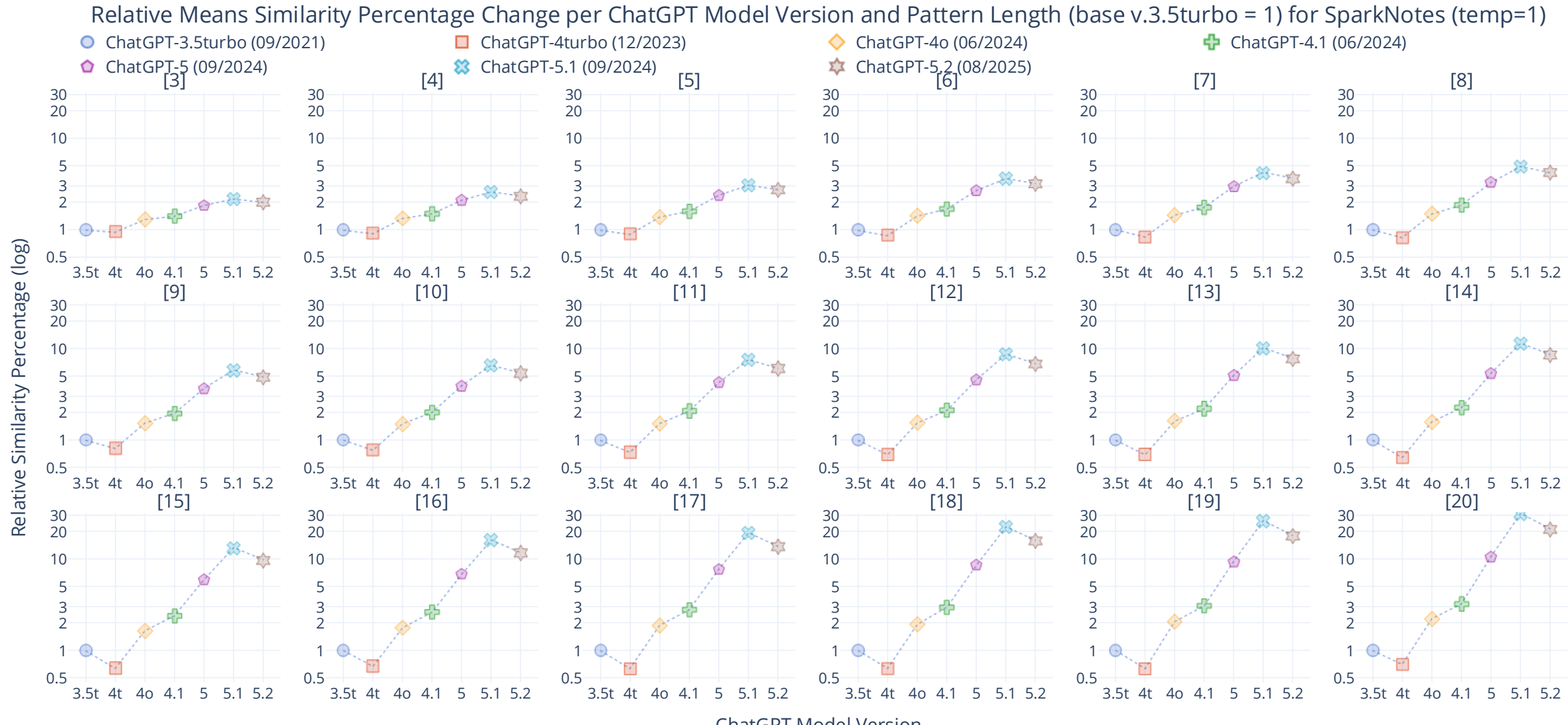